\RequirePackage[loading]{tracefnt}
\documentclass{article} 

\usepackage{microtype}
\usepackage{graphicx,wrapfig}
\usepackage{subfigure}
\usepackage{booktabs} 

\usepackage{hyperref}

\usepackage[accepted]{icml2024} 

\usepackage{amsmath}
\usepackage{amssymb}
\usepackage{mathtools}
\usepackage{amsthm}
\usepackage{pifont}

\usepackage[export]{adjustbox}
\usepackage{url}
\usepackage{threeparttable}
\usepackage{xspace}
\usepackage{dsfont}
\usepackage{bbm}
\usepackage[bottom]{footmisc}
\usepackage{algorithm}[H] 
\usepackage{xcolor,soul}
\usepackage[htt]{hyphenat}

\usepackage{color}
\newbox\statebox
\newcommand{\myState}[1]{%
    \setbox\statebox=\vbox{#1}%
    \edef\thealgruleheight{\dimexpr \the\ht\statebox+1pt\relax}%
    \edef\thealgruledepth{\dimexpr \the\dp\statebox+1pt\relax}%
    \ifdim\thealgruleheight<.75\baselineskip
        \def\thealgruleheight{\dimexpr .75\baselineskip+1pt\relax}%
    \fi
    \ifdim\thealgruledepth<.25\baselineskip
        \def\thealgruledepth{\dimexpr .25\baselineskip+1pt\relax}%
    \fi
    \State #1%
    \def\thealgruleheight{\dimexpr .75\baselineskip+1pt\relax}%
    \def\thealgruledepth{\dimexpr .25\baselineskip+1pt\relax}%
}

\definecolor{perf}{HTML}{E1144B}
\definecolor{dist}{HTML}{0053D6}

\sethlcolor{lightgray}

\newcommand{\cmark}{\ding{51}}%
\newcommand{\xmark}{\ding{55}}%
\newcommand\numberthis{\addtocounter{equation}{1}\tag{\theequation}}

\usepackage{amsmath,amsfonts,bm}

\def\eqref#1{equation~\ref{#1}}

\def\1{\bm{1}}

\def\eps{{\epsilon}}

\DeclareMathAlphabet{\mathsfit}{\encodingdefault}{\sfdefault}{m}{sl}
\SetMathAlphabet{\mathsfit}{bold}{\encodingdefault}{\sfdefault}{bx}{n}

\DeclareMathOperator*{\argmin}{arg\,min}

\renewcommand{\vec}[1]{{\boldsymbol{\mathbf{#1}}}}
\newcommand{\mat}[1]{{\boldsymbol{\mathbf{#1}}}}

\newcommand{\state}{s}
\newcommand{\stateSpace}{\mathcal{S}}
\newcommand{\stateDist}{\rho}

\newcommand{\action}{a}
\newcommand{\actionSpace}{\mathcal{A}}

\newcommand{\reward}{r}
\newcommand{\feat}{\vec \phi}
\newcommand{\featSpace}{\Phi}

\newcommand{\discount}{\gamma}

\newcommand{\goal}{\vec g}

\newcommand{\skill}{\vec z}
\newcommand{\skillDiayn}{\vec z_\text{DIAYN}}

\newcommand{\skillImg}{\widetilde{\skill}}
\newcommand{\skillEnv}{\skill}
\newcommand{\skillSpace}{\mathcal{Z}}

\newcommand{\horizonImg}{H}
\newcommand{\replay}{\mathcal{D}}

\newcommand{\params}{\theta}
\newcommand{\paramsV}{\theta_{\valueFunction}}
\newcommand{\paramsSF}{\theta_{\successorFeatures}}
\newcommand{\temperatureSac}{\beta}
\newcommand{\policy}{\pi}
\newcommand{\policySkill}{\policy_\skill}

\newcommand{\lagrange}{\lambda}
\newcommand{\valueFunction}{V}
\newcommand{\QFunction}{Q}
\newcommand{\successorFeatures}{\vec \psi}
\newcommand{\featAvg}{\expect{\policySkill}{\feat(\state, \action)}}

\newcommand{\wm}{\mathcal{W}}
\newcommand{\stateImg}{\widetilde{s}}
\newcommand{\cont}{c}
\newcommand{\successorFeaturesLambda}{\vec \psi_\lambda}
\newcommand{\valueFunctionLambda}{V_\lambda}

\newcommand{\uniform}{\mathcal{U}}
\newcommand{\expectSymbol}[1]{\mathbb{E}_{{#1}}}
\newcommand{\expect}[2]{\mathbb{E}_{{#1}} \left[{#2}\right]}

\newcommand{\norm}[1]{\left\lVert#1\right\rVert_2}
\newcommand{\threshold}{\delta}
\newcommand{\sg}[1]{\mathrm{sg}\left( { #1 } \right)}

\newcommand{\defeq}{\vcentcolon=}

\newcommand{\ours}{QDAC\xspace}
\newcommand{\oursmb}{QDAC-MB\xspace}
\newcommand{\oursLong}{Quality-Diversity Actor-Critic\xspace}

\newcommand{\ppga}{PPGA\xspace}
\newcommand{\dcgme}{DCG-ME\xspace}
\newcommand{\qdpg}{QD-PG\xspace}
\newcommand{\me}{MAP-Elites\xspace}

\newcommand{\cmamaega}{CMA-MAEGA\xspace}

\newcommand{\domino}{DOMiNO\xspace}
\newcommand{\smerl}{SMERL\xspace}
\newcommand{\smerlReverse}{Reverse SMERL\xspace}

\newcommand{\uvfa}{UVFA\xspace}
\newcommand{\oursFixedLambda}{Fixed-$\lambda$\xspace}
\newcommand{\oursSepSkill}{No-SF\xspace}

\newcommand{\sac}{SAC\xspace}
\newcommand{\diayn}{DIAYN\xspace}

\newcommand{\dreamer}{DreamerV3\xspace}

\newcommand{\aurora}{AURORA\xspace}

\makeatletter
\newcommand{\subalign}[1]{%
  \vcenter{%
    \Let@ \restore@math@cr \default@tag
    \baselineskip\fontdimen10 \scriptfont\tw@
    \advance\baselineskip\fontdimen12 \scriptfont\tw@
    \lineskip\thr@@\fontdimen8 \scriptfont\thr@@
    \lineskiplimit\lineskip
    \ialign{\hfil$\m@th\scriptstyle##$&$\m@th\scriptstyle{}##$\hfil\crcr
      #1\crcr
    }%
  }%
}
\makeatother

\newlength\dlf
\newcommand\alignedbox[2]{
  &
  \begingroup
  \settowidth\dlf{$\displaystyle #1$}
  \addtolength\dlf{\fboxsep+\fboxrule}
  \hspace{-\dlf}
  \boxed{#1 #2}
  \endgroup
}

\newtheorem*{proposition*}{Proposition}
\newtheorem{propositionAppendix}{Proposition}[section]

\icmltitlerunning{Quality-Diversity Actor-Critic}

\begin{document}

\twocolumn[
\icmltitle{Quality-Diversity Actor-Critic:\\Learning High-Performing and Diverse Behaviors\\via Value and Successor Features Critics}

\icmlsetsymbol{equal}{*}

\begin{icmlauthorlist}
\icmlauthor{Luca Grillotti}{equal,comp}
\icmlauthor{Maxence Faldor}{equal,comp}
\icmlauthor{Borja González León}{comp,iconic}
\icmlauthor{Antoine Cully}{comp}
\end{icmlauthorlist}

\icmlaffiliation{comp}{Department of Computing, Imperial College London, London, United Kingdom}
\icmlaffiliation{iconic}{Iconic AI}

\icmlcorrespondingauthor{Luca Grillotti}{luca.grillotti16@imperial.ac.uk}

\icmlkeywords{Reinforcement Learning, Quality-Diversity}

\vskip 0.3in
]

\printAffiliationsAndNotice{\icmlEqualContribution} 

\begin{abstract}
A key aspect of intelligence is the ability to demonstrate a broad spectrum of behaviors for adapting to unexpected situations.
Over the past decade, advancements in deep reinforcement learning have led to groundbreaking achievements to solve complex continuous control tasks.
However, most approaches return only one solution specialized for a specific problem.
We introduce \oursLong (\ours), an off-policy actor-critic deep reinforcement learning algorithm that leverages a value function critic and a successor features critic to learn high-performing and diverse behaviors.
In this framework, the actor optimizes an objective that seamlessly unifies both critics using constrained optimization to (1) maximize return, while (2) executing diverse skills.
Compared with other Quality-Diversity methods, \ours achieves significantly higher performance and more diverse behaviors on six challenging continuous control locomotion tasks.
We also demonstrate that we can harness the learned skills to adapt better than other baselines to five perturbed environments.
Finally, qualitative analyses showcase a range of remarkable behaviors:
\href{https://adaptive-intelligent-robotics.github.io/QDAC/}{adaptive-intelligent-robotics.github.io/QDAC}.
\end{abstract}

\newcommand{\MyLabel}[1]{\label{\MyTag#1}}
\newcommand{\MyRef}[1]{\ref{\MyTag#1}}

\newcommand{\MyTag}{main:}
\section{Introduction}
\label{sec:introduction}
Reinforcement Learning (RL) has enabled groundbreaking achievements like mastering discrete games~\citep{mnih_PlayingAtariDeep_2013,silver_MasteringGameGo_2016} but also continuous control domains for locomotion~\citep{haarnoja_SoftActorCriticAlgorithms_2019,heess_EmergenceLocomotionBehaviours_2017}. These milestones have showcased the extraordinary potential of RL algorithms in solving specific problems.

In contrast, human intelligence is beyond mastering a single task, and adapts to unforeseen environments by combining skills.
Empowering artificial agents with diverse skills was shown to improve exploration~\citep{gehring_HierarchicalSkillsEfficient_2021}, to facilitate knowledge transfer~\citep{eysenbach_DiversityAllYou_2018}, to enable hierarchical problem-solving~\citep{allard_HierarchicalQualitydiversityOnline_2022}, to enhance robustness and adaptation~\citep{kumar_OneSolutionNot_2020,cully_RobotsThatCan_2015} and finally, to foster creativity~\citep{zahavy_DiversifyingAICreative_2023,lehman_SurprisingCreativityDigital_2020}.

\begin{figure*}[t]
\centering
\includegraphics[width=0.99\textwidth]{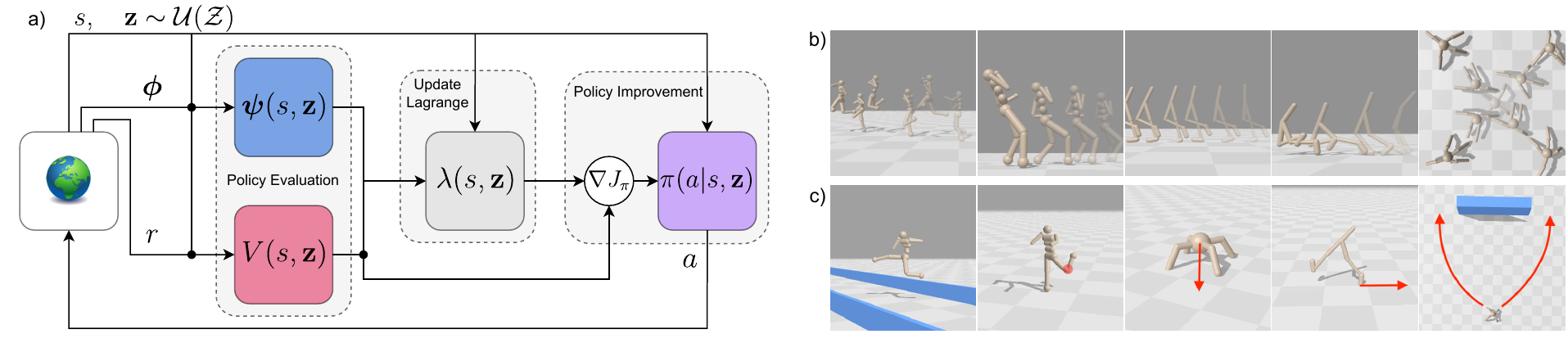}

\caption{%
\textbf{a)} \ours's architecture: the agent $\policy(\action | \state, \skill)$ learns high-performing and diverse behaviors with a dual critics optimization $\valueFunction(\state, \skill)$ and $\successorFeatures(\state, \skill)$ which are balanced with a Lagrange multiplier $\lagrange(\state, \skill)$. \textbf{b)} Example of diverse behaviors on a set of challenging continuous control tasks. \textbf{c)} Few-shot adaptation tasks and hierarchical learning tasks using the diversity of skills learned by \ours.}
\end{figure*}

Following this observation, methods have been developed to make agents more versatile, including Goal-Conditioned Reinforcement Learning (GCRL)~\citep{liu_GoalConditionedReinforcementLearning_2022}, Unsupervised Reinforcement Learning (URL)~\citep{eysenbach_DiversityAllYou_2018,sharma_DynamicsAwareUnsupervisedDiscovery_2019}, and reward design~\citep{margolis_WalkTheseWays_2022}.
However, designing algorithms to learn expressive skills that are useful to solve downstream tasks remains a challenge.
Reward design requires a lot of manual work and fine-tuning while being very brittle. GCRL and URL try to achieve goals or execute skills while disregarding other objectives like safety or efficiency, leaving a gap in our quest for machines that can execute expressive and optimal skills to solve complex tasks.

Quality-Diversity (QD) optimization~\cite{pugh_QualityDiversityNew_2016} is a family of methods, originating from Evolutionary Algorithms, that generate a diverse population of high-performing solutions.
QD algorithms have shown promising results in hard-exploration settings~\citep{ecoffet_FirstReturnThen_2021b}, to recover from damage~\citep{cully_RobotsThatCan_2015} or to reduce the reality gap~\citep{chatzilygeroudis_Reset_2018}.
In particular, QD algorithms have been scaled to challenging, continuous control tasks, by synergizing evolutionary methods with reinforcement learning~\cite{faldor_Synergizing_2023,pierrot_DiversityPolicyGradient_2022}.
Other approaches like \smerl~\citep{kumar_OneSolutionNot_2020} and \domino~\citep{zahavy_DiscoveringPoliciesDOMiNO_2022} share the same objective of finding diverse and near-optimal policies and optimize a quality-diversity trade-off employing a pure reinforcement learning formulation.
Most QD algorithms guide the diversity search towards relevant behaviors using a manually defined behavior descriptor function, that meaningfully characterizes solutions for the type of diversity desired~\cite{cully_QualityDiversityOptimization_2018,mouret_IlluminatingSearchSpaces_2015}. Two notable exceptions are \aurora~\cite{grillotti_UnsupervisedBehaviourDiscovery_2022} and \smerl, that learn an unsupervised diversity measure, using an autoencoder architecture and \diayn respectively.

In this work, we aim to solve the Quality-Diversity problem, i.e. to learn a large number of high-performing and diverse behaviors, where the diversity measure is given as part of the task, as a function of state-action occupancy (see Section~\ref{sec:problem-statement} for a detailed problem statement).
First, we introduce an approximate policy skill improvement update based on successor features, analogous to the classic policy improvement update based on value function (Section~\ref{sec:methods:actorObjective}).
Second, we show that the policy skill improvement update based on successor features enables the policy to efficiently learn to execute skills with a theoretical justification (see Proposition in Section~\ref{sec:methods:actorObjective}).
Third, we formalize the goal of Quality-Diversity into a problem that seamlessly unifies value function and successor features critics using constrained optimization to (1) maximize performance, while (2) executing desired skills (see Problem~\ref{eq:problem-3} in Section~\ref{sec:methods:actorObjective}).
Finally, we introduce \oursLong (\ours), a practical algorithm that solves this problem by leveraging two independent critics --- the value function criticizes the actions made by the actor to improve quality while the successor features criticizes the actions made by the actor to improve diversity (Section~\ref{sec:methods:practicalAlgorithm}).

We evaluate our approach on six continuous control tasks and show that \ours achieves 15\% more diverse behaviors and 38\% higher performance than other baselines (Section~\ref{sec:results-qd}). Finally, we show that the skills can be used to adapt to downstream tasks in a few shots or via hierarchical learning (Section~\ref{sec:results-adaptation}).

\section{Background}
\label{sec:background}
We consider the reinforcement learning framework~\citep{sutton_ReinforcementLearningIntroduction_2018} where an agent interacts with a \emph{Markov Decision Process} (MDP) to maximize the expected sum of rewards. At each time step $t$, the agent observes a \emph{state} $\state_t \in \stateSpace$ and takes an \emph{action} $\action_t \in \actionSpace$, which causes the environment to transition to a next state $\state_{t+1} \in \stateSpace$, sampled from the dynamics $p(\state_{t+1} \mid \state_t, \action_t)$. Additionally, the agent receives a reward $\reward_t = r(\state_t, \action_t)$ and observes features $\feat_t = \feat(\state_t, \action_t) \in \featSpace \subset \mathbb{R}^d$. In this work, we assume the features $\feat_t$ are provided by the environment as part of the task, akin to the rewards, and are not learned by the agent. We denote $\stateDist^\pi(\state) = \lim_{t\rightarrow\infty} P(\state_t = \state | \state_0, \policy)$ the stationary distribution of states under a policy $\policy$, which we assume exists and is independent of $\state_0$~\citep{sutton_PolicyGradientMethods_1999}.

The objective of the agent is to find a policy $\policy$ that maximizes the expected discounted sum of rewards, or expected return $\expect{\policy}{\sum_{t} \discount^t \reward_{t}}$. The so-called value-based methods in RL rely on the concept of \emph{value function} $\valueFunction^\policy(\state)$, defined as the expected return obtained when starting from state $\state$ and following policy $\policy$ thereafter~\citep{puterman_MarkovDecisionProcesses_1994}: $\valueFunction^{\policy}(\state) = \expect{\policy}{\left.\sum_{i=0}^\infty \discount^i \reward_{t+i} \right\vert \state_t = \state}$.
In this work, the value function is approximated via a neural network parameterized by $\paramsV$.
Similarly to \citet{mnih_PlayingAtariDeep_2013}, those parameters are optimized by minimizing the Bellman error:
\begin{equation}
    \label{eq:lossV}
    J_V(\paramsV) = \expect{\policy}{ \left( \valueFunction_{\paramsV} (\state_t) -  r_t - \gamma \valueFunction_{\paramsV '} (\state_{t+1})  \right)^2 }
\end{equation}
where $\paramsV '$ are the parameters of a target network, which are updated at a lower pace to improve training stability~\citep{mnih2015human}.

In addition to the value function, we also leverage the concept of \emph{successor features} $\successorFeatures^{\policy}(\state)$, which is the expected discounted sum of features obtained when starting from state $\state$ and following policy $\policy$ thereafter~\citep{barreto_SuccessorFeaturesTransfer_2017}: $\successorFeatures^{\policy}(\state) = \expect{\policy}{\left.\sum_{i=0}^{\infty} \discount^i \feat_{t+i} \right\vert \state_t = \state}$. The successor features captures the expected features under a given policy, offering insights into the agent's future behavior and satisfies a Bellman equation in which $\feat_t$ plays the role of the reward $\successorFeatures^\policy(\state) = \expect{\policy}{\feat_t + \discount \successorFeatures^\policy(\state_{t+1}) | \state_t = s}$, and can be learned with any RL methods~\citep{dayan_ImprovingGeneralizationTemporal_1993}.
In this work specifically, the successor features are approximated via a neural network parameterized by $\paramsSF$.
Analogously to the value function network, $\paramsSF$ is optimized by minimizing the Bellman error:
\begin{equation}
    \label{eq:lossSF}
    J_{\successorFeatures}(\paramsSF) = \expect{\policy}{ \norm{ \successorFeatures_{\paramsSF} (\state_t) -  \feat_t - \gamma \successorFeatures_{\paramsSF '} (\state_{t+1}) }^2 }
\end{equation}
where $\paramsSF '$ are the parameters of the corresponding target network. 

In practice, we make use of a universal value function approximator $\valueFunction^{\policy}(\state, \skill)$~\citep{schaul_UniversalValueFunction_2015} and of a universal successor features approximator $\successorFeatures^{\policy}(\state, \skill)$~\citep{borsa_UniversalSuccessorFeatures_2018} that depend on state $\state$ but also on the skill $\skill$ conditioning the policy. The value function quantifies the performance while the successor features characterizes the behavior of the agent. For conciseness, we omit $\policy$ from the notations $\stateDist^\policy$, $\valueFunction^{\policy}$, $\successorFeatures^{\policy}$ and we note $\policySkill(\action | \state) \defeq \policy(\action | \state, \skill)$.

\section{Problem Statement}
\label{sec:problem-statement}
In this work, we aim to solve the Quality-Diversity problem, i.e. to learn a policy that can execute a large number of different and high-performing behaviors. In this section, we formalize this intuitive goal into a concrete optimization problem.
The behavior of a policy $\policy$ is characterized by the expected features under the policy's stationary distribution, $\lim_{T \rightarrow \infty} \frac{1}{T} \sum_{t=0}^{T-1} \feat_t = \expect{\policy}{\feat(\state, \action)}$ and we define the space of all possible behaviors to be the skill space $\skillSpace$.

Given this definition, we intend to learn a skill-conditioned policy $\policy(\action | \state, \skill)$ that (1) maximizes the expected return, and (2) is subject to the expected features converge to the desired skill $\skill$. In other words, we solve the following constrained optimization problem, for all $\skill \in \skillSpace$,
\begin{align}
\label{eq:problem-1}
\begin{split}
&\text{maximize } \expect{\policySkill}{\sum_{i=0}^{\infty} \discount^i \reward_{t+i}}\\
&\text{subject to } \featAvg = \skill
\end{split}
\tag{P1}
\end{align}
The feature function $\feat$ can be any arbitrary function of the state of the MDP and of the action taken by the agent.
Computing diversity based on the raw observations in high-dimensional environments (e.g., pixel observations) may not lead to interesting behaviors.
Thus, the features can be thought of as relevant characteristics or events for the type of diversity desired, such as joint positions, contact with the ground, speed and so on.
To illustrate the generality of this problem formulation, we now give two examples.
Consider a robot whose objective is to minimize energy consumption, and where the features characterize the velocity of the robot $\feat_t = \vec v_t = \begin{bmatrix}v_x(t) & v_y(t)\end{bmatrix}^\intercal$ and the skill space $\skillSpace = \mathbb{R}^2$. For each desired velocity $\skill \in \skillSpace$, $\policy(\action | \state, \skill)$ is expected to (1) minimize energy consumption, while (2) following the desired velocity $\skill$ in average, $\lim_{T\rightarrow\infty} \frac{1}{T} \sum_{t=0}^{T-1} \vec v_t = \skill$.

Now consider another example with a legged robot, where the objective is to go forward as fast as possible, and the features characterize which foot is in contact with the ground at each time step. For example, $\feat_t = \begin{bmatrix}1 & 0\end{bmatrix}^\intercal$ for a biped robot that is standing on its first leg and with the second leg not touching the ground at time step $t$.
With these features, the $i$-th component of the skill $\skill$ (i.e. average features) will be the proportion of time during which the $i$-th foot of the robot is in contact with the ground, denoted as feet contact rate.
In that case, the skill space characterizes the myriad of ways the robot can walk and specifically, how often each leg is being used. Notice that to achieve a feet contact of $\skill = \begin{bmatrix}0.1 & 0.6\end{bmatrix}^\intercal$, the robot needs to use 10\% of the time the first foot and 60\% of the time the second foot over a trajectory of \textit{multiple} time steps.

\section{Methods}
\label{sec:methods}
In this section, we present \oursLong (\ours), a quality-diversity reinforcement learning algorithm that discovers high-performing and diverse skills.
First, we present a concrete optimization problem that optimizes for quality and diversity as defined in Section~\ref{sec:problem-statement}.
Second, we define the notion of successor features policy iteration that we combine with value function policy iteration to derive a tractable objective for the actor, that solves problem~\ref{eq:problem-1} approximately.
Third, we derive a practical algorithm that optimizes this objective.

\subsection{Actor Objective}
\label{sec:methods:actorObjective}
\begin{figure}[t]
\centering
\includegraphics[width=0.45\textwidth]{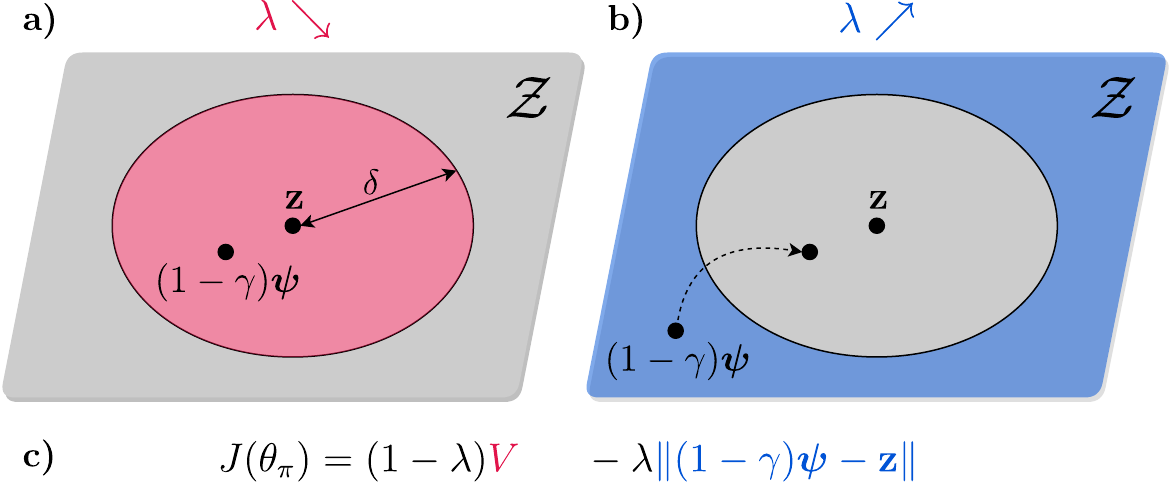}

\caption{%
The Lagrange multiplier is optimized to balance the quality-diversity trade-off, see Eq.~\ref{eq:lagrange-obj}.
\textbf{a)} If the expected features $(1 - \discount) \successorFeatures(\state, \skill)$ is in the neighborhood of $\skill$, then $\lagrange(\state, \skill)$ decreases to focus on maximizing the return.
\textbf{b)} Otherwise, $\lambda(\state, \skill)$ increases to focus on executing $\skill$.
\textbf{c)} After the Lagrange multiplier is updated, the policy is optimized according to the objective.
}
\label{fig:lagrangian-function}
\end{figure}

First, we relax the constraint from Problem~\ref{eq:problem-1} using the $L^2$ norm and $\threshold$, a threshold that quantifies the maximum acceptable distance between the desired skill and the expected features. We solve the optimization problem, for all $\skill \in \skillSpace$,
\begin{align}
\label{eq:problem-2}
\begin{split}
&\text{maximize } \expect{\policySkill}{\sum_{i=0}^{\infty} \discount^i \reward_{t+i}}\\
&\text{subject to } \norm{\featAvg - \skill} \leq \threshold
\end{split}
\tag{P2}
\end{align}
Second, we derive an upper bound for the distance between desired skill and expected features, whose proof is provided in Appendix~\ref{appendix:proofs}. A similar proposition is proven in a more general case in Appendix~\ref{proposition-2}. The goal is to minimize the bound so that the constraint in Problem~\ref{eq:problem-2} is satisfied.
\begin{proposition*}
\MyLabel{proposition-1}
Consider an infinite horizon, finite MDP with observable features in $\featSpace$. Let $\policy$ be a policy and let $\successorFeatures$ be the discounted successor features.
Then, for all skills $\skill \in \skillSpace$, we can derive an upper bound for the distance between $\skill$ and the expected features under $\policy$:
\begin{equation}
\MyLabel{eq:upper-bound}
\norm{\featAvg - \skill} \leq \expect{\policySkill}{\norm{(1 - \discount) \successorFeatures(\state, \skill) - \skill}}
\end{equation}
\end{proposition*}

Third, we derive a new Problem~\ref{eq:problem-3} by replacing the intractable constraint from Problem~\ref{eq:problem-2} with the tractable upper bound in Equation~\MyRef{eq:upper-bound}. 
The constraint in Problem~\ref{eq:problem-3} is more restrictive than that of Problem~\ref{eq:problem-2}.
Indeed, the above proposition ensures that if the constraint in Problem~\ref{eq:problem-3} is satisfied, then the constraint in Problem~\ref{eq:problem-2} is necessarily satisfied as well. 
For all $\skill \in \skillSpace$,
\begin{align}
\label{eq:problem-3}
\begin{split}
&\text{maximize } \expect{\policySkill}{{\color{perf} \valueFunction(\state, \skill)}}\\
&\text{subject to } \expect{\policySkill}{{\color{dist} \norm{(1 - \discount) \successorFeatures(\state, \skill) - \skill}}} \leq \threshold
\end{split}
\tag{P3}
\end{align}
Finally, we solve Problem~\ref{eq:problem-3} using the method of Lagrange multipliers as described by~\citet{abdolmaleki_MaximumPosterioriPolicy_2018,abdolmaleki_MultiobjectivePolicyOptimization_2023}. For all states $\state$, and all skills $\skill \in \skillSpace$, we maximize the Lagrangian function, subject to $0 \leq \lagrange(\state, \skill) \leq 1$,
\begin{equation}
\label{eq:actor-obj}
\left(1 - \lagrange(\state, \skill) \right) {\color{perf} \valueFunction(\state, \skill)} - \lagrange(\state, \skill) {\color{dist} \norm{ (1-\discount) \successorFeatures(\state, \skill) - \skill}}
\end{equation}
The first term in {\color{perf}{red}} aims at maximizing the return, while the second term in {\color{dist}{blue}} aims at executing the desired skill. To optimize the actor to be high-performing while executing diverse skills, we use a generalized policy iteration method. The algorithm consists in (1) policy evaluation for both critics $\valueFunction(\state, \skill)$ and $\successorFeatures(\state, \skill)$, and (2) policy improvement via optimization of the Lagrangian function introduced in Equation~\ref{eq:actor-obj}. This formulation combines the classic policy improvement based on value function with a novel policy skill improvement based on successor features.

The Lagrange multiplier $\lagrange$ is optimized to balance the quality-diversity trade-off.
If ${\color{dist}\norm{(1-\discount) \successorFeatures(\state_1, \skill_1) - \skill_1}} \leq \threshold$ is satisfied for $(\state_1, \skill_1)$, we expect $\lagrange(\state_1, \skill_1)$ to decrease to encourage maximizing the return.
On the contrary, if the constraint is not satisfied for $(\state_2, \skill_2)$, we expect $\lagrange(\state_2, \skill_2)$ to increase to encourage satisfying the constraint.

\subsection{Practical Algorithm}
\label{sec:methods:practicalAlgorithm}
\begin{algorithm*}[t]
\caption{\ours}
\label{algo:ours}
\newcommand{\obs}{\state}
\renewcommand{\algorithmiccomment}[1]{\hfill$\triangleright$#1}
\newcommand{\LINECOMMENT}[1]{$\triangleright$#1}
\small

\begin{algorithmic}
\INPUT{Parameters $\theta_\policy$, $\theta_\valueFunction$, $\theta_\successorFeatures$, $\theta_\lagrange$} \COMMENT{ Initial parameters for the actor, critics and Lagrange multiplier}
\STATE{$\replay \leftarrow \emptyset$} \COMMENT{ Initialize an empty replay buffer}
\REPEAT
    \STATE{$\skillEnv \sim \uniform\left(\skillSpace\right)$} \COMMENT{ Sample skill uniformly from skill space}
    \FOR{$T$ steps}
        \STATE{\LINECOMMENT{ Environment steps}}
        \STATE{$\action_t \sim \policy(\action_t | \obs_t, \skill)$} \COMMENT{ Sample action from policy}
        \STATE{$\obs_{t+1} \sim p(\obs_{t+1} | \obs_t, \action_t, \skill)$} \COMMENT{ Sample transition from the environment}
        \STATE{$\replay \leftarrow \replay \cup \{(\obs_t, \action_t, \reward(\obs_t, \action_t), \feat(\obs_t, \action_t), \obs_{t+1}, \skill)\}$} \COMMENT{ Store transition in the replay buffer}

        \STATE{\LINECOMMENT{ Training steps}}
        \STATE{$\params_\lagrange \leftarrow \params_\lagrange - \alpha_\lagrange \nabla J_\lagrange(\params_\lagrange)$} \COMMENT{ Update Lagrange multiplier with Eq. \ref{eq:lagrange-obj}}
        \STATE{$\params_\valueFunction \leftarrow \params_\valueFunction - \alpha_\valueFunction \nabla J_\valueFunction(\params_\valueFunction)$} \COMMENT{ Policy evaluation for value function with Eq.~\ref{eq:lossV}}
        \STATE{$\params_\successorFeatures \leftarrow \params_\successorFeatures - \alpha_\successorFeatures \nabla J_\successorFeatures(\params_\successorFeatures)$} \COMMENT{ Policy evaluation for successor features with Eq.~\ref{eq:lossSF}}
        \STATE{$\params_\policy \leftarrow \params_\policy + \alpha_\policy \nabla J_\policy(\params_\policy)$} \COMMENT{ Policy improvement with Eq.~\ref{eq:actor-obj}}
    \ENDFOR
\UNTIL{convergence}
\end{algorithmic}
\end{algorithm*}

The objective in Equation~\ref{eq:actor-obj} can be optimized with any reinforcement learning algorithm that implements generalized policy iteration. We give two variants of our method, one variant named \ours, that is model-free and that builds on top of \sac, and one variant named \oursmb, that is model-based and that builds on top of \dreamer.
Additional details about \oursmb are provided in Appendix~\ref{appendix:model-based}. In this section, we detail the model-free variant.

\ours's model-free pseudocode is provided in Algorithm~\ref{algo:ours}. At each iteration, a skill $\skill$ is uniformly sampled for an episode of length $T$, during which the agent interacts with the environment following skill $\skill$ with $\pi(\cdot | \state, \skill)$. At each time step $t$, the transition is stored in a replay buffer $\replay$, augmented with the features $\feat(\state_t, \action_t)$ and with the current desired skill $\skill$.

Then, the Lagrange multiplier is updated to balance the quality-diversity trade-off. The parameters $\params_\lagrange$ are optimized so that $\lagrange(\state, \skill)$ increases when the actor is unable to execute the desired skill $\skill$, to put more weight on executing the skill. Conversely, the parameters $\params_\lagrange$ are optimized so that $\lagrange(\state, \skill)$ decreases when the actor is able to execute the desired skill $\skill$, to put more weight on maximizing the return.
The update of the Lagrange multiplier and its role in the actor objective are depicted in Figure~\ref{fig:lagrangian-function}.
In practice, we use a cross-entropy loss to optimize $\params_\lagrange$:
\begin{equation}
\label{eq:lagrange-obj}
\begin{split}
    &J_\lagrange(\params_\lagrange) = \expectSymbol{\subalign{&\state \sim \stateDist\\ & \skill \sim \mathcal{U}(\skillSpace)}}
    \begin{aligned}[t]
    [ & - (1 - y) \log\left( 1 - \lagrange\left(\state, \skill\right) \right) \\ 
    & - y \log\left( \lagrange\left(\state, \skill\right)\right)]
    \end{aligned} \\
    &\mbox{where}\: y=\begin{cases}
    0 &\mbox{if}\quad {\color{dist} \norm{ (1-\discount) \successorFeatures(\state, \skill) - \skill } } \leq \threshold\\
    1 \quad &\mbox{otherwise}
    \end{cases}
\end{split}
\end{equation}

Finally, the critics $\valueFunction$, $\successorFeatures$ and the actor $\policySkill$ are trained with a policy iteration step adapted from \sac and following Equation~\ref{eq:actor-obj}. The objective is optimized with stochastic gradient descent using a mini-batch of transitions sampled from the replay buffer.
To improve sample efficiency, the transitions from the mini-batch are duplicated with new random skills sampled uniformly in the skill space.
Additional information about \ours's training procedure are provided in Appendix~\ref{appendix:qdac-expanded-info}.

\section{Experiments}
\label{sec:experiments}
The goal of our experiments is twofold: (1) evaluate \ours's ability to learn high-performing and diverse skills based on a wide range of features, (2) evaluate \ours's ability to harness learned skills to solve downstream tasks.
\subsection{Tasks}
\label{sec:tasks}
\subsubsection{Learning Diverse High-Performing Skills}
We evaluate our method on a range of challenging continuous control tasks using the Google Brax~\citep{freeman_BraxDifferentiablePhysics_2021} physics engine.
We consider the three classic locomotion environments Walker, Ant and Humanoid that we combine with four different feature functions that we call \emph{feet contact}, \emph{velocity}, \emph{jump} and \emph{angle}. The first two features are traditional benchmark tasks that have been extensively studied in the Quality-Diversity and GCRL literature~\cite{cully_RobotsThatCan_2015,faldor_MAPElitesDescriptorConditionedGradients_2023,nilsson_PolicyGradientAssisted_2021,zhu_MapGoModelAssistedPolicy_2021,finn_ModelAgnosticMetaLearningFast_2017}, while the two last ones are challenging tasks that we introduce in this work.
In these locomotion tasks, the objective is to go forward as fast as possible while minimizing energy consumption.

\textit{Feet Contact} features indicate for each foot of the agent, if the foot is in contact or not with the ground, exactly as defined in \dcgme's original paper~\citep{faldor_MAPElitesDescriptorConditionedGradients_2023}. For example, if the Ant only touches the ground with its second foot at time step $t$, then $\feat(\state_t, \action_t) = \begin{bmatrix}0 & 1 & 0 & 0\end{bmatrix}^\intercal$. The diversity of feet contact found by such QD algorithms has been demonstrated to be very useful in downstream tasks such as damage recovery~\citep{cully_RobotsThatCan_2015}. The expected features correspond to the proportion of time each foot is in contact with the ground.

\textit{Velocity} features are two-dimensional vectors indicating the velocity of the agent in the $xy$-plane, $\feat(\state_t, \action_t) = \begin{bmatrix}v_x(t) & v_y(t)\end{bmatrix}^\intercal$. We evaluate on the velocity features to show that our method works on classic GCRL tasks. Moreover, the velocity features are interesting because satisfying a velocity that is negative on the $x$-axis is directly opposite to maximizing the forward velocity reward.

\textit{Jump} features are one-dimensional vectors indicating the height of the lowest foot. For example, if the left foot of the humanoid is 10 cm above the ground and if its right foot is 3.5 cm above the ground, then the features $\feat(\state_t, \action_t) = \begin{bmatrix}0.035\end{bmatrix}$. The skills derived from the jump features are also challenging to execute because to maintain an average $\skill = \frac{1}{T} \sum_{i=0}^{T-1} \feat_{t+i}$, the agent is forced to oscillate around that value $\skill$ because of gravity.

\textit{Angle} features are two-dimensional vectors indicating the angle $\alpha$ of the main body about the $z$-axis, $\feat(\state_t, \action_t) = \begin{bmatrix}\cos(\alpha) & \sin(\alpha)\end{bmatrix}^\intercal$. The goal of this task is to go as fast as possible in the $x$-direction while facing any directions, forcing the agent to sidestep or moonwalk.

\subsubsection{Harnessing Skills for Few-Shot Adaptation and Hierarchical Learning}
We evaluate our method on few-shot adaptation scenarios with four types of perturbation and on one hierarchical learning task. For each task, the reward is the same but the MDP's dynamics is perturbed. Additional details are available in Appendix~\ref{appendix:adaptation-tasks}.

In few-shot adaption tasks, no re-training is allowed and we evaluate the top-performing skills for each method while varying the perturbation to measure the robustness of the different algorithms, see Appendix~\ref{appendix:adaptation-tasks} for more details. \textit{Humanoid - Hurdles} requires the agent to jump over hurdles of varying heights. \textit{Humanoid - Motor Failure} requires the agent to adapt to different degrees of failure in the motor controlling its left knee. In \textit{Ant - Gravity}, the agent needs to adapt to different gravity conditions. Finally, \textit{Walker - Friction} requires the agent to adapt to varying levels of ground friction. Here, we evaluate the agent's ability to adjust its locomotion strategy to a new perturbed MDP.

In the hierarchical learning task, named \textit{Ant - Wall}, the agent is faced with navigating around a wall. A meta-controller is trained with Soft Actor-Critic (SAC) to maximize forward movement. Here, we evaluate the ability to use the diversity of skills discovered by \ours for hierarchical RL.

\subsection{Baselines}
We compare \ours with two families of methods that both balance a quality-diversity trade-off. The first family consists in evolutionary algorithms that maintain a diverse population of high-performing individuals whereas the second family uses a pure reinforcement learning formulation. Additionally, we perform three ablation studies.

\paragraph{Quality-Diversity via Evolutionary Algorithms}  We compare our method with \ppga~\citep{batra_ProximalPolicyGradient_2023}, \dcgme~\citep{faldor_MAPElitesDescriptorConditionedGradients_2023,faldor_Synergizing_2023} and \qdpg~\cite{pierrot_DiversityPolicyGradient_2022}, three evolutionary algorithms that optimize a diverse population of high-performing individuals.
\ppga is a state-of-the-art Quality-Diversity algorithm that mixes Proximal Policy Optimization (PPO)~\citep{schulman_ProximalPolicyOptimization_2017} with \cmamaega~\citep{fontaine2023covariance}; it alternates between (1) estimating the performance-feature gradients with PPO and (2) maintaining a population of coefficients to linearly combine those
performance-feature gradients, those coefficients are optimized to maximize archive improvement.
\dcgme is another state-of-the-art Quality-Diversity algorithm that evolves a population of both high-performing and diverse solutions, and simultaneously distills those solutions into a single skill-conditioned policy. 
\qdpg is a Quality-Diversity algorithm that uses quality and diversity policy gradients to optimize its population of policies.

\paragraph{Quality-Diversity via Reinforcement Learning} We also compare our method with \domino~\citep{zahavy_DiscoveringPoliciesDOMiNO_2022}, \smerl~\citep{kumar_OneSolutionNot_2020} and \smerlReverse~\citep{zahavy_DiscoveringPoliciesDOMiNO_2022} that balance a quality-diversity trade-off using a pure reinforcement learning formulation. \domino is a reinforcement learning algorithm designed to discover diverse behaviors while preserving near-optimality. Analogous to our method, it characterizes policies' behaviors using successor features. \smerl learns a latent-conditioned policy that maximizes the mutual information between states $\state$ and latent variables $z$, with a threshold to toggle the diversity reward in the objective $\reward + \alpha \mathbbm{1}(R \geq R^* - \epsilon) \tilde{\reward}$. The diversity reward $\tilde{\reward}$ is measured from the likelihood of a discriminator $q(z | \state)$ coming from \diayn. In other words, \smerl maximizes a weighted combination of environment reward and diversity reward when the policy is near-optimal, and only the environment reward otherwise. \smerlReverse maximizes a similar reward $\mathbbm{1}(R < R^* - \epsilon) \reward + \alpha \tilde{\reward}$. In other words, \smerlReverse maximizes a weighted combination of environment reward and diversity reward when the policy is not near-optimal, and only the diversity reward otherwise.

\paragraph{Ablations} We perform three additional ablation studies, that we call \oursSepSkill, \oursFixedLambda and \uvfa.
For \oursSepSkill, we remove the successor features representation and use a naive distance to skill instead $\sum_{t}\discount^t\norm{\feat_t - \skill}$ to understand the contribution of the successor features critic to optimize diversity.
For \oursFixedLambda, we remove the Lagrange multiplier and use a fixed trade-off instead to understand the contribution of constrained optimization.
\uvfa~\citep{schaul_UniversalValueFunction_2015} is an algorithm that corresponds to the combination of our two previous ablations, as such we consider it to be an ablation in this work.
A summarized description of all baselines under study is provided in Table~\ref{table:baselines}.

\subsection{Evaluation Metrics}
\begin{figure}[t]
\centering
\includegraphics[width=0.48\textwidth]{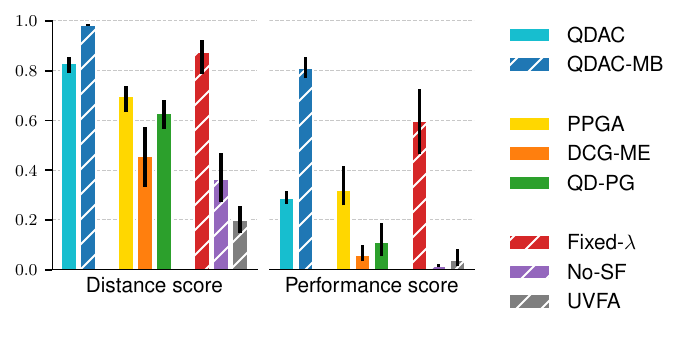}

\caption{Distance and performance scores normalized and aggregated across all tasks. The values correspond to the IQM while the error bars represent IQM 95\% CI.}
\label{fig:scores}
\end{figure}

\label{sec:experiments-metrics}
We evaluate our method using two types of metrics from the Quality-Diversity literature that aim at evaluating the performance and the diversity of the discovered skills: (1) the distance to skill metrics, that evaluate the ability of an agent to execute desired skills, and (2) the performance metrics, that quantify the ability of an agent to maximize return while executing desired skills.
Each experiment is replicated 10 times with random seeds. We report the Inter-Quartile Mean (IQM) value for each metric, with the estimated 95\% Confidence Interval (CI)~\citep{agarwal_DeepReinforcementLearning_2021}.
The statistical significance of the results is evaluated using the Mann-Whitney $U$ test~\citep{mann1947test} and the probabilities of improvement are reported in Appendix~\ref{appendix:supplementary-results:quantitative}.

\begin{figure*}[t]
\centering
\includegraphics[width=0.99\textwidth]{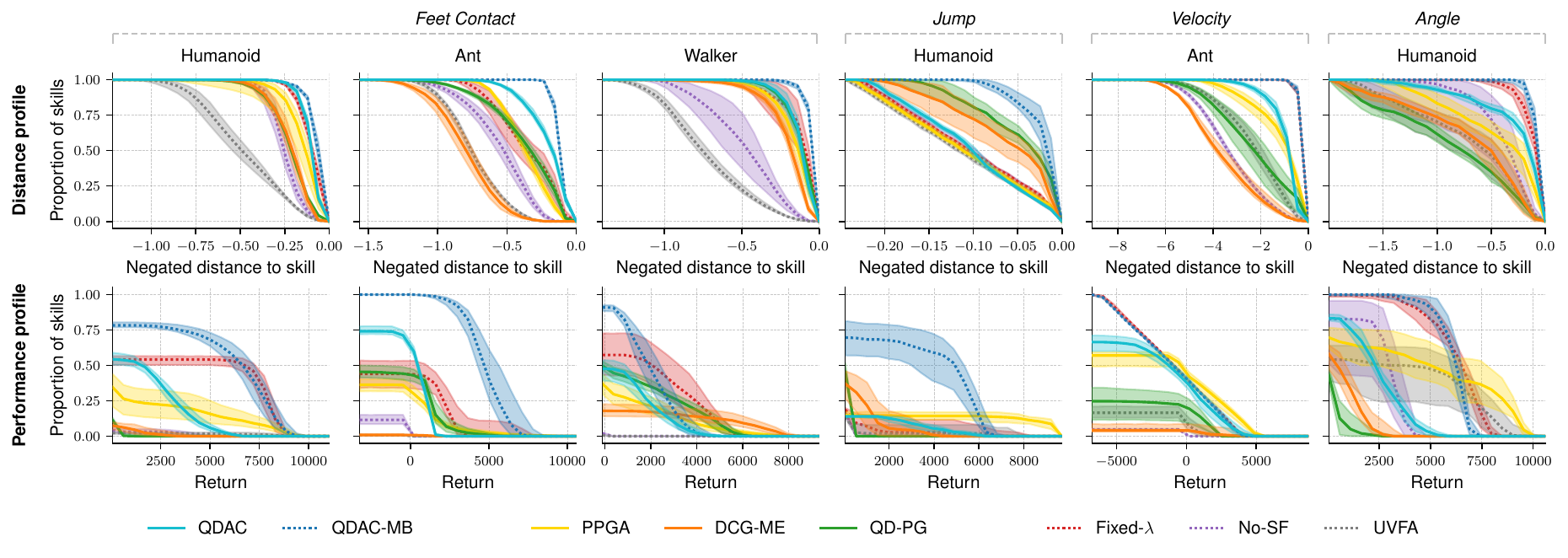}

\caption{%
(\textbf{top}) Distance profiles and (\textbf{bottom}) performance profiles for each task defined in Section~\ref{sec:experiments-metrics}.
The lines represent the IQM for 10 replications, and the shaded areas correspond to the 95\% CI.
Figure~\ref{fig:profiles_explanation} illustrates how to read distance and performance profiles.
}
\label{fig:profiles}
\end{figure*}

\paragraph{Distance to skill metrics}
To evaluate the ability of a policy to achieve a given skill $\skill$, we estimate the \emph{expected distance to skill}, denoted $d(\skill)$, by averaging the euclidean distance between the desired skill $\skill$ and the observed skill over 10 rollouts, as defined by~\citet{faldor_MAPElitesDescriptorConditionedGradients_2023,faldor_Synergizing_2023}.
First, we use $d(\skill)$ to compute \emph{distance profiles} on Figure~\ref{fig:profiles}, which quantify for a given distance $d$, the proportion of skills in the skill space that have an expected distance to skill smaller than $d$, computed with the function $d \mapsto \frac{1}{N_\skill}\sum_{i=1}^{N_\skill} \mathbbm{1}(d(\skill_i) < d)$.
Second, we summarize the ability of a policy to execute skills with the \textit{distance score}, $\frac{1}{N_\skill}\sum_{i=1}^{N_\skill} -d(\skill_i)$.

\paragraph{Performance metrics}

To evaluate the ability of a policy to solve a task given a skill $\skill$, we estimate the \emph{expected undiscounted return}, denoted $R(\skill)$, by averaging the return over 10 rollouts, as defined by~\citet{flageat_UncertainQualityDiversityEvaluation_2023,grillotti_DonBetLuck_2023}.
First, we use $R(\skill)$ to compute \emph{performance profiles} on Figure~\ref{fig:profiles}, which quantify for a given return $R$, the proportion of skills in the skill space that have an expected return larger than $R$, after filtering out the skills that are not achieved by the policy. To this end, we compute the expected distance to skill $d(\skill)$, and discard skills with an expected distance to skill that is larger than a predefined threshold, $d(\skill) > d_\text{eval}$. More precisely, the performance profile is the function $R \mapsto \frac{1}{N_\skill}\sum_{i=1}^{N_\skill} \mathbbm{1}(d(\skill_i) < d_\text{eval}\text{, }R(\skill_i) > R)$.
Second, we summarize the ability of a policy to maximize return while executing skills, with the \textit{performance score}, $\frac{1}{N_\skill}\sum_{i=1}^{N_\skill} R(\skill) \mathbbm{1}(d(\skill_i) < d_\text{eval})$.

\subsection{Results}
The goal of our experiments is to answer two questions: (1) Does \ours solve the Quality-Diversity problem? (2) Can we harness the high-performing and diverse skills to adapt to perturbed MDP?
In Section~\ref{sec:results-qd}, we demonstrate that \ours achieves significantly higher performance and more diverse behaviors on six challenging continuous control locomotion tasks (Fig.~\ref{fig:scores}).
In Section~\ref{sec:results-adaptation}, we show that we can harness the learned skills to adapt better than other baselines to five perturbed MDP (Fig.~\ref{fig:adaptation}).
The code is available at: \href{https://github.com/adaptive-intelligent-robotics/QDAC}{github.com/adaptive-intelligent-robotics/QDAC}.

\subsubsection{Learning Diverse High-Performing Skills}
\label{sec:results-qd}
\begin{figure*}[t]
\centering
\includegraphics[width=0.99\textwidth]{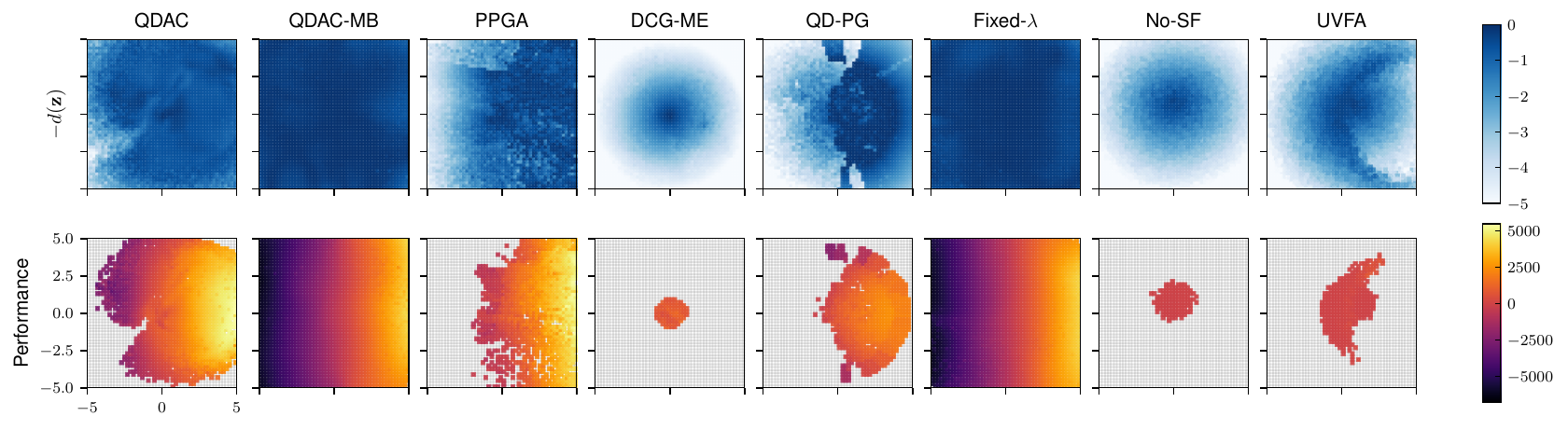}

\caption{
\textbf{Ant Velocity} Heatmaps of \textbf{(top)} negative distance to skill, \textbf{(bottom)} performance defined in Section~\ref{sec:experiments-metrics}.
The heatmap represents the skill space $\skillSpace = [-5 \text{ m/s}, 5 \text{ m/s}]^2$, of target velocities. This space is discretized into cells, with each cell representing a distinct skill $\skill = \begin{bmatrix}v_x & v_y\end{bmatrix}^\intercal$.
In the bottom row, empty cells show which skills are not successfully executed (i.e. $d(\skill) > d_{\mathrm{eval}}$).
The heatmaps for other tasks are presented in section~\ref{appendix:heatmaps}.
}
\label{fig:heatmap-ant-velocity}
\end{figure*}

In this section, we evaluate \ours with metrics coming from the Quality-Diversity literature, namely the distance to skill and performance metrics. However, \domino and \smerl optimize for diversity, without the focus on executing specific skills. Consequently, the concept of `distance to skill' does not apply and thus, the traditional QD metrics are not applicable. Nonetheless, we compare our approach with these baselines on adaptation tasks, for which they were initially designed, in Section~\ref{sec:results-adaptation}.

\ours and \oursmb outperform all baselines in executing a wide range of skills (Fig.~\ref{fig:profiles}), except on Humanoid Jump where \dcgme and \qdpg achieve a better distance profile than our model-free variant.
Yet, \oursmb outperforms those baselines, due to the representation capabilities of the world model. The jump features are challenging because of the $\min$ operator, and because the features are not explicitly available in the observations given to the agent.

Notably, \ours and \oursmb are capable of achieving skills that are contrary to the task reward, as illustrated by the velocity features in Figure~\ref{fig:profiles} and Figure~\ref{fig:heatmap-ant-velocity}, which is not the case for \ppga, \dcgme and \qdpg.
Finally, our approach outperforms \dcgme that fails to explicitly minimize the expected distance to skill, a common issue among QD algorithms~\citep{flageat_UncertainQualityDiversityEvaluation_2023}.

\oursmb outperforms \oursFixedLambda on all tasks, showing the importance of using constrained optimization to solve the QD problem.
Furthermore, \oursmb achieves better performance than \oursSepSkill on all tasks, showing the importance of the successor features critics to optimize diversity.
Finally, \ours significantly outperforms \oursSepSkill and \uvfa on all tasks. On feet contact tasks, \oursSepSkill and \uvfa can only execute skills in the corners of the skill space where $\skill = \feat_t$, as shown in Figure~\ref{fig:appendix:humanoid-feet-contact}. This is because these baselines employ a naive approach that consists in minimizing the distance between the features and the desired skill. Thus, they can only execute skills where the legs are always or never in contact with the ground.
These comparisons supports the claim that \ours is capable of accurately executing a diversity of skills and highlight the significance of the policy skill improvement term in blue in Equation~\ref{eq:actor-obj}.

\ours and \oursmb outperform \dcgme and \qdpg in maximizing return (Fig.~\ref{fig:scores}), as the latter don't achieve many skills in the first place, and the performance score only evaluates the performance of skills successfully executed by the policy.
While \ppga achieves a performance score comparable to \ours, it does so by finding fewer robust policies, albeit with better performance (Fig.~\ref{fig:profiles}).
\oursFixedLambda is the only baseline that gets performance scores and profiles comparable to \oursmb.
However, \oursFixedLambda covers a smaller range of skills, as evidenced by the edges of the skill space on Figure~\ref{fig:appendix:humanoid-feet-contact}.
Additionally, \oursmb outperforms \oursFixedLambda on the challenging jump task (Fig.~\ref{fig:profiles}), due to the necessity of a strong weight on the constraint.
Ultimately, using an adaptive $\lagrange$ proves advantageous for our approach.
%

As summarized in Figure~\ref{fig:scores}, \ours and \oursmb achieve a better quality-diversity trade-off than other baselines, quantified by the distance score and the performance score.

\subsubsection{Harnessing Skills for Few-Shot Adaptation and Hierarchical Learning}
\label{sec:results-adaptation}
\begin{figure*}[t]
\centering
\includegraphics[width=0.99\textwidth]{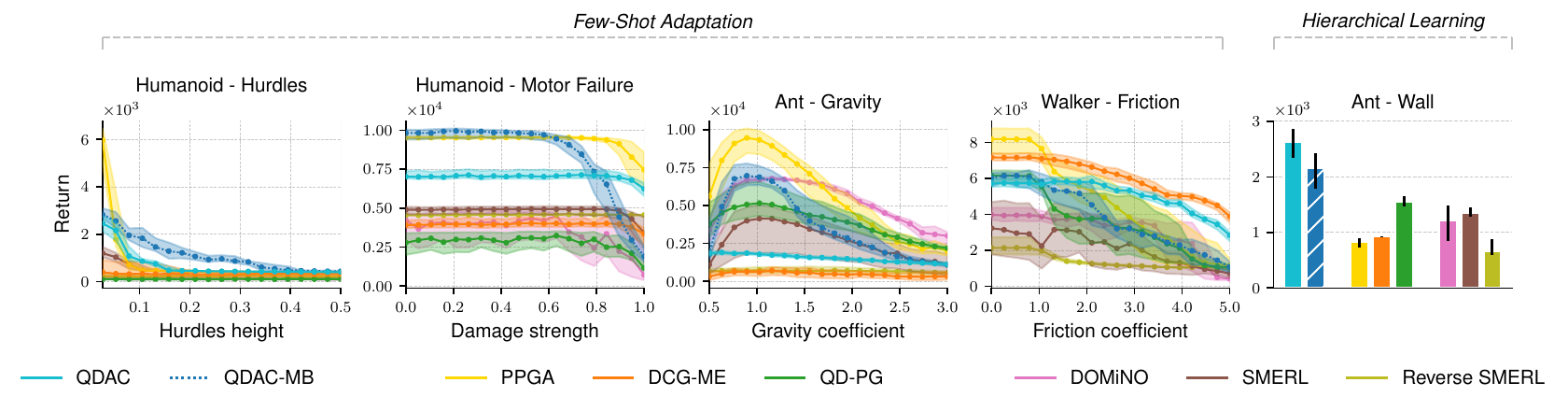}

\caption{Performance for each algorithm in environments with different levels of perturbations after few-shot adaptation or after hierarchical learning.
The lines represent the IQM for 10 replications, and the shaded areas correspond to the 95\% CI.}
\label{fig:adaptation}
\end{figure*}

\ours and \oursmb demonstrate competitive performance in few-shot adaptation and hierarchical RL tasks, see Figure~\ref{fig:adaptation}.
On the hurdles task, when considering hurdle heights strictly greater than 0, \oursmb significantly outperforms other baselines by consistently jumping over higher hurdles and showcases remarkable behaviors.
On the motor failure task, although performing worse than \ppga, \ours shows great robustness, especially in the high damage regime. \oursmb performs better than \ours on low damage, but \ours can adapt to 100\% damage strength on the left knee, still achieving more than 5,000 in return. 
In other words, \ours has found a way to continue walking despite not being able to control at all the left knee.
\ours does not seem able to adapt to gravity variations, but \oursmb shows competitive performance although performing slightly worse than \ppga and \domino.
On Walker - Friction, \ours outperforms all baselines except \ppga and \dcgme that achieve marginally better performance.
Finally, \ours's learned skills appear to be the best on the hierarchical RL task, as it achieves significantly higher performance than other baselines.

Our extensive experiments and analyses in Sections~\ref{sec:results-qd} and \ref{sec:results-adaptation} firmly establish the efficacy of \ours and \oursmb in addressing the dual challenges of learning high-performing and diverse skills. These methods not only surpass traditional Quality-Diversity algorithms in optimizing for specific pre-defined skills but also demonstrate remarkable adaptability and robustness in perturbed environments.

\section{Related Work}
\label{sec:related-work}
QD optimization~\citep{pugh_QualityDiversityNew_2016,cully_QualityDiversityOptimization_2018} is a family of algorithms that generate large collections of solutions, such as policies, that are both diverse and high-performing.
Those methods originate from Novelty Search~\cite{lehman2011NoveltySearch,lehman2011evolving} and Evolutionary Algorithms literature, where the diversity is defined across a population of solutions.
Quality-Diversity algorithms have been shown to be competitive with skill discovery reinforcement learning methods~\citep{chalumeau_NeuroevolutionCompetitiveAlternative_2022}, and promising for adaptation to unforeseen situations~\citep{cully_RobotsThatCan_2015}.
When considering large neural network policies, the sample efficiency of QD algorithms can be improved by using Evolution Strategies~\citep{fontaine_CovarianceMatrixAdaptation_2020, colas2020mapelitesES}, RL-based methods~\citep{pierrot_DiversityPolicyGradient_2022,nilsson_PolicyGradientAssisted_2021,faldor_MAPElitesDescriptorConditionedGradients_2023,tjanaka_ApproximatingGradientsDifferentiable_2022,batra_ProximalPolicyGradient_2023, xue2024sampleefficient}. The sample efficiency can be further improved by decomposing the policies into several parts and coevolving a sub-population for each part~\citep{xue2024sampleefficient}.
However, most QD algorithms output a large number of policies, which can be difficult to deal with in downstream tasks.
Similarly to \ours, \dcgme addresses that issue by optimizing a single skill-conditioned policy~\citep{faldor_MAPElitesDescriptorConditionedGradients_2023}.
Finally, approaches like \smerl~\cite{kumar_OneSolutionNot_2020} or \domino~\cite{zahavy_DiscoveringPoliciesDOMiNO_2022} also solve the QD objective employing a pure reinforcement learning formulation. Contrary to \ours, the policies discovered by \domino are not trained to execute specific target skills.

Most Unsupervised Reinforcement Learning approaches discover diverse behaviors by maximizing an intrinsic reward defined in terms of the discriminability of the trajectories.
Usually, the methods maximize the Mutual Information (MI) between the trajectories and the skills $I(\tau, \skill)$~\citep{gregor_VariationalIntrinsicControl_2016,sharma_DynamicsAwareUnsupervisedDiscovery_2019,mazzaglia_ChoreographerLearningAdapting_2022}, simplified to an MI-maximization between skills and states with the following lower bound: $I(\tau, \skill) \geq \sum_{t=1}^{T} I(s_t, \skill)$.
It has been shown that MI-based algorithms are equivalent to distance-to-skill minimization algorithms~\citep{choi_VariationalEmpowermentRepresentation_2021,gu_BraxlinesFastInteractive_2021}, and therefore present similarities with our work.
However, most URL algorithms maximize an intrinsic reward while disregarding any other objective, making it difficult to discover useful and expressive behaviors.

While diversity can be achieved by maximizing a mutual information objective, it can also be explicitly defined as a distance between behavioral descriptors.
Such descriptors can take the form of successor features \citep{zahavy_DiscoveringPoliciesDOMiNO_2022, zahavy_DiversifyingAICreative_2023} or of expected features obtained though entire episodes \citep{cully_RobotsThatCan_2015, batra_ProximalPolicyGradient_2023}. 
In this work, we rely on this latter definition, as expressed in Problems~\ref{eq:problem-1} and~\ref{eq:problem-2}.
%
%
The features $\feat$ can be defined in different ways.
First, they can be a subpart of the state of the agent such as the joint positions and velocities \citep{zahavy_DiscoveringPoliciesDOMiNO_2022}, torso velocity \citep{cheng2023learning}, or feet contacts \citep{cully_RobotsThatCan_2015}.
In this case, the state of the agent may guide the search towards relevant notions of diversity; however, this requires expert knowledge about the task, and the choice of feature definition strongly influences the quality and diversity of the generated solutions~\citep{InfluenceEncodings2016}.
Second, to avoid hand-defining features, we could define $\feat$ as the full state \citep{kumar_OneSolutionNot_2020} or as an unsupervised low-dimensional encoding of it \citep{grillotti_UnsupervisedBehaviourDiscovery_2022, mazzaglia_ChoreographerLearningAdapting_2022, liu2021behavior}.
In this case, additional techniques can be used to ensure the learned behaviors are relevant, such as adding an extrinsic reward term \citep{chalumeau_NeuroevolutionCompetitiveAlternative_2022}, promoting diversity in the neighborhood of relevant solutions \citep{grillotti_RelevanceguidedUnsupervisedDiscovery_2022}, or adding constraints for near-optimality \citep{zahavy_DiscoveringPoliciesDOMiNO_2022, kumar_OneSolutionNot_2020}.
Instead, \ours constrains the agent’s behavior to follow a given hand-defined skill $\skill$, and maximizes the performance for all skills $\skill \in \skillSpace$.

\section{Conclusion}
\label{sec:conclusion}
In this work, we present \ours, an actor-critic deep reinforcement learning algorithm, that leverages a value function critic and a successor features critic to learn high-performing and diverse behaviors. In this framework, the actor optimizes an objective that seamlessly unifies both critics using constrained optimization to (1) maximize return, while (2) executing diverse skills.

Empirical evaluations suggest that \ours outperforms previous Quality-Diversity methods on six continuous control locomotion tasks. Quantitative results demonstrate that \ours is competitive in adaptation tasks, while qualitative analyses reveal a range of diverse and remarkable behaviors.

In the future, we hope to apply \ours to other tasks with different properties than the tasks from this paper. 
For example, it would be interesting to apply \ours to non-ergodic environments such as Atari games.

Furthermore, like the vast majority of Quality-Diversity algorithms, \ours uses a manually defined diversity measure to guide the diversity search towards relevant behaviors. 
An exciting direction for future work would be to learn the feature function in an unsupervised manner to discover task-agnostic skills.

\section*{Impact Statement}
This paper presents work whose goal is to advance the field of Machine Learning. There are many potential societal consequences of our work, none which we feel must be specifically highlighted here.

\bibliography{bibliography}
\bibliographystyle{icml2024}

\newpage

\renewcommand\thefigure{\thesection.\arabic{figure}}
\renewcommand\thetable{\thesection.\arabic{table}}

\renewcommand{\MyTag}{appendix:}
\appendix
\onecolumn
\newpage
\section{Supplementary Results}
\label{appendix:supplementary-results}
\subsection{Quantitative Results}
\label{appendix:supplementary-results:quantitative}
\begin{figure}[H]
\centering
\includegraphics[width=0.5\textwidth]{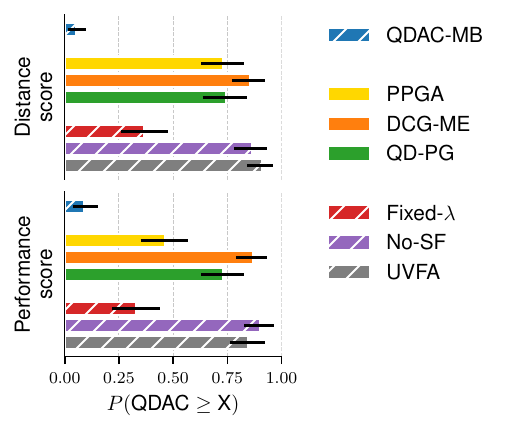}
\caption{%
Probabilities of improvement of \ours over all other baselines, aggregated across all tasks, as defined by~\citet{agarwal_DeepReinforcementLearning_2021}.
}
\end{figure}

\begin{figure}[H]
\centering
\includegraphics[width=0.5\textwidth]{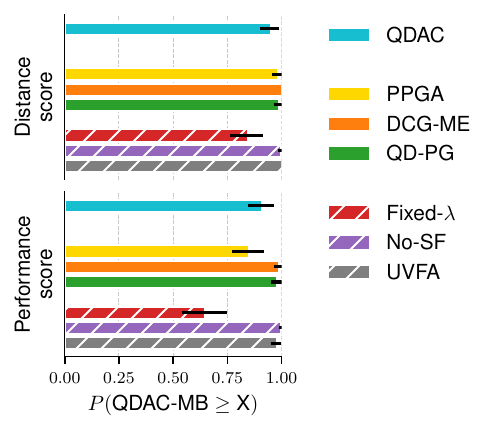}
\caption{%
Probabilities of improvement of \oursmb over all other baselines, aggregated across all tasks, as defined by~\citet{agarwal_DeepReinforcementLearning_2021}.
}
\end{figure}

\begin{figure}[H]
\centering
\includegraphics[width=\textwidth]{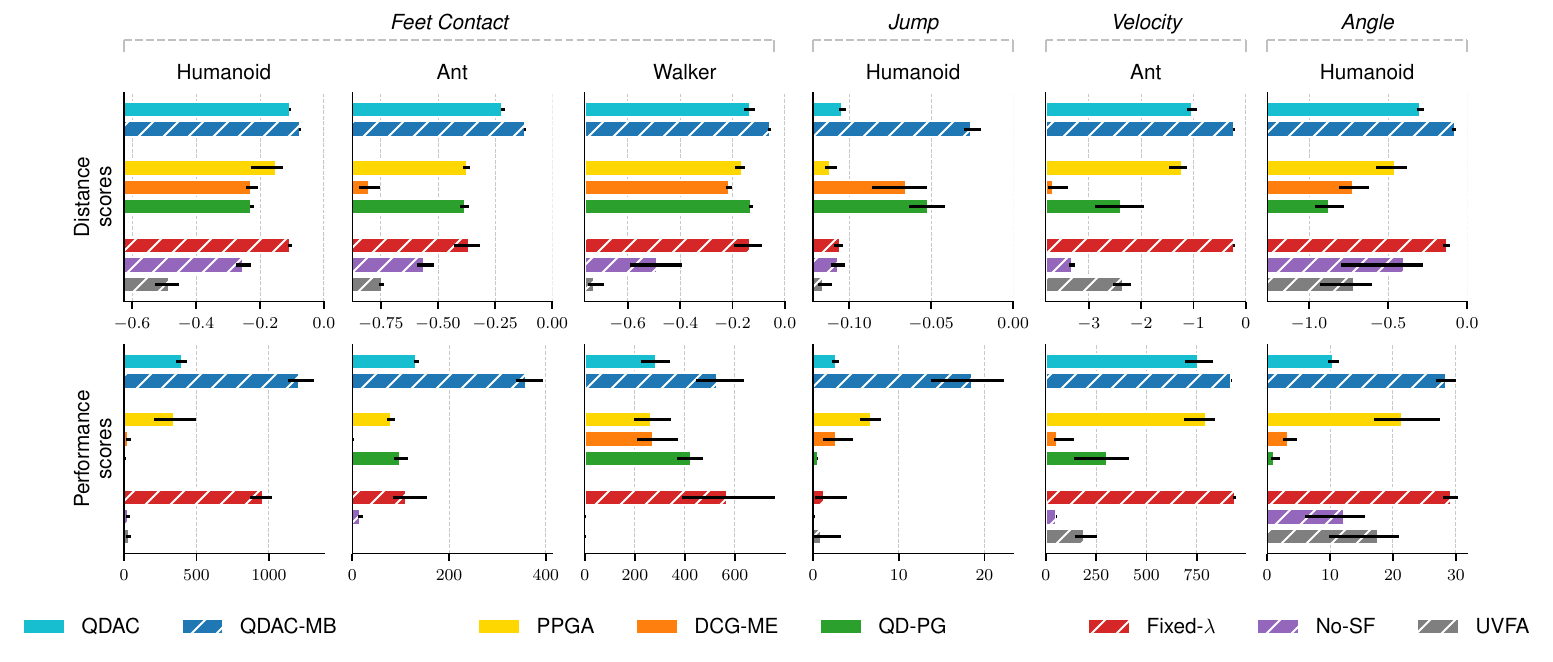}
\caption{%
IQM for distance and performance scores per task.
}
\end{figure}

\begin{figure}[H]
\centering
\includegraphics[width=\textwidth]{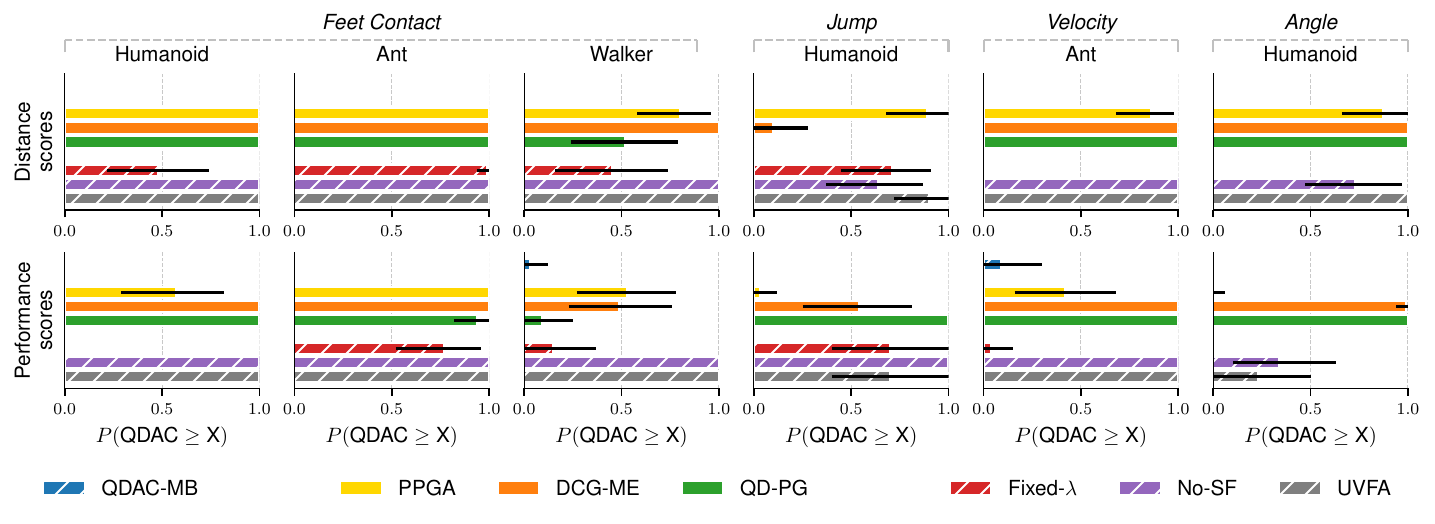}
\caption{%
Per-task probabilities of improvement (as defined by~\citet{agarwal_DeepReinforcementLearning_2021}) of \ours over all other baselines.}
\end{figure}

\begin{figure}[H]
\centering
\includegraphics[width=\textwidth]{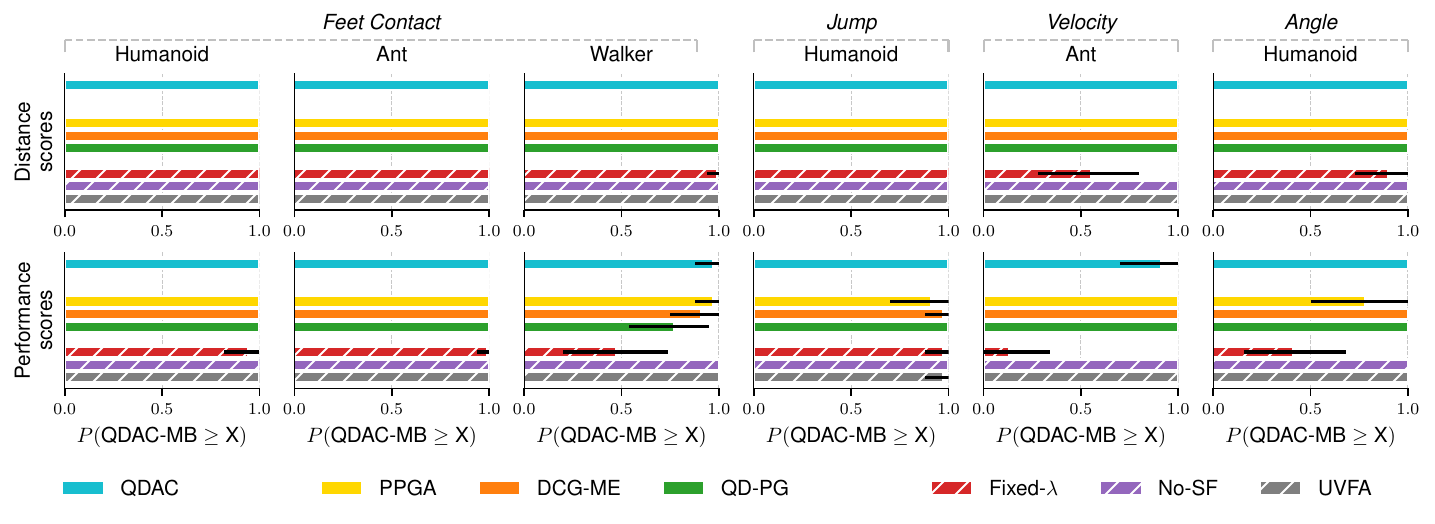}
\caption{%
Per-task probabilities of improvement (as defined by~\citet{agarwal_DeepReinforcementLearning_2021}) of \oursmb over all other baselines.}
\end{figure}

\newpage
\subsection{Heatmaps}

\label{appendix:heatmaps}

\begin{figure}[H]
\centering
\includegraphics[width=\textwidth]{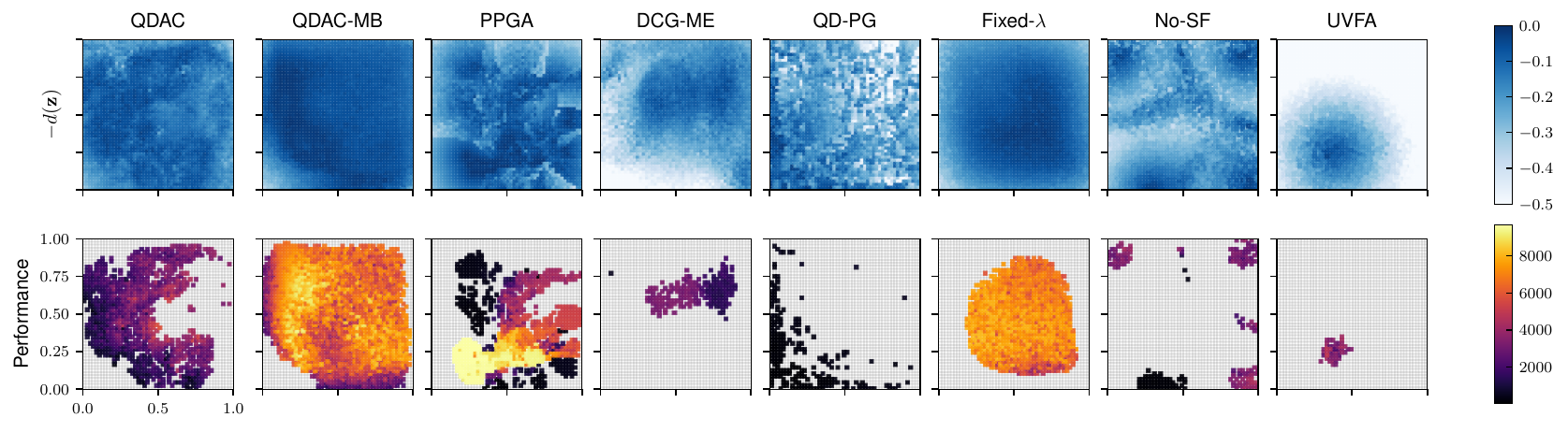}
\caption{%
\textbf{Humanoid Feet Contact} Heatmaps of \textbf{(top)} negative distance to skill, \textbf{(bottom)} performance defined in Section~\ref{sec:experiments-metrics}.
The heatmap represents the skill space of feet contacts $\skillSpace = [ 0, 1]^2$. 
This space is discretized into cells, with each cell representing a distinct skill $\skill = \begin{bmatrix}z_1 & z_2\end{bmatrix}^\intercal$, where $z_i$ is the proportion of time that leg $i$ touches the ground over an entire episode.
In the bottom row, empty cells show which skills are not successfully executed (i.e. $d(\skill) > d_{\mathrm{eval}}$), while colorized cells indicate the performance score obtained for the corresponding skill.
}
\label{fig:appendix:humanoid-feet-contact}
\end{figure}

\begin{figure}[H]
\centering
\includegraphics[width=\textwidth]{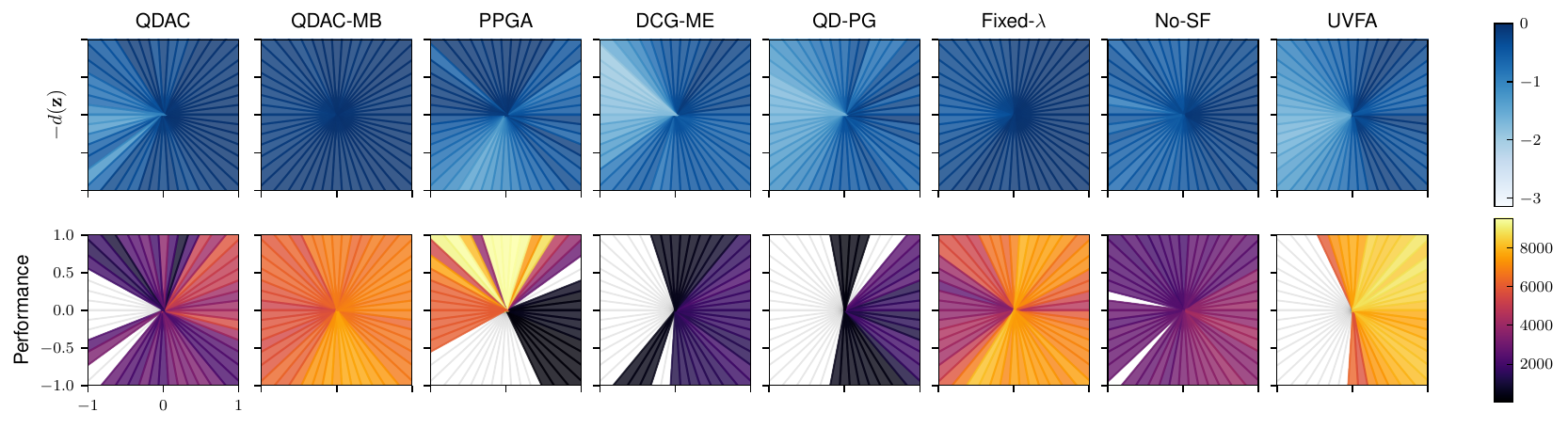}
\caption{%
\textbf{Humanoid Angle} Heatmaps of \textbf{(top)} negative distance to skill, \textbf{(bottom)} performance defined in Section~\ref{sec:experiments-metrics}.
The heatmap represents the skill space of body angles $\skillSpace = ]-\pi, \pi]$. 
This space is discretized into cells, with each cell representing a distinct skill; in this task, the skills refer to the angle of the humanoid body about the $z$-axis.
In the bottom row, empty cells show which skills are not successfully executed (i.e. $d(\skill) > d_{\mathrm{eval}}$), while colorized cells indicate the performance score obtained for the corresponding skill.
}
\label{fig:appendix:humanoid-angle}
\end{figure}

In Figures~\ref{fig:appendix:humanoid-feet-contact} to~\ref{fig:appendix:humanoid-angle}, we report the heatmaps for the metrics defined in Section~\ref{sec:experiments-metrics}: the negative distance to skill (in the top row) and the performance (in the bottom row). 
Each heatmap represents the skill space of the corresponding task.
In the first row, the color of each cell represents the negated distance to the closest skill achieved by the policy (the darker the better).
In the bottom row, empty cells show which skills are not successfully executed (i.e. $d(\skill) > d_{\mathrm{eval}}$); while colorized cells indicate the performance score (i.e. the return) achieved by the agent for the corresponding skill.
\begin{figure}[H]
\centering
\includegraphics[width=\textwidth]{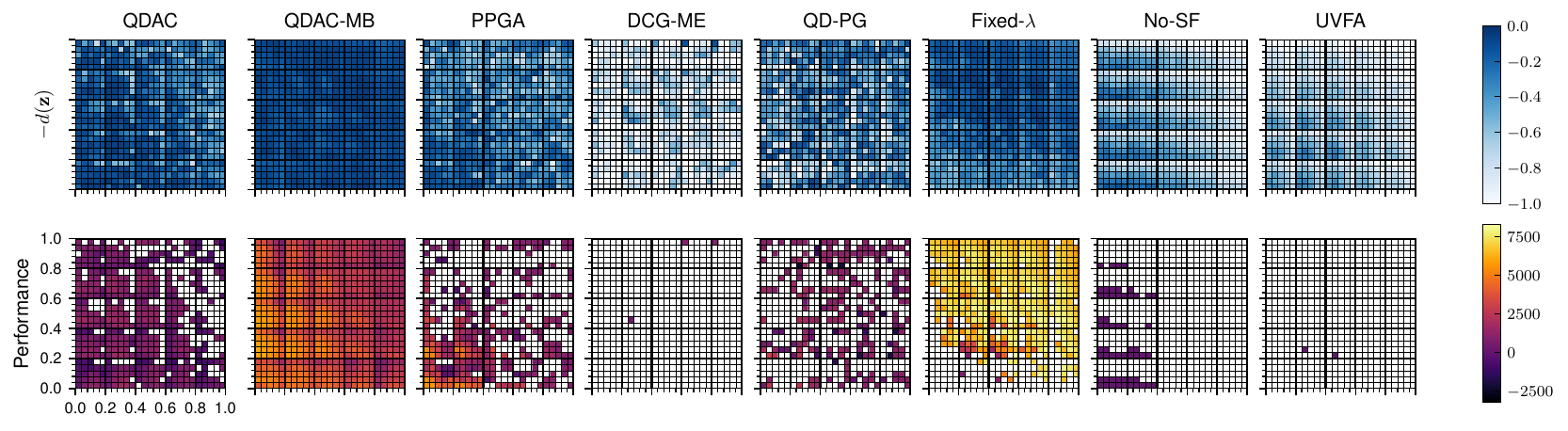}
\caption{%
\textbf{Ant Feet Contact} Heatmaps of \textbf{(top)} negative distance to skill, \textbf{(bottom)} performance defined in Section~\ref{sec:experiments-metrics}.
The heatmap represents the skill space of feet contacts $\skillSpace = [ 0, 1]^4$. 
This space is discretized into cells, with each cell representing a distinct skill $\skill = \begin{bmatrix}z_1 & z_2 & z_3 & z_4\end{bmatrix}^\intercal$, where $z_i$ is the proportion of time that leg $i$ touches the ground over an entire episode.
In the bottom row, empty cells show which skills are not successfully executed (i.e. $d(\skill) > d_{\mathrm{eval}}$), while colorized cells indicate the performance score obtained for the corresponding skill.
}
\end{figure}
\begin{figure}[H]
\centering
\includegraphics[width=\textwidth]{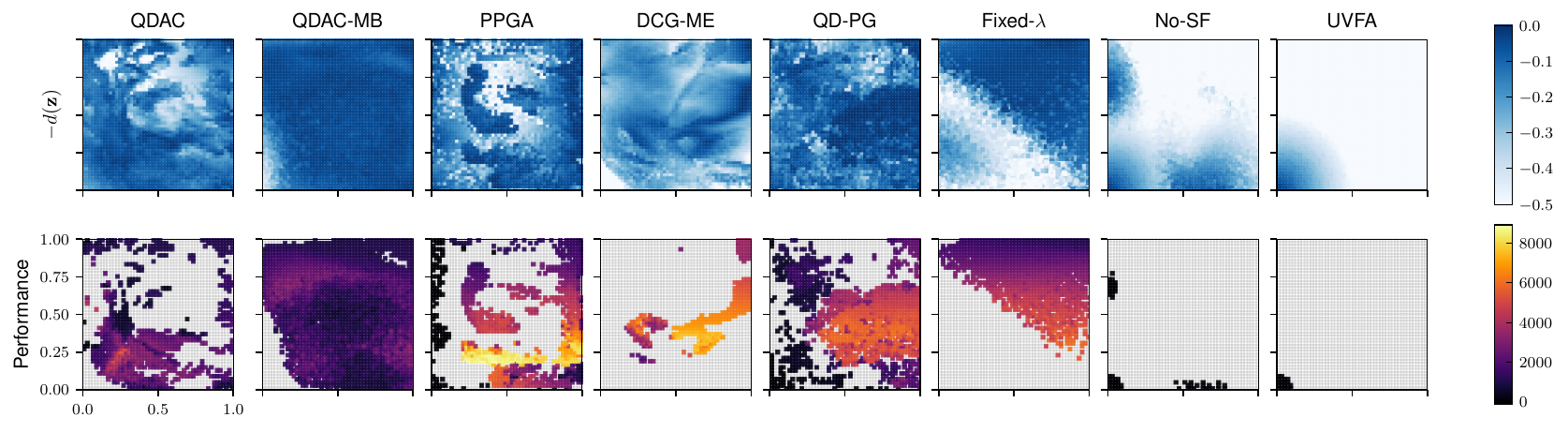}
\caption{%
\textbf{Walker Feet Contact} Heatmaps of \textbf{(top)} negative distance to skill, \textbf{(bottom)} performance defined in Section~\ref{sec:experiments-metrics}.
The heatmap represents the skill space of feet contacts $\skillSpace = [ 0, 1]^2$. 
This space is discretized into cells, with each cell representing a distinct skill $\skill = \begin{bmatrix}z_1 & z_2\end{bmatrix}^\intercal$, where $z_i$ is the proportion of time that leg $i$ touches the ground over an entire episode.
In the bottom row, empty cells show which skills are not successfully executed (i.e. $d(\skill) > d_{\mathrm{eval}}$), while colorized cells indicate the performance score obtained for the corresponding skill.
}
\end{figure}
\begin{figure}[H]
\centering
\includegraphics[width=\textwidth]{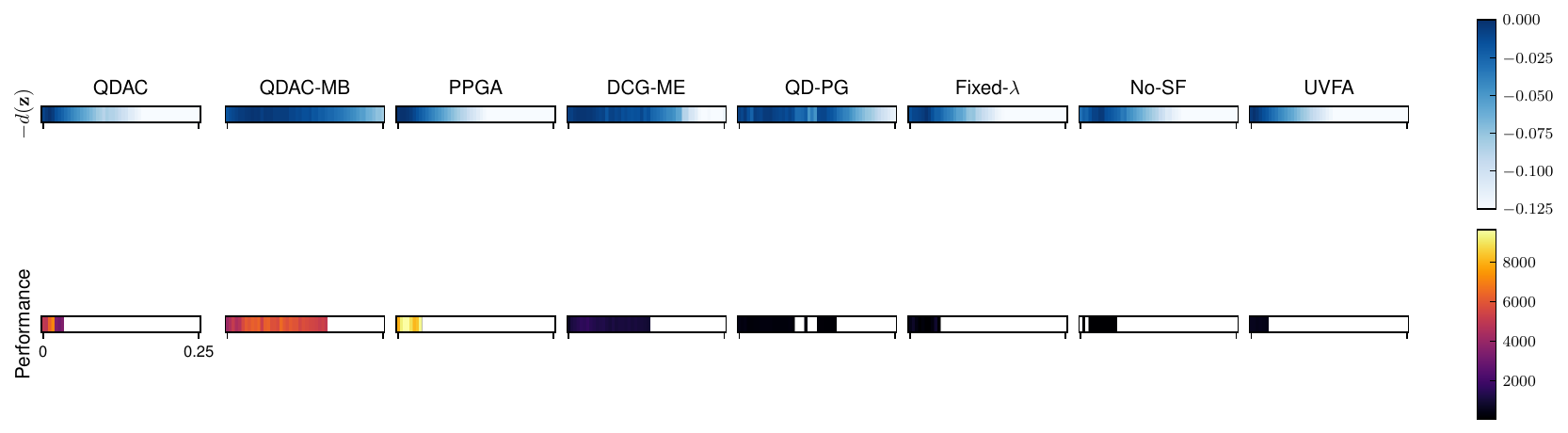}
\caption{%
\textbf{Humanoid Jump} Heatmaps of \textbf{(top)} negative distance to skill, \textbf{(bottom)} performance defined in Section~\ref{sec:experiments-metrics}.
The heatmap represents the skill space of jumping skills $\skillSpace = [0, 0.25]$. 
This space is discretized into cells, with each cell representing a distinct skill; in this task, the skills refer to the average of the lowest foot heights over an entire episode.
In the bottom row, empty cells show which skills are not successfully executed (i.e. $d(\skill) > d_{\mathrm{eval}}$), while colorized cells indicate the performance score obtained for the corresponding skill.
}
\end{figure}
\begin{figure}[H]
\centering
\includegraphics[width=\textwidth]{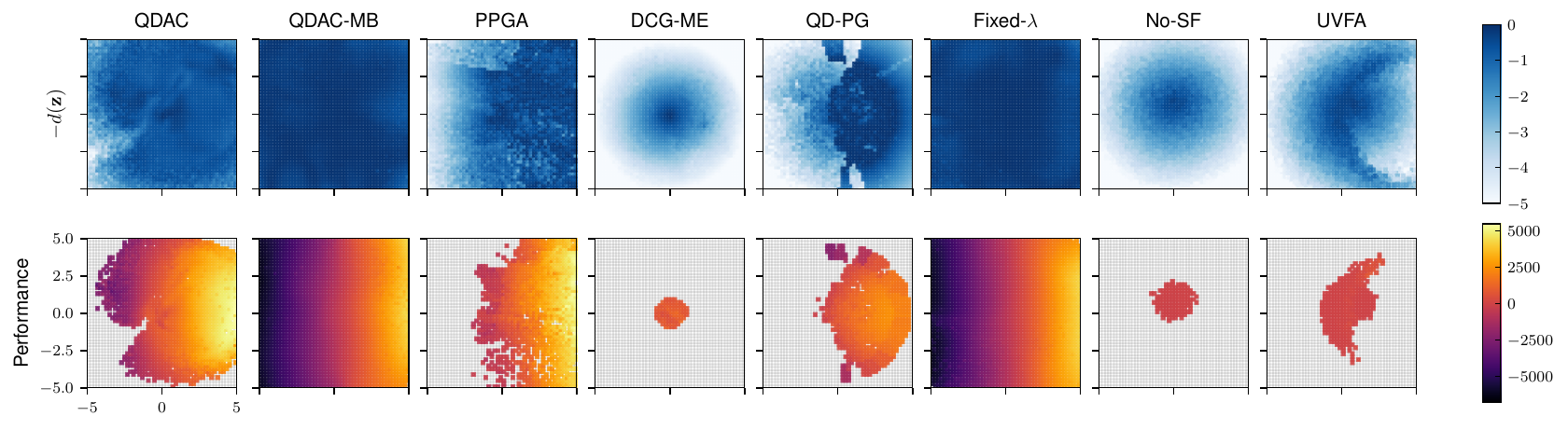}
\caption{%
\textbf{Ant Velocity} Heatmaps of \textbf{(top)} negative distance to skill, \textbf{(bottom)} performance defined in Section~\ref{sec:experiments-metrics}.
The heatmap represents the skill space $\skillSpace = [-5 \text{ m/s}, 5 \text{ m/s}]^2$, of target velocities. 
This space is discretized into cells, with each cell representing a distinct skill $\skill = \begin{bmatrix}v_x & v_y\end{bmatrix}^\intercal$.
In the bottom row, empty cells show which skills are not successfully executed (i.e. $d(\skill) > d_{\mathrm{eval}}$), while colorized cells indicate the performance score obtained for the corresponding skill.
}
\end{figure}

\newpage
\subsection{Results without filtering with $d_{\text{eval}}$}
\begin{figure}[H]
\centering
\includegraphics[width=\textwidth]{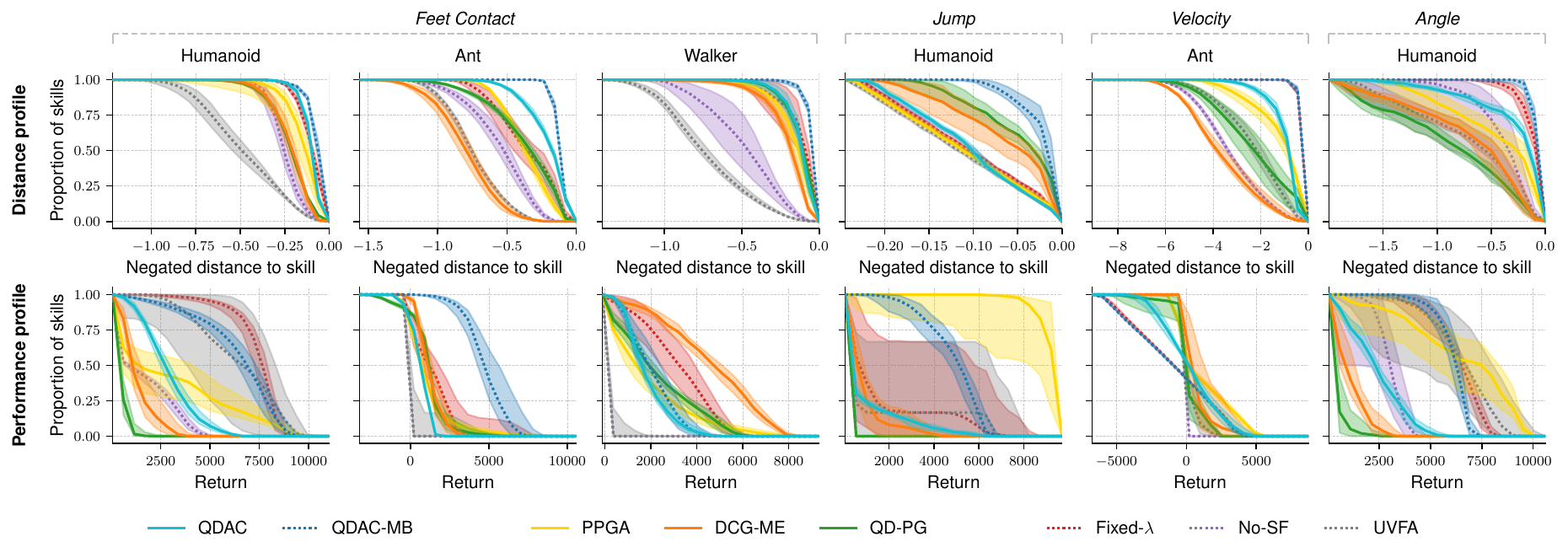}
\caption{%
(\textbf{top}) Distance profiles and (\textbf{bottom}) performance profiles for each task defined in Section~\ref{sec:experiments-metrics} similar to Figure~\ref{fig:profiles} except that skills that are not successfully executed (i.e. $d(\skill) > d_{\mathrm{eval}}$) are \textbf{not} filtered out.
The lines represent the IQM for 10 replications, and the shaded areas correspond to the 95\% CI.
Figure~\ref{fig:profiles_explanation} illustrates how to read distance and performance profiles.
}
\label{fig:profiles_no_d}
\end{figure}

\begin{figure}[H]
\centering
\includegraphics[width=\textwidth]{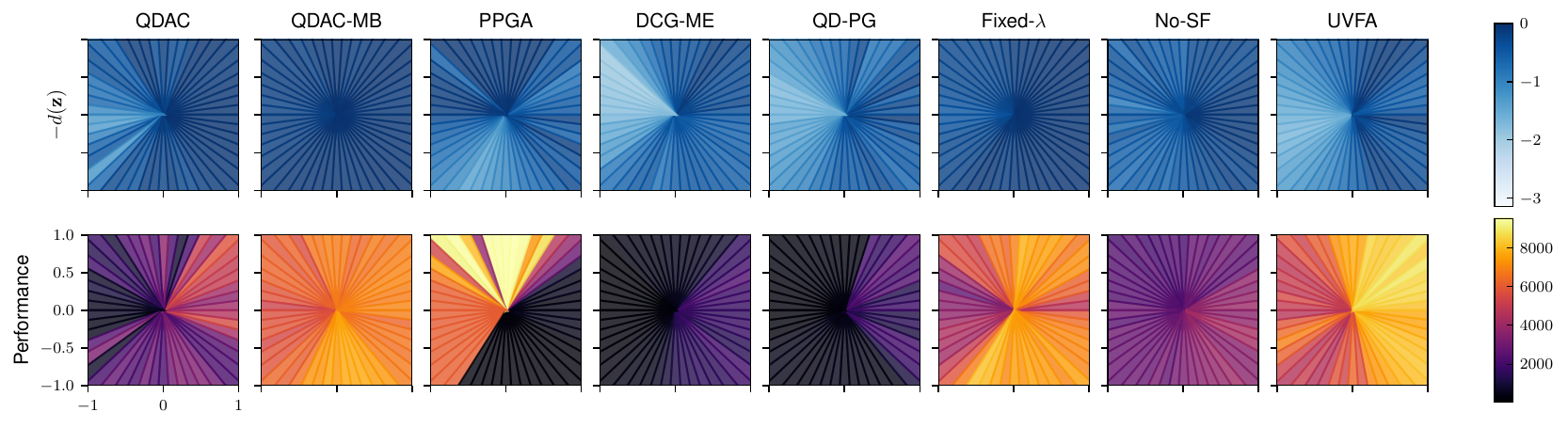}
\caption{%
\textbf{Humanoid Angle} Heatmaps of \textbf{(top)} negative distance to skill, \textbf{(bottom)} performance defined in Section~\ref{sec:experiments-metrics}. In the bottom row, the skills that are not successfully executed (i.e. $d(\skill) > d_{\mathrm{eval}}$) are \textbf{not} filtered out.
}
\label{fig:humanoid-angle-no-d}
\end{figure}

In Figures~\ref{fig:profiles_no_d} to~\ref{fig:humanoid-angle-no-d}, we report the profiles and heatmaps defined in Section~\ref{sec:experiments-metrics} except that skills that are not successfully executed (i.e. $d(\skill) > d_{\mathrm{eval}}$) are \textbf{not} filtered out.

\begin{figure}[H]
\centering
\includegraphics[width=\textwidth]{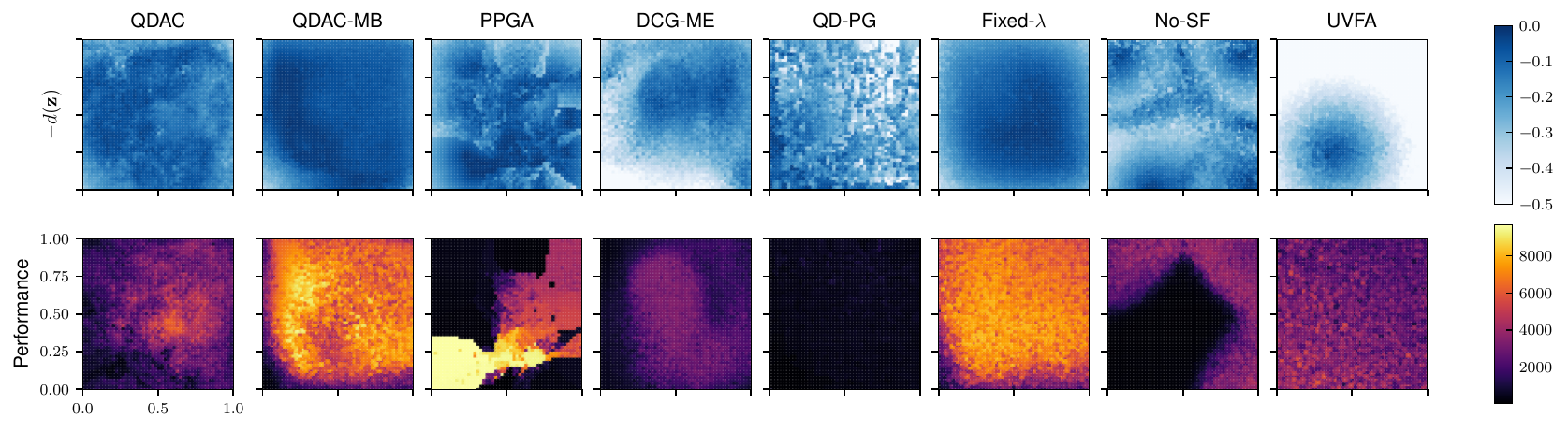}
\caption{%
\textbf{Humanoid Feet Contact} Heatmaps of \textbf{(top)} negative distance to skill, \textbf{(bottom)} performance defined in Section~\ref{sec:experiments-metrics}. In the bottom row, the skills that are not successfully executed (i.e. $d(\skill) > d_{\mathrm{eval}}$) are \textbf{not} filtered out.
}
\end{figure}
\begin{figure}[H]
\centering
\includegraphics[width=\textwidth]{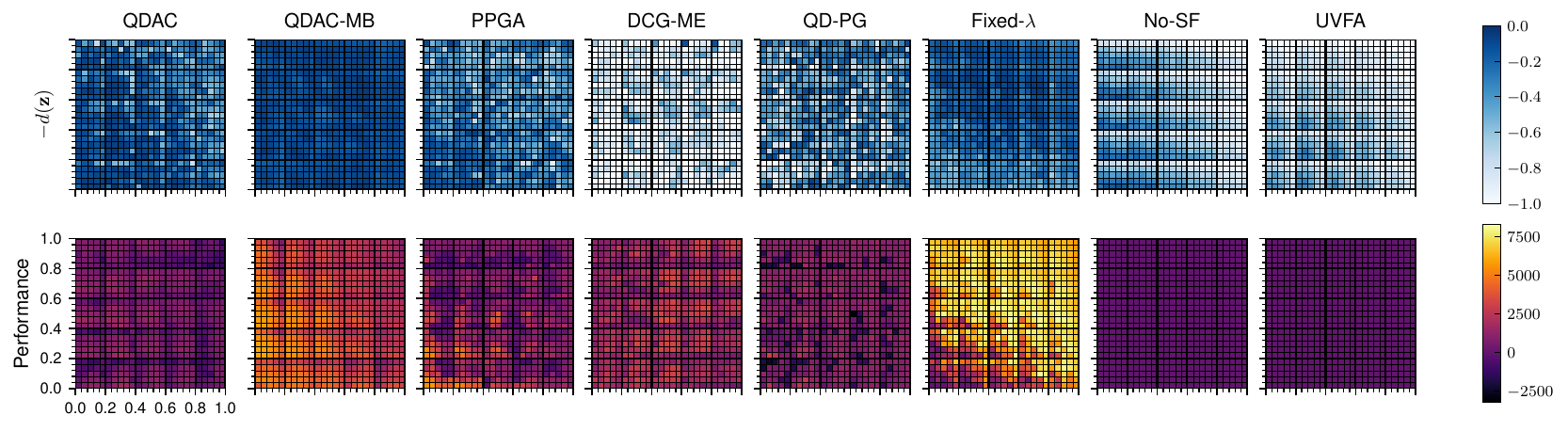}
\caption{%
\textbf{Ant Feet Contact} Heatmaps of \textbf{(top)} negative distance to skill, \textbf{(bottom)} performance defined in Section~\ref{sec:experiments-metrics}. In the bottom row, the skills that are not successfully executed (i.e. $d(\skill) > d_{\mathrm{eval}}$) are \textbf{not} filtered out.
}
\end{figure}
\begin{figure}[H]
\centering
\includegraphics[width=\textwidth]{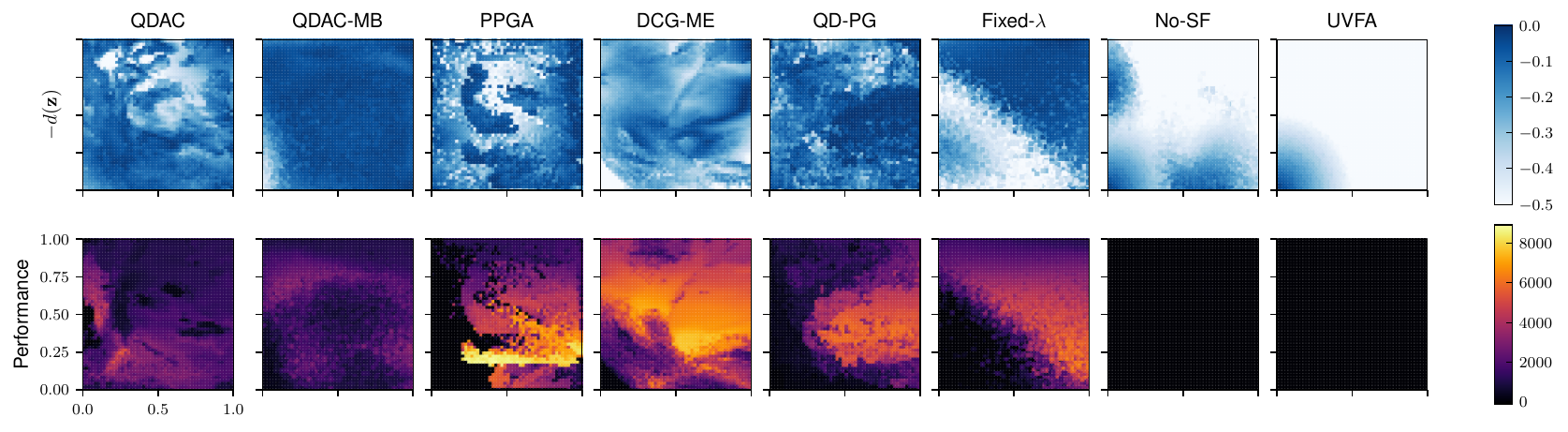}
\caption{%
\textbf{Walker Feet Contact} Heatmaps of \textbf{(top)} negative distance to skill, \textbf{(bottom)} performance defined in Section~\ref{sec:experiments-metrics}. In the bottom row, the skills that are not successfully executed (i.e. $d(\skill) > d_{\mathrm{eval}}$) are \textbf{not} filtered out.
}
\end{figure}
\begin{figure}[H]
\centering
\includegraphics[width=\textwidth]{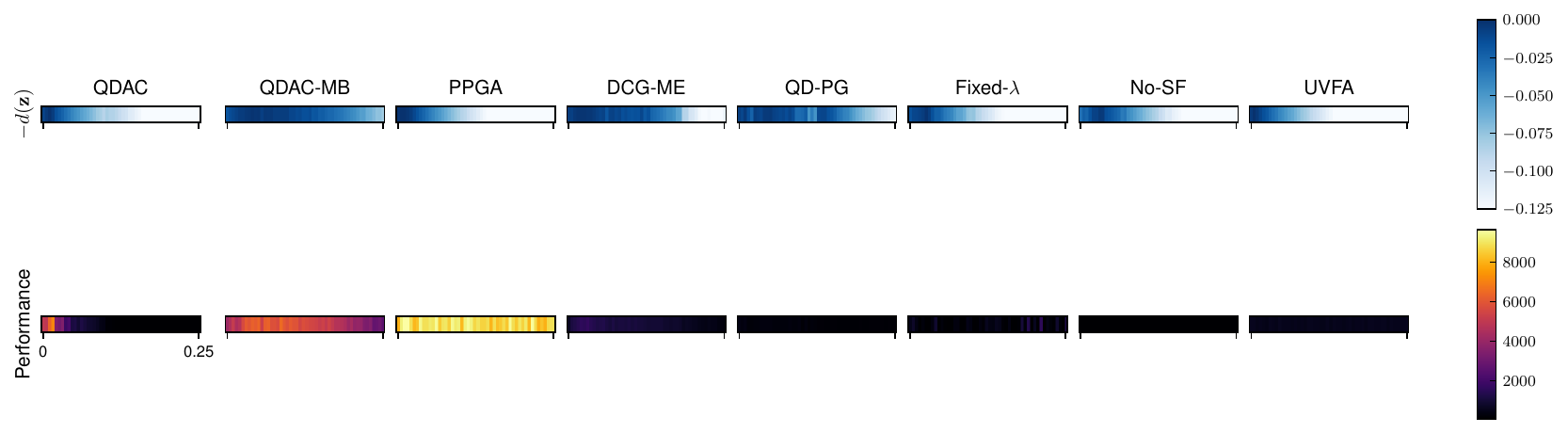}
\caption{%
\textbf{Humanoid Jump} Heatmaps of \textbf{(top)} negative distance to skill, \textbf{(bottom)} performance defined in Section~\ref{sec:experiments-metrics}. In the bottom row, the skills that are not successfully executed (i.e. $d(\skill) > d_{\mathrm{eval}}$) are \textbf{not} filtered out.
}
\end{figure}
\begin{figure}[H]
\centering
\includegraphics[width=\textwidth]{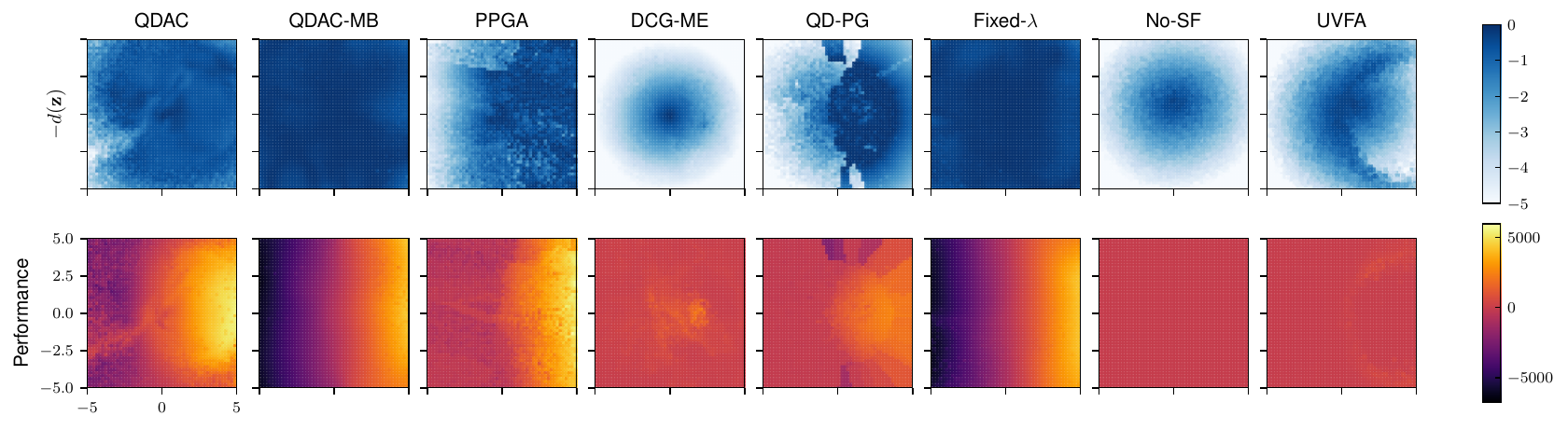}
\caption{%
\textbf{Ant Velocity} Heatmaps of \textbf{(top)} negative distance to skill, \textbf{(bottom)} performance defined in Section~\ref{sec:experiments-metrics}. In the bottom row, the skills that are not successfully executed (i.e. $d(\skill) > d_{\mathrm{eval}}$) are \textbf{not} filtered out.
}
\end{figure}

\newpage
\subsection{Archive Profiles and Heatmaps for \domino, \smerl and \smerlReverse}

\begin{figure}[H]
\centering
\includegraphics[width=0.99\textwidth]{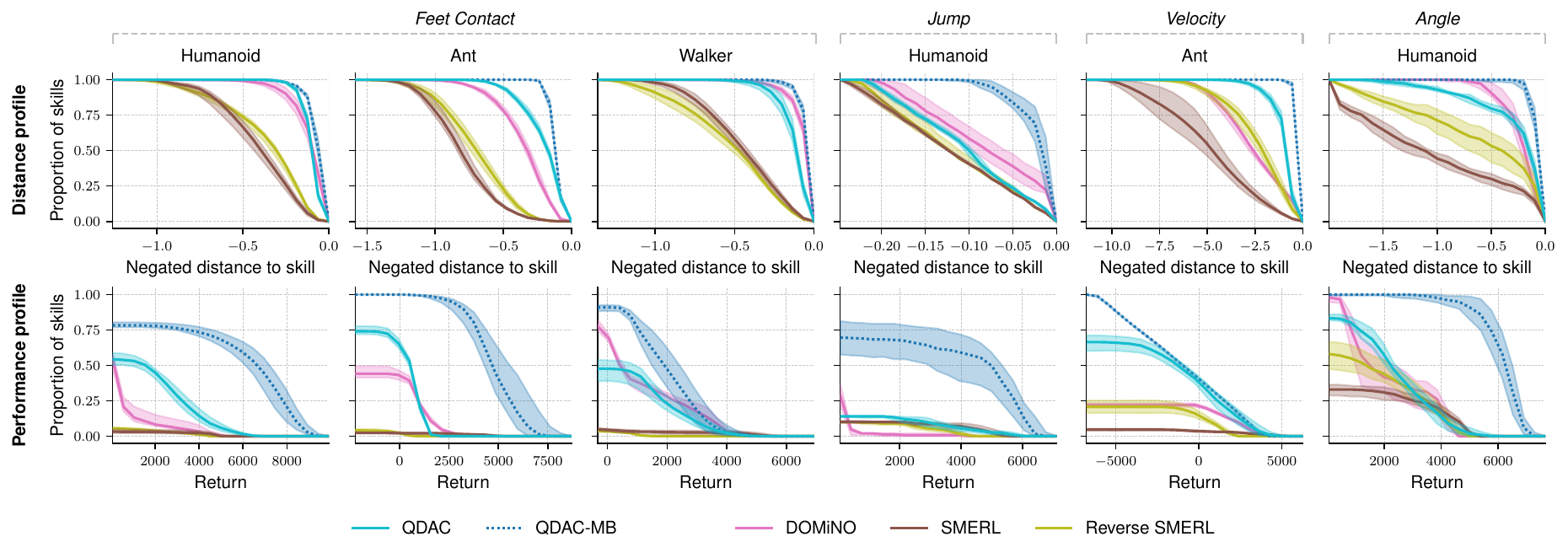}

\caption{%
(\textbf{top}) Distance profiles and (\textbf{bottom}) performance profiles for each task defined in Section~\ref{sec:experiments-metrics}.
We present here the results from \domino, \smerl and \smerlReverse using a method analogous to that in \citep{chalumeau_NeuroevolutionCompetitiveAlternative_2022}. 
The lines represent the IQM for 10 replications, and the shaded areas correspond to the 95\% CI.
Figure~\ref{fig:profiles_explanation} illustrates how to read distance and performance profiles.
}
\label{fig:appendix:profiles:smerl-domino}
\end{figure}

\begin{figure}[H]
\centering
\includegraphics[width=\textwidth]{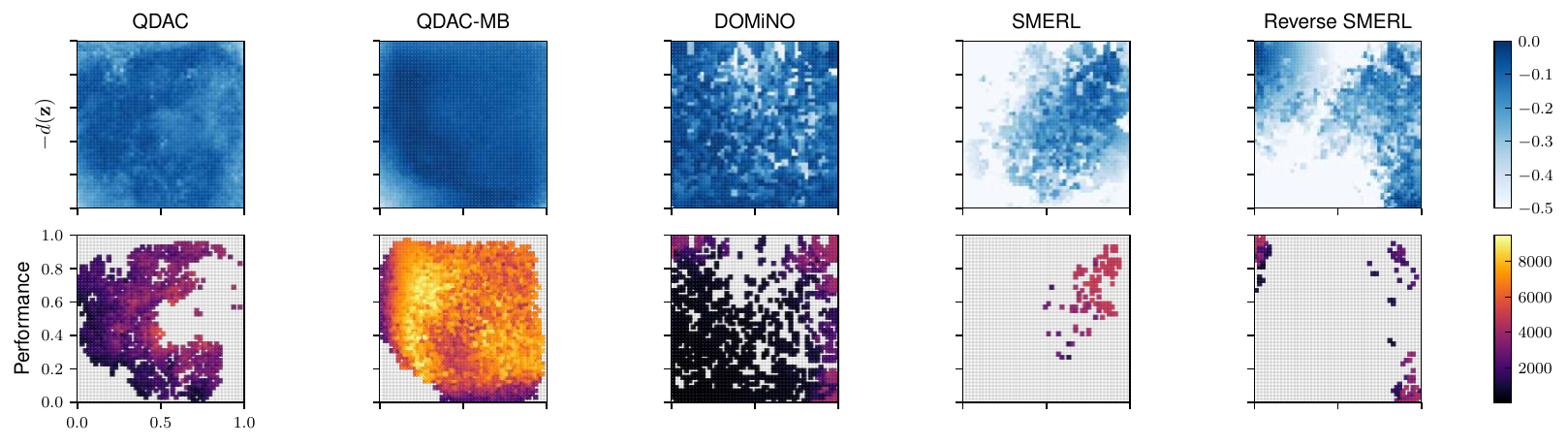}
\caption{%
\textbf{Humanoid Feet Contact} Heatmaps of \textbf{(top)} negative distance to skill, \textbf{(bottom)} performance defined in Section~\ref{sec:experiments-metrics}. We present here the results from \domino, \smerl and \smerlReverse using a method analogous to that in \citep{chalumeau_NeuroevolutionCompetitiveAlternative_2022}. 
The heatmap represents the skill space of feet contacts $\skillSpace = [ 0, 1]^2$. 
This space is discretized into cells, with each cell representing a distinct skill $\skill = \begin{bmatrix}z_1 & z_2\end{bmatrix}^\intercal$, where $z_i$ is the proportion of time that leg $i$ touches the ground over an entire episode.
In the bottom row, empty cells show which skills are not successfully executed (i.e. $d(\skill) > d_{\mathrm{eval}}$), while colorized cells indicate the performance score obtained for the corresponding skill.
}
\label{fig:appendix:humanoid-feet-contact:smerl-domino}
\end{figure}

\begin{figure}[H]
\centering
\includegraphics[width=\textwidth]{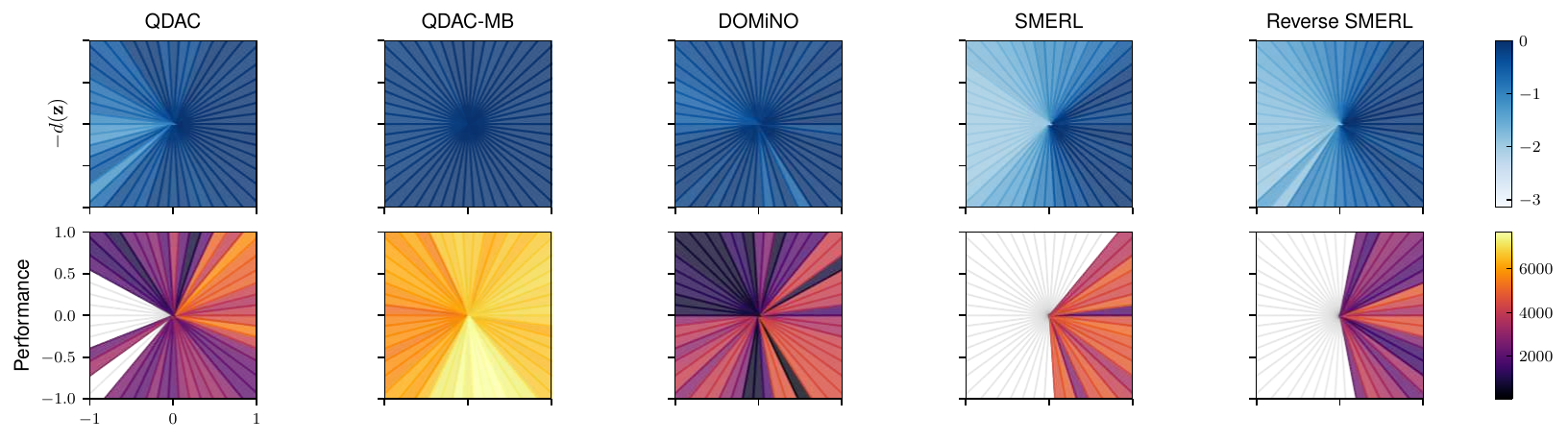}
\caption{%
\textbf{Humanoid Angle} Heatmaps of \textbf{(top)} negative distance to skill, \textbf{(bottom)} performance defined in Section~\ref{sec:experiments-metrics}. We present here the results from \domino, \smerl and \smerlReverse using a method analogous to that in \citep{chalumeau_NeuroevolutionCompetitiveAlternative_2022}.
The heatmap represents the skill space of body angles $\skillSpace = ]-\pi, \pi]$. 
This space is discretized into cells, with each cell representing a distinct skill; in this task, the skills refer to the angle of the humanoid body about the $z$-axis.
In the bottom row, empty cells show which skills are not successfully executed (i.e. $d(\skill) > d_{\mathrm{eval}}$), while colorized cells indicate the performance score obtained for the corresponding skill.
}
\label{fig:appendix:humanoid-angle:smerl-domino}
\end{figure}

In Figures~\ref{fig:appendix:profiles:smerl-domino} to~\ref{fig:appendix:humanoid-angle:smerl-domino}, we present the archive profiles and heatmaps achieved by \domino, \smerl, and \smerlReverse using a method analogous to that in \citep{chalumeau_NeuroevolutionCompetitiveAlternative_2022}: we generate an archive from the intermediate policies encountered during training, and use this archive to compare against \ours and \oursmb.

\begin{figure}[H]
\centering
\includegraphics[width=\textwidth]{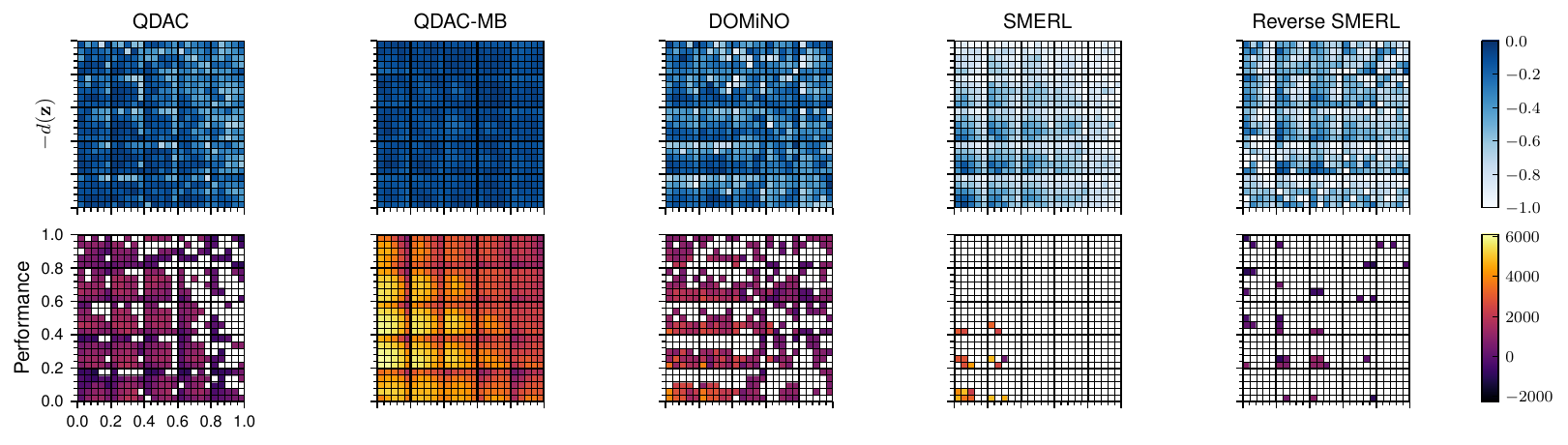}
\caption{%
\textbf{Ant Feet Contact} Heatmaps of \textbf{(top)} negative distance to skill, \textbf{(bottom)} performance defined in Section~\ref{sec:experiments-metrics}. We present here the results from \domino, \smerl and \smerlReverse using a method analogous to that in \citep{chalumeau_NeuroevolutionCompetitiveAlternative_2022}.
The heatmap represents the skill space of feet contacts $\skillSpace = [ 0, 1]^4$. 
This space is discretized into cells, with each cell representing a distinct skill $\skill = \begin{bmatrix}z_1 & z_2 & z_3 & z_4\end{bmatrix}^\intercal$, where $z_i$ is the proportion of time that leg $i$ touches the ground over an entire episode.
In the bottom row, empty cells show which skills are not successfully executed (i.e. $d(\skill) > d_{\mathrm{eval}}$), while colorized cells indicate the performance score obtained for the corresponding skill.
}
\end{figure}
\begin{figure}[H]
\centering
\includegraphics[width=\textwidth]{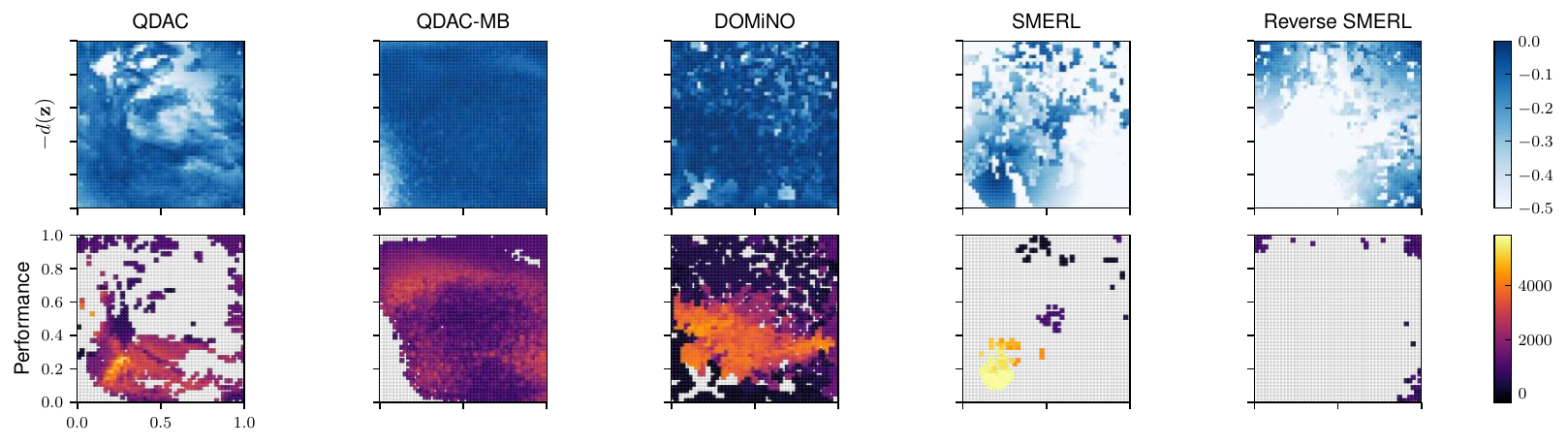}
\caption{%
\textbf{Walker Feet Contact} Heatmaps of \textbf{(top)} negative distance to skill, \textbf{(bottom)} performance defined in Section~\ref{sec:experiments-metrics}. We present here the results from \domino, \smerl and \smerlReverse using a method analogous to that in \citep{chalumeau_NeuroevolutionCompetitiveAlternative_2022}.
The heatmap represents the skill space of feet contacts $\skillSpace = [ 0, 1]^2$. 
This space is discretized into cells, with each cell representing a distinct skill $\skill = \begin{bmatrix}z_1 & z_2\end{bmatrix}^\intercal$, where $z_i$ is the proportion of time that leg $i$ touches the ground over an entire episode.
In the bottom row, empty cells show which skills are not successfully executed (i.e. $d(\skill) > d_{\mathrm{eval}}$), while colorized cells indicate the performance score obtained for the corresponding skill.
}
\end{figure}
\begin{figure}[H]
\centering
\includegraphics[width=\textwidth]{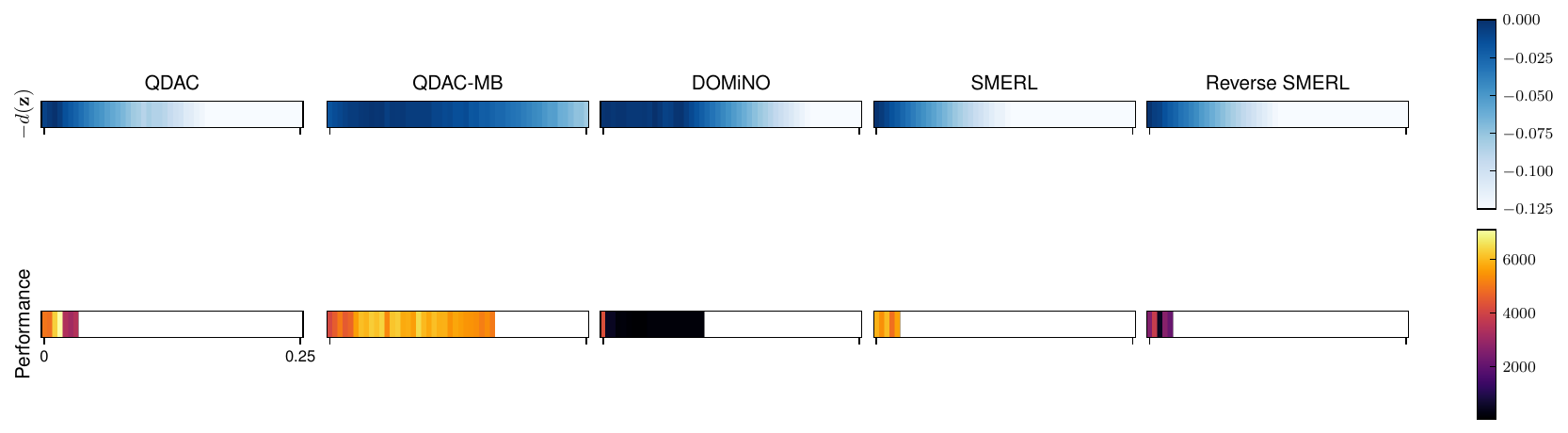}
\caption{%
\textbf{Humanoid Jump} Heatmaps of \textbf{(top)} negative distance to skill, \textbf{(bottom)} performance defined in Section~\ref{sec:experiments-metrics}. We present here the results from \domino, \smerl and \smerlReverse using a method analogous to that in \citep{chalumeau_NeuroevolutionCompetitiveAlternative_2022}.
The heatmap represents the skill space of jumping skills $\skillSpace = [0, 0.25]$. 
This space is discretized into cells, with each cell representing a distinct skill; in this task, the skills refer to the average of the lowest foot heights over an entire episode.
In the bottom row, empty cells show which skills are not successfully executed (i.e. $d(\skill) > d_{\mathrm{eval}}$), while colorized cells indicate the performance score obtained for the corresponding skill.
}
\end{figure}
\begin{figure}[H]
\centering
\includegraphics[width=\textwidth]{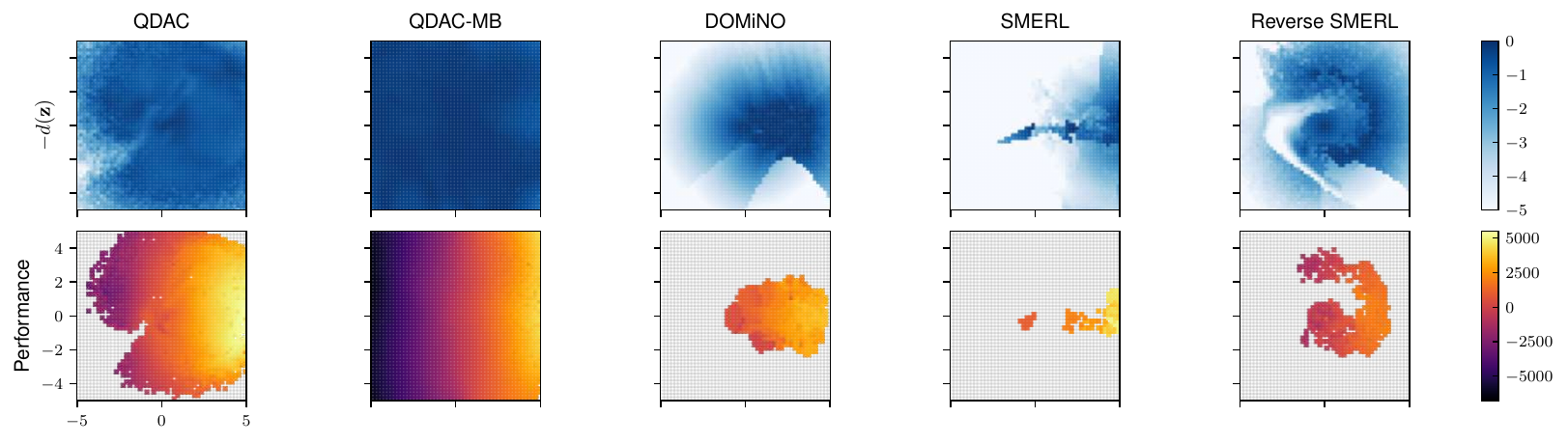}
\caption{%
\textbf{Ant Velocity} Heatmaps of \textbf{(top)} negative distance to skill, \textbf{(bottom)} performance defined in Section~\ref{sec:experiments-metrics}. We present here the results from \domino, \smerl and \smerlReverse using a method analogous to that in \citep{chalumeau_NeuroevolutionCompetitiveAlternative_2022}.
The heatmap represents the skill space $\skillSpace = [-5 \text{ m/s}, 5 \text{ m/s}]^2$, of target velocities. 
This space is discretized into cells, with each cell representing a distinct skill $\skill = \begin{bmatrix}v_x & v_y\end{bmatrix}^\intercal$.
In the bottom row, empty cells show which skills are not successfully executed (i.e. $d(\skill) > d_{\mathrm{eval}}$), while colorized cells indicate the performance score obtained for the corresponding skill.
}
\end{figure}
\newpage
\section{Theoretical Results}
\label{appendix:proofs}
\begin{proposition*}
\MyLabel{proposition-1}
Consider an infinite horizon, finite MDP with observable features in $\featSpace$. Let $\policy$ be a policy and let $\successorFeatures$ be the discounted successor features.
Then, for all skills $\skill \in \skillSpace$, we can derive an upper bound for the distance between $\skill$ and the expected features under $\policy$:
\begin{equation}
\MyLabel{eq:upper-bound}
\norm{\featAvg - \skill} \leq \expect{\policySkill}{\norm{(1 - \discount) \successorFeatures(\state, \skill) - \skill}}
\end{equation}
\end{proposition*}

\begin{proof}
For all states $\state \in \stateSpace$, the Bellman equation for $\successorFeatures$ gives:
\begin{align}
\successorFeatures(\state, \action, \skill) = \feat(\state, \action) + \discount \expect{\state^\prime \sim p(. | \state, \action)}{\successorFeatures(\state^\prime, \skill)}\label{eq:sf-bellman}
\end{align}

For all skills $\skill \in \skillSpace$ and for all sequences of $T$ states $(\state_0, \action_0, \dots, \state_{T-1})$ sampled from $\policySkill$, we have:
\begin{align*}
&\norm{\frac{1}{T} \sum_{t=0}^{T-1} \feat(\state_t, \action_t) - \skill}\\
%
%
&= \norm{\frac{1}{T} \sum_{t=0}^{T-1} \left( \successorFeatures(\state_t, \action_t, \skill) - \discount \expect{\state_{t+1} \sim p(. | \state_t, \action_t)}{\successorFeatures(\state_{t+1}, \skill)} - \skill \right)} \tag*{(Equation~\ref{eq:sf-bellman})}\\
%
%
&= \norm{\frac{1}{T} \sum_{t=0}^{T-1} \left( (1 - \discount) \successorFeatures(\state_t, \action_t, \skill) - \skill \right) + \frac{\discount}{T} \sum_{t=0}^{T-1} \left( \successorFeatures(\state_t, \action_t, \skill) - \expect{\state_{t+1} \sim p(. | \state_t, \action_t)}{\successorFeatures(\state_{t+1}, \skill)} \right)}\\
&\leq \norm{\frac{1}{T} \sum_{t=0}^{T-1} (1 - \discount) \successorFeatures(\state_t, \action_t, \skill) - \skill} + \discount \norm{\frac{1}{T} \sum_{t=0}^{T-1} \left( \successorFeatures(\state_t, \action_t, \skill) - \expect{\state_{t+1} \sim p(. | \state_t, \action_t)}{\successorFeatures(\state_{t+1}, \skill)} \right)} \tag*{(triangular inequality)}
\end{align*}

We denote $\stateDist(\state) = \lim_{t\rightarrow\infty} P(\state_t = \state | \state_0, \policy_\skill)$ the stationary distribution of states under $\policy_\skill$, which we assume exists and is independent of $\state_0$. Consequently, by taking the right term to the limit as $T \rightarrow \infty$:
\begin{align*}
&\lim_{T\rightarrow\infty} \frac{1}{T} \sum_{t=0}^{T-1} \left( \successorFeatures(\state_t, \action_t, \skill) - \expect{\state_{t+1} \sim p(. | \state_t, \action_t)}{\successorFeatures(\state_{t+1}, \skill)} \right)\\
&= \lim_{T\rightarrow\infty} \frac{1}{T} \sum_{t=0}^{T-1} \successorFeatures(\state_t, \action_t, \skill) - \lim_{T\rightarrow\infty} \frac{1}{T} \sum_{t=0}^{T-1} \left( \expect{\state_{t+1} \sim p(. | \state_t, \action_t)}{\successorFeatures(\state_{t+1}, \skill)} \right)\\
&= \expect{\substack{\state \sim \stateDist\\ \action \sim \policy(. | \state, \skill)}}{\successorFeatures(\state, \action, \skill)} - \expect{\substack{\state \sim \stateDist\\ \action \sim \policy(. | \state, \skill)}}{\expect{\state^\prime \sim p(. | \state, \action)}{\successorFeatures(\state^\prime, \skill)}}\\
&= \expect{\state \sim \stateDist}{\successorFeatures(\state, \skill)} - \expect{\substack{\state \sim \stateDist\\ \action \sim \policy(. | \state, \skill)\\ \state^\prime \sim p(. | \state, \action)}}{\successorFeatures(\state^\prime, \skill)}\\
&= \expect{\state \sim \stateDist}{\successorFeatures(\state, \skill)} - \expect{s^\prime \sim \stateDist}{\successorFeatures(\state^\prime, \skill)}\\
&= 0
\end{align*}

Furthermore, by taking the left term to the limit as $T \rightarrow \infty$:
\begin{align*}
\lim_{T\rightarrow\infty} \frac{1}{T} \sum_{t=0}^{T-1} (1 - \discount) \successorFeatures(\state_t, \action_t, \skill) - \skill &= (1 - \discount) \expect{\substack{\state \sim \stateDist\\ \action \sim \policy(. | \state, \skill)}}{\successorFeatures(\state, \action, \skill)} - \skill\\
&= (1 - \discount) \expect{\state \sim \stateDist}{\successorFeatures(\state, \skill)} - \skill
\end{align*}

Finally, by taking the inequality to the limit as $T \rightarrow \infty$, we get:
\begin{align*}
\norm{\featAvg - \skill } &\leq \norm{(1 - \discount) \expect{\policySkill}{\successorFeatures(\state, \skill)} - \skill} + \norm{\vec 0}\\
\alignedbox{\norm{\featAvg - \skill}}{\leq \expect{\policySkill}{\norm{(1 - \discount) \successorFeatures(\state, \skill) - \skill}}} \tag*{(Jensen's inequality)}
\end{align*}
\end{proof}

\newpage
\begin{propositionAppendix}
\label{proposition-2}
Consider a continuous MDP with a bounded feature space $\featSpace$, a skill $\skill \in \skillSpace$, and $\policy$ a policy such that the sequence $\left(\frac{1}{T} \sum_{t=0}^{T-1}\feat_{t}\right)_{T\geq 1}$ almost surely\footnotemark converges for trajectories sampled from $\policySkill$. 
If we write $\eps\defeq\sup_t\expect{\policySkill}{\norm{\feat_t + \gamma \successorFeatures(\state_t, \action_t, \skill) - \successorFeatures(\state_{t+1}, \action_{t+1}, \skill)}}$, then:
\begin{align}
    \expect{\policySkill}{\norm{ \lim_{T\rightarrow\infty} \frac{1}{T} \sum_{t=0}^{T-1} \feat_{t} - \skill } } \leq \sup_t \expect{\policySkill}{\norm{(1-\discount) \successorFeatures(\state_t, \action_t, \skill) - \skill}} + \eps 
\end{align}
Furthermore, it is worth noting that if the MDP dynamics $p$ and $\pi$ are deterministic, then $\epsilon=0$.
\end{propositionAppendix}
\footnotetext{\textit{almost sure} refers to the almost sure convergence from probability theory where rollouts are sampled from $\policySkill$.}

\newcommand{\piZ}{\pi_\skill}
\newcommand{\EZ}[1]{\expect{\piZ}{ { #1 } }}
\newcommand{\SF}{\successorFeatures}
\newcommand{\SFA}{\SF^a}

\begin{proof}
    Let $\skill\in\skillSpace$.

    To make the proof easier to read, we use the following notations:
    \begin{align*}
        \successorFeatures_t &\defeq \successorFeatures\left( \state_t, \action_t, \skill \right)
    \end{align*}

    We define $\beta$ as follows:
\begin{align}
    \beta\defeq \sup_t \EZ{\norm{(1-\discount) \successorFeatures_t - \skill}}
\end{align}

    Then we have, for all $t$,
    \begin{align}
       \EZ{\norm{(1-\discount) \successorFeatures_t - \skill}} &\leq \beta \label{appendix:eq:hyp:1} \\
        \EZ{\norm{\feat_t + \gamma \successorFeatures_t - \successorFeatures_{t+1}}} &\leq \epsilon \label{appendix:eq:hyp:2}
    \end{align}

    The Bellman equation applied to Successor Features (SF) can be written:
    \begin{align}
        \SF_t &= \feat_t + \gamma \expect{\piZ}{ \left. \successorFeatures_{t+1} \right\vert \state_t, \action_t} \label{appendix:eq:BE:SF} \\
        \text{or also: }\feat_t &= \SF_t - \gamma \expect{\piZ}{  \left. \successorFeatures_{t+1} \right\vert \state_t, \action_t} \label{appendix:eq:BE:feat}
    \end{align}

    We can now derive an upper bound for $\norm{ \frac{1}{T} \sum_{t=0}^{T-1} \feat_t - \skill }$. 
    For all sequences of $T$ states $\state_{0:T-1}$ we have:
    \begin{align*}
        \norm{ \frac{1}{T} \sum_{t=0}^{T-1} \feat_t - \skill }
        &= \norm{ \frac{1}{T} \sum_{t=0}^{T-1} \left( \feat_t - \skill \right) } \\
        &= \norm{ \frac{1}{T} \sum_{t=0}^{T-1} \left( \successorFeatures_t - \gamma \expect{\piZ}{  \left. \successorFeatures_{t+1} \right\vert \state_t, \action_t} - \skill \right) } \tag*{(from Equation~\ref{appendix:eq:BE:feat})} \\
        &= \norm{ \frac{1}{T} \sum_{t=0}^{T-1} \left( (1-\discount) \successorFeatures_t + \discount \successorFeatures_t - \gamma \expect{\piZ}{  \left. \successorFeatures_{t+1} \right\vert \state_t, \action_t} - \skill \right) } \\
        &= \norm{ \frac{1}{T} \sum_{t=0}^{T-1} \left( (1-\discount) \successorFeatures_t - \skill \right) + \frac{1}{T} \sum_{t=0}^{T-1}\left( \discount \successorFeatures_t - \gamma \expect{\piZ}{  \left. \successorFeatures_{t+1} \right\vert \state_t, \action_t} \right) } \\
        &\leq \norm{ \frac{1}{T} \sum_{t=0}^{T-1} \left( (1-\discount) \successorFeatures_t - \skill \right)} + \norm{\frac{1}{T} \sum_{t=0}^{T-1}\left( \discount \successorFeatures_t - \gamma \expect{\piZ}{  \left. \successorFeatures_{t+1} \right\vert \state_t, \action_t} \right) } \tag*{(triangular inequality)}
        \end{align*}

Thus,
\begin{align}
\begin{split}
\label{appendix:ineq:tempResultProof}
    \EZ{\norm{ \frac{1}{T} \sum_{t=0}^{T-1} \feat_t - \skill }} 
&\leq \EZ{\norm{ \frac{1}{T} \sum_{t=0}^{T-1} \left( (1-\discount) \successorFeatures_t - \skill \right)}} \\
&\qquad + \EZ{\norm{\frac{1}{T} \sum_{t=0}^{T-1}\left( \discount \successorFeatures_t - \gamma \expect{\piZ}{  \left. \successorFeatures_{t+1} \right\vert \state_t, \action_t} \right) }}
\end{split}
    \end{align}

We consider now the two terms on the right hand-side separately.
First of all, we prove that the first term is lower than or equal to $\beta$:
\begin{align*}
\expect{\piZ}{\norm{ \frac{1}{T} \sum_{t=0}^{T-1} \left( (1-\discount) \successorFeatures_t - \skill \right)} }
&\leq \expect{\piZ}{\frac{1}{T} \sum_{t=0}^{T-1} \norm{  \left( (1-\discount) \successorFeatures_t - \skill \right)}} \tag*{(triangular inequality)}\\
&\leq \frac{1}{T} \sum_{t=0}^{T-1} \expect{\piZ}{\norm{  \left( (1-\discount) \successorFeatures_t - \skill \right)}}\\
&\leq \frac{1}{T} \sum_{t=0}^{T-1} \beta \tag*{(from Equation~\ref{appendix:eq:hyp:1})}\\
&\leq \beta \numberthis \label{appendix:proof:tempResult1}
\end{align*}

Also, we can prove that the second term of the right-hand side in Equation~\ref{appendix:ineq:tempResultProof} is lower than or equal to $\epsilon+\eta_T$ where $\lim_{T\rightarrow\infty}\eta_T = 0$.
For all sequences of $T$ states $\state_{0:T-1}$, we have:
\begin{align*}
\sum_{t=0}^{T-1}\left( \discount \successorFeatures_t - \gamma \expect{\piZ}{  \left. \successorFeatures_{t+1} \right\vert \state_t, \action_t} \right) 
&= \sum_{t=0}^{T-1} \discount \successorFeatures_t  - \sum_{t=0}^{T-1} \gamma \expect{\piZ}{  \left. \successorFeatures_{t+1} \right\vert \state_t, \action_t}\\
&= \gamma \successorFeatures_0 - \gamma \expect{\piZ}{  \left. \successorFeatures_T \right\vert \state_{T-1}, \action_{T-1}} + \sum_{t=1}^{T-1} \discount \successorFeatures_t  - \sum_{t=0}^{T-2} \gamma \expect{\piZ}{  \left. \successorFeatures_{t+1} \right\vert \state_t, \action_t} \\
&= \gamma \successorFeatures_0 - \gamma \expect{\piZ}{  \left. \successorFeatures_T \right\vert \state_{T-1}, \action_{T-1}} + \sum_{t=0}^{T-2} \discount \successorFeatures_{t+1}  - \sum_{t=0}^{T-2} \gamma \expect{\piZ}{  \left. \successorFeatures_{t+1} \right\vert \state_t, \action_t} \\
&= \gamma \successorFeatures_0 - \gamma \expect{\piZ}{  \left. \successorFeatures_T \right\vert \state_{T-1}, \action_{T-1}} +  \sum_{t=0}^{T-2} \left( \discount \successorFeatures_{t+1}  - \gamma \expect{\piZ}{  \left. \successorFeatures_{t+1} \right\vert \state_t, \action_t} \right)
\end{align*}

Thus, after dividing by $T$ and applying the norm and expectation, we get:
\begin{align*}
\EZ{\norm{\frac{1}{T}\sum_{t=0}^{T-1}\left( \discount \successorFeatures_t - \gamma \expect{\piZ}{  \left. \successorFeatures_{t+1} \right\vert \state_t, \action_t} \right) }} 
&\leq  \EZ{\norm{\frac{1}{T} \left( \gamma \successorFeatures_0 - \gamma \expect{\piZ}{  \left. \successorFeatures_T \right\vert \state_{T-1}, \action_{T-1}} \right)}}  \\
&\quad  + \EZ{\norm{\frac{1}{T} \sum_{t=0}^{T-2} \left( \discount \successorFeatures_{t+1}  - \gamma \expect{\piZ}{  \left. \successorFeatures_{t+1} \right\vert \state_t, \action_t} \right)}}  \tag*{(triangular inequality)}
\end{align*}

Let $\eta_T \defeq \EZ{\norm{\frac{1}{T} \left( \gamma \successorFeatures_0 - \gamma \expect{\piZ}{  \left. \successorFeatures_T \right\vert \state_{T-1}, \action_{T-1}} \right)}}$, we then have:
\begin{align*}
&\EZ{\norm{\frac{1}{T}\sum_{t=0}^{T-1}\left( \discount \successorFeatures_t - \gamma \expect{\piZ}{  \left. \successorFeatures_{t+1} \right\vert \state_t, \action_t} \right) }} \\
&\leq  \eta_T + \EZ{\norm{\frac{1}{T} \sum_{t=0}^{T-2} \left( \discount \successorFeatures_{t+1}  - \gamma \expect{\piZ}{  \left. \successorFeatures_{t+1} \right\vert \state_t, \action_t} \right)}}  \tag*{(triangular inequality)} \\
&\leq \eta_T + \EZ{\frac{1}{T} \sum_{t=0}^{T-2} \norm{ \discount \successorFeatures_{t+1}  - \gamma \expect{\piZ}{  \left. \successorFeatures_{t+1} \right\vert \state_t, \action_t} }} \\
&\leq \eta_T + \EZ{\frac{1}{T} \sum_{t=0}^{T-2} \norm{ \discount \successorFeatures_{t+1}  + \feat_t - \successorFeatures_t }} \tag*{(from Equation~\ref{appendix:eq:BE:SF})} \\
&\leq \eta_T + \frac{1}{T} \sum_{t=0}^{T-2} \EZ{ \norm{ \discount \successorFeatures_{t+1}  + \feat_t - \successorFeatures_t }} \\
&\leq \eta_T + \frac{1}{T} \sum_{t=0}^{T-2} \EZ{ \norm{ \feat_t + \discount \successorFeatures_{t+1} - \successorFeatures_t }} \\
&\leq \eta_T + \frac{1}{T} \sum_{t=0}^{T-2} \epsilon \\
&\leq \eta_T + \frac{T-1}{T} \epsilon \numberthis \label{appendix:proof:tempResultEpsilon2}
\end{align*}

After combining the two previously derived Equations~\ref{appendix:proof:tempResult1} and~\ref{appendix:proof:tempResultEpsilon2}, we get:
\begin{align}
    \EZ{\norm{ \frac{1}{T} \sum_{t=0}^{T-1} \feat_t - \skill }}  \leq \beta + \eta_T + \frac{T-1}{T} \epsilon
\end{align}

Now we intend to prove that $\lim_{T\rightarrow \infty}\eta_T = 0$
\begin{align*}
\eta_T &= \EZ{\norm{\frac{1}{T} \left( \gamma \successorFeatures_0 - \gamma \expect{\piZ}{  \left. \successorFeatures_T \right\vert \state_{T-1}, \action_{T-1}} \right)}} \\
&= \frac{\gamma}{T}\EZ{\norm{ \left( \successorFeatures_0 - \expect{\piZ}{  \left. \successorFeatures_T \right\vert \state_{T-1}, \action_{T-1}} \right)}} \\
&\leq \frac{\gamma}{T}\EZ{\norm{ \successorFeatures_0} + \norm{\expect{\piZ}{  \left. \successorFeatures_T \right\vert \state_{T-1}, \action_{T-1}} }} \tag*{(triangular inequality)} \\
&\leq \frac{\gamma}{T}\left(\EZ{\norm{ \successorFeatures_0}} + \EZ{\norm{\expect{\piZ}{  \left. \successorFeatures_T \right\vert \state_{T-1}, \action_{T-1}} }} \right)
\end{align*}
As the space of features $\featSpace$ is bounded, there exist a $\rho>0$ such that for all $\feat\in\featSpace$, $\norm{\feat} \leq \rho$.
Hence, for all $t$, $\norm{\successorFeatures_t}=\EZ{\left. \norm{\sum_{i=0}^{\infty}   \gamma^i\feat_{t+i}  } \right\vert \state_t, \action_t } 
\leq \EZ{\left. \sum_{i=0}^{\infty}   \gamma^i\norm{\feat_{t+i}  } \right\vert \state_t, \action_t } \leq \frac{\rho}{1-\gamma}$. Hence,
\begin{align*}
\eta_T 
&\leq \frac{\discount}{T} \left( \frac{\rho}{1-\discount} + \EZ{\norm{\expect{\piZ}{  \left. \successorFeatures_T \right\vert \state_{T-1}, \action_{T-1}} }} \right) \\
&\leq \frac{\discount}{T} \left( \frac{\rho}{1-\discount} + \EZ{\expect{\piZ}{  \left. \norm{\successorFeatures_T} \right\vert \state_{T-1}, \action_{T-1}} } \right) \tag*{(Jensen's inequality)} \\
&\leq \frac{\discount}{T} \left( \frac{\rho}{1-\discount} + \EZ{ \norm{\successorFeatures_T} }  \right) \tag*{(law of total expectation)} \\
&\leq \frac{\discount}{T} \left( \frac{\rho}{1-\discount} + \frac{\rho}{1-\discount}  \right)  \\
&\leq \frac{1}{T} \left( \frac{2\rho\gamma}{1-\discount} \right)  \numberthis \label{appendix:proof:sandwitchUpper}
\end{align*}
Then, knowing that for all $T$, we have $0 \leq \eta_T$, the squeeze theorem ensures that $\lim_{T\rightarrow\infty}\eta_T = 0$.

\newcommand{\absv}[1]{\left\vert { #1 } \right\vert}

Now we will prove that the left-hand side of Equation~\ref{appendix:ineq:tempResultProof} converges, let $X_T \defeq \norm{ \frac{1}{T} \sum_{t=0}^{T-1} \feat_t - \skill }$.
For all $T$, 
\begin{align*}
    \absv{X_T} 
&\leq \frac{1}{T} \sum_{t=0}^{T-1} \norm{\feat_t} + \norm{\skill} \tag*{(triangular inequality)} \\
&\leq \rho + \norm{\skill}
\end{align*}
Moreover, $\skill$ is a fixed variable, which means that $\absv{X_T}$ is bounded.
In addition, $\EZ{\rho + \norm{\skill}} < \infty$, and the sequence $(X_T)_{T\geq 1}$ converges almost surely (since $\left(\frac{1}{T} \sum_{t=0}^{T-1} \feat_t\right)_{T\geq 1 }$ converges almost surely by hypothesis).
The dominated convergence theorem then ensures that $(\EZ{X_T})_{T\geq 1}$ converges and:
\begin{align}
    \lim_{T\rightarrow\infty}\EZ{X_T} = \EZ{\lim_{T\rightarrow\infty}X_T}
\end{align}

Finally, by taking the Equation~\ref{appendix:ineq:tempResultProof} to the limit as $T\rightarrow\infty$, we get:
\begin{align*}
\lim_{T\rightarrow\infty}\EZ{\underbrace{\norm{ \frac{1}{T} \sum_{t=0}^{T-1} \feat_t - \skill }}_{X_T}}  &\leq \lim_{T\rightarrow\infty} \left(\beta + \eta_T + \frac{T-1}{T} \epsilon \right) \\
\EZ{\norm{\lim_{T\rightarrow\infty} \frac{1}{T} \sum_{t=0}^{T-1} \feat_t - \skill }}  &\leq  \beta + \underbrace{\lim_{T\rightarrow\infty}\eta_T}_{=0} + \underbrace{\lim_{T\rightarrow\infty}\frac{T-1}{T}}_{=1} \epsilon \\
\alignedbox{{\EZ{\norm{\lim_{T\rightarrow\infty} \frac{1}{T} \sum_{t=0}^{T-1} \feat_t - \skill }}}}{\leq  \beta + \epsilon}
\end{align*}

\end{proof}





\newpage
\section{Additional Training Details}

\subsection{Expanded Information on \ours}
\label{appendix:qdac-expanded-info}

The policy parameters $\params_{\policy}$ are optimized to maximize the objective function from Equation~\ref{eq:actor-obj}.
To that end, we use the Soft Actor-Critic (SAC) algorithm with adjusted temperature $\temperatureSac$~\citep{haarnoja_SoftActorCriticAlgorithms_2019}.
Then the objective from Equation~\ref{eq:actor-obj} needs to be slightly modified to be optimized by SAC:
\begin{equation}
    J_{\policy} (\theta_\policy) = \left(1 - \lagrange(\state, \skill) \right) {\color{perf} \QFunction(\state, \action, \skill)} - \lagrange(\state, \skill) {\color{dist} \norm{ (1-\discount) \successorFeatures(\state, \action, \skill) - \skill}} + \underbracket[0.5pt]{\temperatureSac \log \policy (\action | \state, \skill)}_{\substack{\text{Entropy regularization}\\\text{term used in SAC}}}
\end{equation}

Using the same notations as~\citep{haarnoja_SoftActorCriticAlgorithms_2019}, each action $\action$ returned by the policy $\policy$ can be seen as the function of the state $\state$, the skill $\skill$, and a random noise $\epsilon$: $\action = f_{\params_\policy} (\state, \skill, \epsilon)$.
Then the complete form of the actor's objective function is as follows:
\begin{equation}
\label{eq:oursSacObjective}
    J_{\policy} (\theta_\policy) = \left(1 - \lagrange(\state, \skill) \right) {\color{perf} \QFunction(\state, f_{\params_\policy} (\state, \skill, \epsilon), \skill)} - \lagrange(\state, \skill) {\color{dist} \norm{ (1-\discount) \successorFeatures(\state, f_{\params_\policy} (\state, \skill, \epsilon), \skill) - \skill}} + \temperatureSac \log \policy (f_{\params_\policy} (\state, \skill, \epsilon) | \state, \skill)
\end{equation}

In our setup, the policy $\policy$ outputs a vector $\mu$ and a vector of standard deviations $\begin{pmatrix}\sigma_1 & \cdots & \sigma_n\end{pmatrix}$. 
The action $\action$ is computed as follows: $\action=\mu + \begin{pmatrix}\sigma_1 \epsilon_1 & \cdots & \sigma_n \epsilon_n \end{pmatrix}$, where $\epsilon\sim\mathcal{N}(\vec 0, \mat I)$.

The Q-network is trained using the exact same procedure as in~\citep{haarnoja_SoftActorCriticAlgorithms_2019}, with a clipped double-Q trick~\citep{fujimoto2018addressing}.
The successor features network $\successorFeatures$ is trained to minimize the Bellman error (see Eq.~\ref{eq:sf-bellman}).
The targets of the double Q-network and of the successor features network are updated at each iteration using soft target updates, in order to stabilize training~\citep{lillicrap2015continuousDDPG}.

Algorithm~\ref{algo:oursDetailed} provides a detailed description of the training procedure of \ours.

\begin{algorithm*}[t]
\caption{Detailed training procedure of \ours}
\label{algo:oursDetailed}
\newcommand{\obs}{\state}
\renewcommand{\algorithmiccomment}[1]{\hfill$\triangleright$#1}
\newcommand{\LINECOMMENT}[1]{$\triangleright$#1}

\begin{algorithmic}
\INPUT{Parameters $\theta_\policy$, $\theta_\valueFunction$, $\theta_\successorFeatures$, $\theta_\lagrange$} \COMMENT{ Initial parameters for the actor, critics and Lagrange multiplier}
\STATE{$\replay \leftarrow \emptyset$} \COMMENT{ Initialize an empty replay buffer}
\REPEAT
    \STATE{$\skillEnv \sim \uniform\left(\skillSpace\right)$} \COMMENT{ Sample skill uniformly from skill space}
    \FOR{$T$ steps}
        \STATE{\LINECOMMENT{ Environment steps}}
        \STATE{$\action_t \sim \policy(\action_t | \obs_t, \skill)$} \COMMENT{ Sample action from policy}
        \STATE{$\obs_{t+1} \sim p(\obs_{t+1} | \obs_t, \action_t, \skill)$} \COMMENT{ Sample transition from the environment}
        \STATE{$\replay \leftarrow \replay \cup \{(\obs_t, \action_t, \reward(\obs_t, \action_t), \feat(\obs_t, \action_t), \obs_{t+1}, \skill)\}$} \COMMENT{ Store transition in the replay buffer}

        \STATE{\LINECOMMENT{ Training steps}}
        \STATE{$\params_\lagrange \leftarrow \params_\lagrange - \alpha_\lagrange \nabla J_\lagrange(\params_\lagrange)$} \COMMENT{ Update Lagrange multiplier with Eq. \ref{eq:lagrange-obj}}
        \STATE{$\params_{\QFunction, i} \leftarrow \params_{\QFunction, i} - \alpha_\QFunction \nabla J_\QFunction(\params_{\QFunction, i}) \text{ for } i\in\{1,2\}$} \COMMENT{ Policy evaluation for the Q-networks \citep{haarnoja_SoftActorCriticAlgorithms_2019}}
        \STATE{$\params_\successorFeatures \leftarrow \params_\successorFeatures - \alpha_\successorFeatures \nabla J_\successorFeatures(\params_\successorFeatures)$} \COMMENT{ Policy evaluation for successor features with Eq.~\ref{eq:lossSF}}
        \STATE{$\params_\policy \leftarrow \params_\policy + \alpha_\policy \nabla J_\policy(\params_\policy)$} \COMMENT{ Policy improvement with Eq.~\ref{eq:oursSacObjective}}
        \STATE{$\temperatureSac \leftarrow \temperatureSac - \alpha_\temperatureSac \nabla J_{\temperatureSac}(\temperatureSac)$} \COMMENT{ Adjust temperature as in~\citep{haarnoja_SoftActorCriticAlgorithms_2019}}
        \STATE{\LINECOMMENT{ Update target networks}}
        \STATE{$\params_{\QFunction, i} ' \leftarrow \tau \params_{\QFunction, i} + (1-\tau) \params_{\QFunction, i} ' \text{ for } i\in\{1,2\}$}
        \STATE{$\paramsSF ' \leftarrow \tau \paramsSF + (1-\tau) \paramsSF '$}
    \ENDFOR
\UNTIL{convergence}
\end{algorithmic}
\end{algorithm*}

\subsection{Expanded Information on \oursmb}
\label{appendix:model-based}
We provide here additional details on world models, and on our implementation of \ours's model-based variant.

\subsubsection{World Models}
Learning a skill-conditioned function approximator is challenging because in general, the agent will only see a small subset of possible $(\state, \skill)$ combinations~\citep{schaul_UniversalValueFunction_2015, borsa_UniversalSuccessorFeatures_2018}. In that case, a world model can be used to improve sample efficiency.
One key advantage of model-based methods is to learn a compressed spatial and temporal representation of the environment to train a simple policy that can solve the required task~\citep{ha_WorldModels_2018}.
World models are particularly valuable for conducting simulated rollouts in imagination which can subsequently inform the optimization of the agent's behavior, effectively reducing the number of environment interactions required for learning~\citep{hafner_DreamControlLearning_2019}.
Moreover, world models enable to compute straight-through gradients, which backpropagate directly through the learned dynamics~\citep{hafner_MasteringDiverseDomains_2023}.
Most importantly, the small memory footprint of imagined rollouts enables to sample thousands of on-policy trajectories in parallel~\citep{hafner_MasteringDiverseDomains_2023}, making possible to learn skill-conditioned function approximators with massive skill sampling in imagination.

In this work, we use a Recurrent State Space Model (RSSM) from  \citet{hafner_LearningLatentDynamics_2019}.
At each iteration, the world model $\wm$ is trained to learn the transition dynamics, and to predict the observation, reward, and termination condition.
An \textit{Imagination} MDP $(\widetilde{\mathcal{S}}, \mathcal{A}, \widehat{p}, \discount)$, can then be defined from the latent states $\stateImg\in\widetilde{\mathcal{S}}$ and from the dynamics $\widehat{p}$ of $\wm$.
In parallel, \dreamer trains a critic network $\widehat{\valueFunction}(\stateImg_t)$ to regress the $\lambda$-return $\valueFunctionLambda(\stateImg_t)$~\citep{sutton_ReinforcementLearningIntroduction_2018}.
Then, the actor is trained to maximize $\valueFunctionLambda$, with an entropy regularization for exploration: $J_\policy(\params_{\policy}) = \expect{\substack{\action \sim \policy(. | \stateImg)\\ \stateImg^\prime \sim \widehat{p}(. | \stateImg, \action)}}{\sum_{t=1}^{H} \valueFunctionLambda(\stateImg_t)}$.

\subsubsection{\oursmb}
\begin{algorithm*}[t]
\caption{\oursmb}
\label{algo:ours-mb}
\newcommand{\obs}{\state}
\newcommand{\obsImg}{\stateImg}
\renewcommand{\algorithmiccomment}[1]{\hfill$\triangleright$#1}
\newcommand{\LINECOMMENT}[1]{$\triangleright$#1}

\begin{algorithmic}
\INPUT{Parameters $\theta_\policy$, $\theta_\valueFunction$, $\theta_\successorFeatures$, $\theta_\lagrange$, $\theta_\wm$} \COMMENT{ Initial parameters for the actor, critics, Lagrange multiplier and world model}
\STATE{$\replay \leftarrow \emptyset$} \COMMENT{ Initialize an empty replay buffer}
\REPEAT
    \STATE{$\skillEnv \sim \uniform\left(\skillSpace\right)$} \COMMENT{ Sample skill uniformly from skill space}
    \FOR{$T$ steps}
        \STATE{\LINECOMMENT{ Environment steps}}
        \STATE{$\action_t \sim \policy(\action_t | \obsImg_t, \skill)$} \COMMENT{ Sample action from policy}
        \STATE{$\obs_{t+1} \sim p(\obs_{t+1} | \obs_t, \action_t, \skill)$} \COMMENT{ Sample transition from the environment}
        \STATE{$\replay \leftarrow \replay \cup \{(\obs_t, \action_t, \reward(\obs_t, \action_t), \feat(\obs_t, \action_t), \obs_{t+1})\}$} \COMMENT{ Store transition in the replay buffer}
        \STATE{$\params_\wm \leftarrow \params_\wm - \alpha_\wm \nabla J_\wm(\params_\wm)$} \COMMENT{ Update world model}

        \STATE{\LINECOMMENT{ Training steps from a rollout in imagination with skills $\skillImg \sim \mathcal{U}(\skillSpace)$}}
        \STATE{$\params_\lagrange \leftarrow \params_\lagrange - \alpha_\lagrange \nabla J_\lagrange(\params_\lagrange)$} \COMMENT{ Update Lagrange multiplier with Eq. \ref{eq:lagrange-obj}}
        \STATE{$\params_\valueFunction \leftarrow \params_\valueFunction - \alpha_\valueFunction \nabla J_\valueFunction(\params_\valueFunction)$} \COMMENT{ Policy evaluation for value function with Eq.~\ref{eq:lossV}}
        \STATE{$\params_\successorFeatures \leftarrow \params_\successorFeatures - \alpha_\successorFeatures \nabla J_\successorFeatures(\params_\successorFeatures)$} \COMMENT{ Policy evaluation for successor features with Eq.~\ref{eq:lossSF}}
        \STATE{$\params_\policy \leftarrow \params_\policy + \alpha_\policy \nabla J_\policy(\params_\policy)$} \COMMENT{ Policy improvement with Eq.~\ref{eq:actor-obj-mb}}
    \ENDFOR
\UNTIL{convergence}
\end{algorithmic}
\end{algorithm*}

\begin{figure*}[t]
\centering
\includegraphics[width=\textwidth]{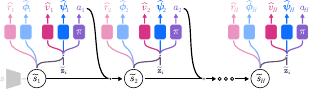}
\caption{%
Imagination rollout performed within the world model $\wm$.
Each individual rollout $i$ generates on-policy transitions following skill $\skillImg_i$, starting from a state $\stateImg_1$ for a fixed number of steps $\horizonImg$. The world model predicts $\widehat{\reward_i}$ and $\widehat{\feat_i}$ that enable to compute $\widehat{\valueFunction_i}$ and $\widehat{\successorFeatures_i}$ respectively.
}
\label{fig:ours-mb}
\end{figure*}

\oursmb's pseudocode is provided in Algorithm~\ref{algo:ours-mb}.
At each iteration, a skill $\skill$ is uniformly sampled and for $T$ steps, the agent interacts with the environment following skill $\skill$ with $\pi(\cdot | \cdot, \skill)$. At each step, the transition is stored in a dataset $\replay$, which is used to perform a world model training step.
Then, $N$ skills are uniformly sampled to perform rollouts in imagination, and those rollouts are used to (1) train the two critics $\valueFunction(\state, \skill)$, $\successorFeatures(\state, \skill)$ and (2) train the actor $\policy$.

\paragraph{World model training}
The dataset is used to train the world model $\wm$ according to DreamerV3. In addition to the reward $\widehat{\reward}_t$, we extend the model to estimate the features $\widehat{\feat}_t$, like shown on Figure~\ref{fig:ours-mb}.
\paragraph{Critic training} The estimated rewards $\widehat{\reward}_t$ and features $\widehat{\feat}_t$ predicted by the world model are used to predict the value function $\widehat{\valueFunction}$ and the successor features $\widehat{\successorFeatures}$ respectively.
Then, similarly to \dreamer, the value function $\widehat{\valueFunction}$ and successor features $\widehat{\successorFeatures}$ are trained to regress the $\lambda$-returns, $\valueFunctionLambda$ and $\successorFeaturesLambda$ respectively. The successor features target is defined recursively as follows:
\begin{equation}
\begin{split}
\label{eq:sfLambdaTargets}
    &\successorFeaturesLambda(\stateImg_t, \skillImg) = \widehat{\feat}_t + \discount \widehat{\cont}_t  \left( \left( 1 - \lambda \right) \widehat{\successorFeatures}(\stateImg_{t+1},\skillImg) +\lambda \successorFeaturesLambda(\stateImg_{t+1}, \skillImg)  \right) \\ &\text{and} \quad \successorFeaturesLambda(\stateImg_H, \skillImg) = \widehat{\successorFeatures}(\stateImg_H, \skillImg)
\end{split}
\end{equation}
\paragraph{Actor training} For each actor training step, we sample $N$ skills $\skillImg_{1}\ldots\skillImg_N\in\skillSpace$.
We then perform $N$ rollouts of horizon $H$ in imagination using the world model and policies $\pi(\cdot | \cdot, \skillImg_{i})$.
Those rollouts are used to train the critic $v$, the successor features network $\successorFeatures$, and the actor by backpropagating through the dynamics of the model.
The actor maximizes the following objective, with an entropy regularization for exploration, where $\sg{\cdot}$ represents the \textit{stop gradient} function.
\begin{equation}
\label{eq:actor-obj-mb}
J_\policy\left(\params_{\pi}\right) = \expect{\subalign{&\stateImg_{1:H}\sim \wm,\pi \\ &\skillImg\sim\mathcal{U}\left(\skillSpace\right)}}{\sum_{t=1}^{H}{\left(1 - \sg{\lagrange} \right) { \underbracket[0.5pt]{{\color{perf}\valueFunctionLambda(\stateImg_t, \skillImg)}}_{\text{Performance}} } - \sg{\lagrange}  \underbracket[0.5pt]{{\color{dist}\norm{ (1-\discount) \successorFeaturesLambda (\stateImg_t, \skillImg) - \skillImg }}}_{\text{Distance to desired skill $\skillImg$}}}}
\end{equation}

\newpage
\section{Tasks and Metrics Details}
\subsection{Tasks}
\label{appendix:tasks}
\begin{table}[h]
\caption{Tasks}
\label{tab:tasks}
\centering
\newdimen\length
\length=1.4cm
\begin{tabular}{l | c c c | c | c | c}
    \toprule
    & \multicolumn{3}{c |}{\textsc{Feet Contact}} & \textsc{Jump} & \textsc{Velocity} & \textsc{Angle}\\
    & \textsc{Humanoid} & \textsc{Ant} & \textsc{Walker} & \textsc{Humanoid} & \textsc{Ant} & \textsc{Humanoid}\\
    & \includegraphics[height=0.8\length, width=\length]{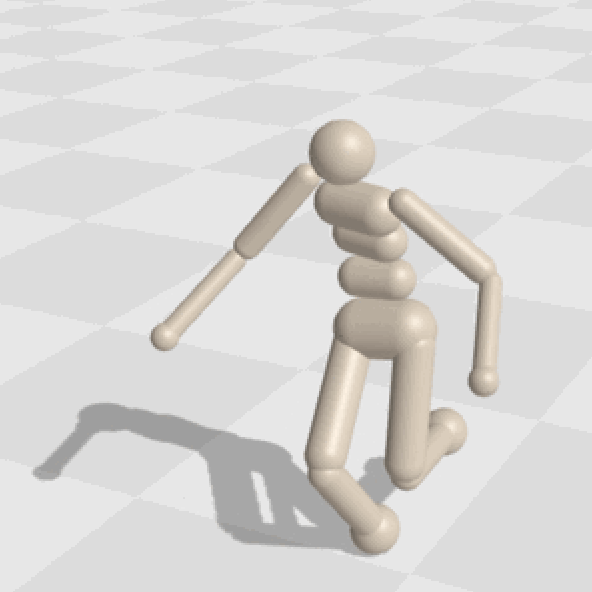} & \includegraphics[height=0.8\length, width=\length]{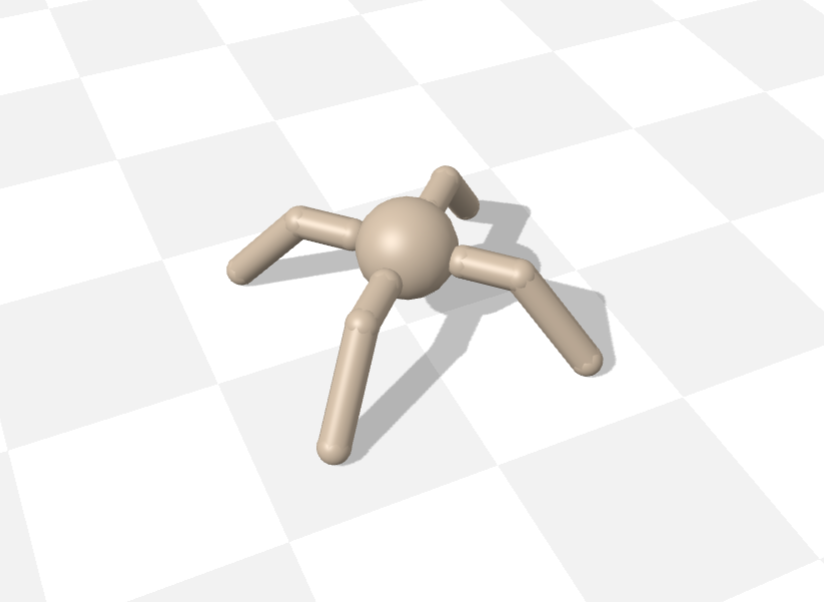} & \includegraphics[height=0.8\length, width=\length]{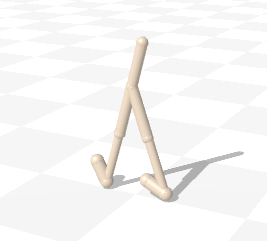} & \includegraphics[height=0.8\length, width=\length]{figures/png/Humanoid.png} & \includegraphics[height=0.8\length, width=\length]{figures/png/Ant.png} & \includegraphics[height=0.8\length, width=\length]{figures/png/Humanoid.png}\\
    \midrule
    \textsc{State dim} & 244 & 27 & 17 & 244 & 27 & 244\\
    \textsc{Action dim} & 17 & 8 & 6 & 17 & 8 & 17\\
    \textsc{Features dim} & $2$ & $4$ & $2$ & $1$ & $1$ & $1$\\
    \textsc{Features space} & $\{0, 1\}^2$ & $\{0, 1\}^4$ & $\{0, 1\}^2$ & $[0, 0.25]$ & $[-5., 5]^2$ & $]-\pi, \pi]$\\
    \textsc{Skill space} & $[0, 1]^2$ & $[0, 1]^4$ & $[0, 1]^2$ & $[0, 0.25]$ & $[-5., 5]^2$ & $]-\pi, \pi]$\\
    \textsc{Episode length} & 1000 & 1000 & 1000 & 1000 & 1000 & 1000\\
    \textsc{Threshold $\threshold$} & 0.01 & 0.1 & 0.01 & 0.0025 & 0.1 & 0.06\\
    \textsc{Distance eval $d_\text{\textnormal{eval}}$} & 0.1 & 0.3 & 0.1 & 0.025 & 1.0 & 0.6\\
    \bottomrule
\end{tabular}
\end{table}

\newpage
\subsection{Few-Shot Adaptation and Hierarchical Learning Tasks}
\label{appendix:adaptation-tasks}
For all adaptation tasks, the reward stays the same but the dynamics of the MDP is changed. The goal is to leverage the diversity of skills to adapt to unforeseen situations.
\begin{table}[h]
\caption{Adaptation tasks}
\label{tab:adaptation-tasks}
\centering
\newdimen\length
\length=2.18cm
\begin{tabular}{l | c | c | c | c | c}
    \toprule
    & \textsc{Hurdles} & \textsc{Motor Failure} & \textsc{Gravity} & \textsc{Friction} & \textsc{Wall}\\
    & \textsc{Humanoid} & \textsc{Humanoid} & \textsc{Humanoid} & \textsc{Walker} & \textsc{Ant}\\
    & \includegraphics[height=\length, width=\length]{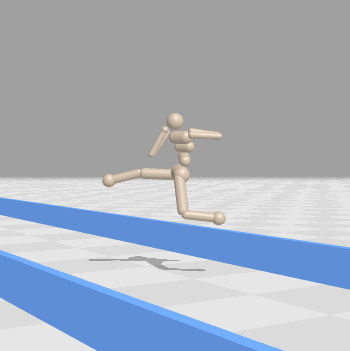} & \includegraphics[height=\length, width=\length]{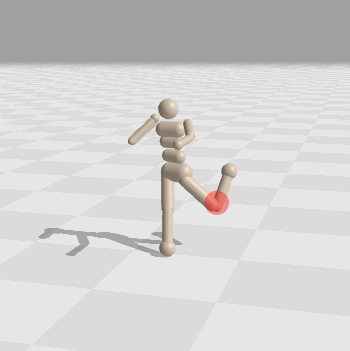} & \includegraphics[height=\length, width=\length]{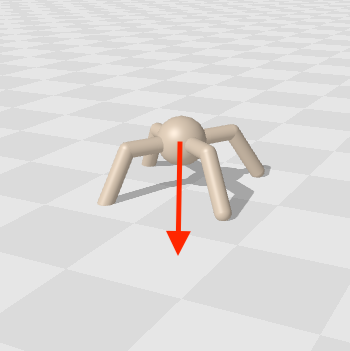} & \includegraphics[height=\length, width=\length]{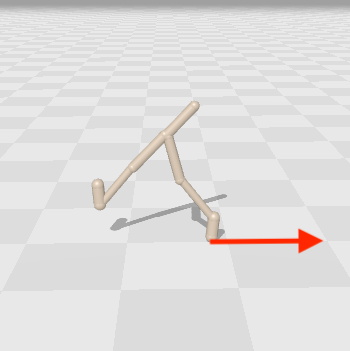} & \includegraphics[height=\length, width=\length]{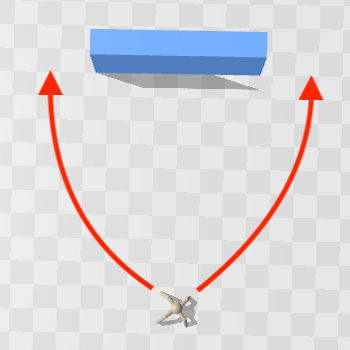}\\
    \midrule
    \textsc{Features} & Jump & Feet Contact & Feet Contact & Feet Contact & Velocity\\
    \textsc{Adaptation} & Few-shot & Few-shot & Few-shot & Few-shot & Hierarchy\\
    \bottomrule
\end{tabular}
\end{table}

\subsubsection{Few-Shot Adaptation}
For all few-shot adaptation tasks, we evaluate all skills for each replication of each method and select the best one to solve the adaptation task. In Figure~\ref{fig:adaptation}, the lines represent the IQM for the 10 replications and the shaded areas correspond to the 95\% CI.

On \textit{Humanoid - Hurdles}, we use the jump features to jump over hurdles varying in height from 0 to 50 cm.

On \textit{Humanoid - Motor Failure}, we use the feet contact features to find the best way to continue walking forward despite the damage. In this experiment, we scale the action corresponding to the torque of the left knee (actuator 10) by the damage strength (x-axis of Figure~\ref{fig:adaptation}) ranging from 0.0 (no damage) to 1.0 (maximal damage).

On \textit{Ant - Gravity}, we use the feet contact features to find the best way to continue walking forward despite the change in gravity. In this experiment, we scale the gravity by a coefficient ranging from 0.5 (low gravity) to 3.0 (high gravity).

On \textit{Walker - Friction}, we use the feet contact features to find the best way to continue walking forward despite the change in friction. In this experiment, we scale the friction by a coefficient ranging from 0.0 (low friction) to 5.0 (high friction).

\subsubsection{Hierarchical Learning}
For the hierarchical learning task, we learn a meta-controller that selects the skills of the policy in order to adapt to the new task.

On \textit{Ant - Wall}, the meta-controller is trained with SAC to select the velocity skills that enables to go around the wall and move forward as fast as possible in order to maximize performance.

\newpage
\subsection{Evaluation Metrics Details}
In this section, we illustrate how to compute and read the distance and performance profiles in Figure~\ref{fig:profiles}. In the Quality-Diversity community, there is a consensus that the best evaluation metric is the ``distance/performance profile''~\citep{flageat_BenchmarkingQualityDiversityAlgorithms_2022,grillotti_DonBetLuck_2023,grillotti_RelevanceguidedUnsupervisedDiscovery_2022,batra_ProximalPolicyGradient_2023}. This metric is also being used in skill learning for robotics~\citep{margolis_RapidLocomotionReinforcement_2022}.

The profiles are favored because it effectively captures the essence of what QD algorithms aim to achieve: not just finding a single optimal solution but exploring a diverse set of high-quality solutions.
For a given distance $d$, the distance profile shows the proportion of skills with distance to skill lower than $d$.
For a given performance $p$, the performance profile shows the proportion of skills with a performance higher than $p$.
The bigger the area under the curve, the better the algorithm is.
The profiles have similarities with the cumulative distribution functions in probability.

\begin{figure}[H]
\centering
\includegraphics[width=0.5\textwidth]{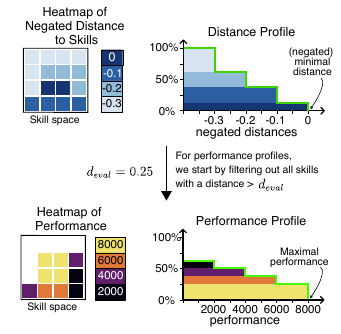}
\caption{\textbf{Left} Heatmaps of negated distance to skills and performance. \textbf{Right} Distance and performance profiles
A skill $\skill$ is considered successfully executed if $d(\skill) < d_\text{eval}$, otherwise it is considered failed. All the skills that are not successfully executed by the policy are filtered out before computing the performance heatmap and profile.
This figure also illustrates how to read the performance of the highest-performing skill (maximal performance of the agent) and the distance to skill of the best executed skill (minimal distance to skill of the agent).
}
\label{fig:profiles_explanation}
\end{figure}

\newpage
\section{Baselines details}

\begin{table}[H]
\centering
\small
\caption{Comparison of the main features for the different algorithms}

\begin{threeparttable}
\begin{tabular}{l l c c c }
\toprule

Algorithm & Objective Function & Model-based & Lagrange Multiplier $\lambda$\\

\midrule
\addlinespace

\ours & $\left(1 - \lambda\right) \sum \discount^t \reward_t - \lambda \norm{(1 - \discount) \sum \discount^t \feat_t - \skill}$ & \xmark & \cmark\\
\oursmb & $\left(1 - \lambda\right) \sum \discount^t \reward_t - \lambda \norm{(1 - \discount) \sum \discount^t \feat_t - \skill}$ & \cmark & \cmark\\

\midrule

\dcgme\tnote{$\dagger$} & $\sum \discount^t \exp(-\frac{\norm{\skill - \skill^\prime}}{l}) \reward_t$ & \xmark & \xmark\\
\qdpg\tnote{$\dagger$} & $\sum \discount^t \reward_t$ or $\sum \discount^t \tilde{\reward}_t$ & \xmark & \xmark\\

\midrule

\domino & $\left(1 - \lambda\right) \sum \discount^t \reward_t + \lambda  \sum \gamma^t \tilde{\reward}_t$ & \xmark & \cmark\\
\smerl\tnote{$\ddagger$} & $\sum \discount^t \left( \reward_t + \lambda \mathbbm{1}(R \geq R^\ast - \eps) \tilde{\reward}_t \right)$ & \xmark & \xmark\\
\smerlReverse\tnote{$\ddagger$} & $\sum \discount^t \left( \mathbbm{1}(R < R^\ast - \eps) \reward_t + \lambda \tilde{\reward}_t \right)$ & \xmark & \xmark\\

\midrule

\oursSepSkill & $\left(1-\lambda \right)\sum \discount^t \reward_t - \lambda  \sum \discount^t \norm{\feat_t - \skill} $ & \cmark & \cmark\\
\oursFixedLambda & $\left(1-\lambda \right)\sum \discount^t \reward_t - \lambda \norm{(1-\discount) \sum \discount^t \feat_t - \skill} $ & \cmark & \xmark\\
\uvfa & $\left(1-\lambda \right)\sum \discount^t \reward_t - \lambda  \sum \discount^t \norm{\feat_t - \skill} $ & \cmark & \xmark\\

\addlinespace
\bottomrule
\end{tabular}

\begin{tablenotes}
\item[$\dagger$] \dcgme and \qdpg learn diverse skills with mechanisms that are not visible in their objective function.
\item[$\ddagger$] see Section~\ref{appendix:baselines-smerl} for a detailed explanation of the reward used in \smerl and \smerlReverse.
\end{tablenotes}

\end{threeparttable}
\label{table:baselines}
\end{table}

\newcommand{\VFE}{\widetilde{\valueFunction}_e}

\newcommand{\ASF}{\widetilde{\successorFeatures}}

\subsection{\dcgme}
\dcgme~\cite{faldor_MAPElitesDescriptorConditionedGradients_2023} is a QD algorithm based on \me~\cite{mouret_IlluminatingSearchSpaces_2015}, that combines evolutionary methods with reinforcement learning to improve sample efficiency. \dcgme addresses these challenges through two key innovations: First, it enhances the Policy Gradient variation operator with a descriptor-conditioned critic. This allows for a more nuanced exploration of the solution space by guiding the search towards uncharted territories of high diversity and performance. Second, by utilizing actor-critic training paradigms, \dcgme learns a descriptor-conditioned policy that encapsulates the collective knowledge of the population into a singular, versatile policy capable of exhibiting a wide range of behaviors without incurring additional computational costs.

\subsection{\qdpg}
\qdpg~\cite{pierrot_DiversityPolicyGradient_2022} is a QD algorithm based on \me~\cite{mouret_IlluminatingSearchSpaces_2015} that integrates Policy Gradient methods with QD approaches, aiming to generate a varied collection of neural policies that perform well within continuous control environments. The core innovation of \qdpg lies in its Diversity Policy Gradient (DPG), a mechanism designed to enhance diversity among policies in a sample-efficient manner. This is achieved by leveraging information available at each time step to nudge policies toward greater diversity. The policies in the \me grid are subjected to two gradient-based mutation operators, specifically tailored to augment both their quality (performance) and diversity. This dual-focus approach not only addresses the exploration-exploitation dilemma inherent in learning but also enhances robustness by producing multiple effective solutions for the given problem.

The diversity policy gradient is based on the maximization of an intrinsic reward defined as: $\tilde{\reward}_t = \sum_{j=1}^J \norm{\feat(\state_t) - \feat(\state_j)}$, where the $J$ states $\state_j$ are coming from an archive of past encountered states.

\subsection{\domino}
\label{appendix:baselines-domino}
\domino~\citep{zahavy_DiscoveringPoliciesDOMiNO_2022} considers a set of policies $(\pi_i)_{i\in\left[1,N\right]}$, and intends to maximize simultaneously their quality while also maximizing the diversity of those that are near-optimal.
Thus each policy $\pi^i$ maximizes the following objective:
\begin{equation*}
\left(1 - \lambda\right) \sum \discount^t \reward^i_t + \lambda  \sum \gamma^t \tilde{\reward}_t^i
\end{equation*}
where $\reward^i_t$ is the task reward (also called \textit{extrinsic reward}) and $\tilde{\reward}_t^i$ is the \textit{diversity reward}.
Also, $\lambda^i$ refers to the Lagrange multiplier of policy $\pi^i$; it is used to balance between (1) the maximization of the extrinsic reward when the policy is not near-optimal, and (2) the maximization of diversity when the policy is near-optimal.
The definition of \textit{near-optimality} is given by the first policy $\pi^1$.

The first policy $\pi^1$ only maximizes the expected sum of extrinsic rewards without considering any diversity.
Hence, $\tilde{\reward}^1 = 0$ and its Lagrange multiplier $\lambda^1$ is always equal to $0$.
Its average extrinsic reward $\VFE^1$ is estimated empirically and used to define when the other policies are near-optimal.
The other policies $(\pi^i)_{i\geq 2}$ are considered near-optimal when their average extrinsic reward $\VFE^i$ is higher than $\alpha\VFE^1$ (considering $\VFE^1$ is positive) where $\alpha$ is a constant between $0$ and $1$.
If a policy is not near-optimal, its Lagrange coefficient $\lambda^i$ decreases to focus on maximizing the task reward; if it is near-optimal, the coefficient increases to give more importance to the diversity reward $\tilde{\reward}^i$.
For policies $(\pi^i)_{i\geq 2}$, the diversity reward balances between repulsion and attraction of the average features $\ASF^i$ experienced by the policies: 
\begin{equation*}
\tilde{\reward}_t^i = \left(1-\left(\frac{l_i}{l_0}\right)^3\right)\phi^i_t \cdot (\ASF^i - \ASF^{j^\ast})
\end{equation*}
where $j^\ast=\argmin_j \norm{\ASF_i - \ASF_j} $ and $l_0$ is a constant.

\subsection{\smerl and \smerlReverse}
\label{appendix:baselines-smerl}
To estimate the optimal return $R_\mathcal{M}(\policy_\mathcal{M}^*)$ required by \smerl, we apply the same method as \citet{kumar_OneSolutionNot_2020}. We trained SAC on each environment and used SAC performance $R_{\text{SAC}}$ as the the optimal return value for each environment. Similarly to~\citet{kumar_OneSolutionNot_2020}, we choose $\lambda = 2.0$ by taking the best value when evaluated on HalfCheetah environment.

We use \smerl with continuous \diayn with a Gaussian discriminator~\citep{choi_VariationalEmpowermentRepresentation_2021}, so that the policy learns a continuous range of skills instead of a finite number of skills~\citep{kumar_OneSolutionNot_2020}.
Finally, we use \diayn + prior~\citep{eysenbach_DiversityAllYou_2018,chalumeau_NeuroevolutionCompetitiveAlternative_2022} to guide \smerl and \smerlReverse towards relevant skills as explained in \diayn's original paper.

With a Gaussian discriminator $q(\skillDiayn | \state) = \mathcal{N}(\skillDiayn|\mu(\state), \Sigma(\state))$, the intrinsic reward is of the form $\tilde{\reward} = \log q(\skillDiayn | \state) - \log p(\skillDiayn) \propto \norm{\mu(\state) - \skillDiayn}^2$ up to an additive and a multiplicative constant, as demonstrated by \citet{choi_VariationalEmpowermentRepresentation_2021}. Replacing the state $\state$ with the prior information $\feat(s)$ in the discriminator gives $\tilde{\reward} \propto - \norm{\mu(\feat(\state)) - \skillDiayn}^2$.
Consequently, we can see that the intrinsic reward from \diayn corresponds to executing a latent skill $\skillDiayn$ (i.e. achieving a latent goal) in the unsupervised space defined by the discriminator $q(\skillDiayn | \feat(\state))$. Indeed, the intrinsic reward is analogous to the reward used in GCRL of the form $\reward \propto - \norm{\feat(\state) - \goal}$~\citep{liu_GoalConditionedReinforcementLearning_2022}.
Moreover, the bijection between the latent skills (i.e. latent goals) and the features (i.e. goals) is given by $\skillDiayn \sim q(\skillDiayn | \feat(\state))$.


\subsection{\texorpdfstring{\oursSepSkill}{TEXT}}
\label{appendix:sepSkill:triangular}
We can show that the constraint in \ours's objective function is easier to satisfy than \oursSepSkill's constraint.

For all skills $\skill \in \skillSpace$ and for all sequences of states $(\state_t)_{t \geq 0}$, we have:
\begin{align*}
\norm{(1 - \discount) \sum_{t=0}^\infty \discount^t \feat_{t} - \skill} &= \norm{(1 - \discount) \left(\sum_{t=0}^\infty \discount^t \feat_{t} - \sum_{t=0}^\infty \discount^t \skill\right)}\\
&= \norm{(1 - \discount) \sum_{t=0}^\infty \discount^t (\feat_{t} - \skill)}\\
&\leq (1 - \discount) \sum_{t=0}^\infty \discount^t \norm{\feat_{t} - \skill}
\end{align*}
Thus, we have the following inequality:
$$\norm{(1 - \discount) \successorFeatures(\state, \skill) - \skill} \leq (1 - \discount) \sum_{t=0}^\infty \discount^t \norm{\feat_{t} - \skill}$$

At each timestep $t$, \oursSepSkill tries to satisfy $\feat_t = \skill$, whereas \ours approximately tries to satisfy $\lim_{T\rightarrow\infty} \frac{1}{T}\sum_{t=0}^T \feat_{t} = \skill$, which is less restrictive.

\newpage
\section{Hyperparameters}
We provide here all the hyperparameters used for \ours and all baselines.
\oursmb uses the same hyperparameters as the ones used by \dreamer~\citep{hafner_MasteringDiverseDomains_2023}, hence we provide here only the parameters mentioned in this work.

The implementation of \oursmb is based on the implementation of \dreamer.
Its successor features network is also implemented as a distributional critic.
In our implementation, the Lagrange multiplier network $\lagrange$ is only conditioned on the skill $\skill$, as we noticed no difference in performance.

The hyperparameters for \smerl, \smerlReverse and \dcgme are exactly the same as in their original papers, where they were fine-tuned on similar locomotion tasks.
We also tried using hidden layers of size 512 for those baselines, in order to give them an architecture that is closer to \ours.
We noticed a statistically significant decrease in performance for \smerlReverse and \dcgme, and no statistically significant change in performance for \smerl. Each algorithm is run until convergence for $10^7$ environment steps.

The hyperparameters for \domino are based on the ones suggested by~\citet{zahavy_DiscoveringPoliciesDOMiNO_2022} and fine-tuned for our tasks.

\begin{table}[H]
\label{tab:hyperparameters-ours}
\caption{\ours hyperparameters}
\centering
\begin{tabular}{l | c}
\toprule
Parameter & Value\\
\midrule
Actor network & [512, 512, $|\mathcal{A}|$]\\
Value function network & [512, 512, 1]\\
Successor feature network & [512, 512, $|\skillSpace|$]\\
Lagrange multiplier network & [512, 512, 1]\\
Real environment exploration batch size & $256$\\
Total timesteps & $1 \times 10^{7}$ \\
Optimizer & Adam\\
Learning rate & $3 \times 10^{-4}$\\
Replay buffer size & $2 \times 10^6$\\
Discount factor $\discount$ & $0.99$\\
Target smoothing coefficient $\tau$ & $0.005$\\
\bottomrule
\end{tabular}
\end{table}

\begin{table}[H]
\label{tab:hyperparameters-ours-mb}
\caption{\oursmb hyperparameters}
\centering
\begin{tabular}{l | c}
\toprule
Parameter & Value\\
\midrule
Actor network & [512, 512, $|\mathcal{A}|$]\\
Value function network & [512, 512, 1]\\
Successor feature network & [512, 512, $|\skillSpace|$]\\
Lagrangian network & [8, 1]\\
Imagination batch size $N$ & $1024$\\
Real environment exploration batch size & $16$\\
Total timesteps & $1 \times 10^{7}$ \\
Optimizer & Adam\\
Learning rate & $3 \times 10^{-4}$\\
Replay buffer size & $10^6$\\
Discount factor $\discount$ & $0.997$\\
Imagination horizon $H$ & $15$\\
Target smoothing coefficient $\tau$ & $0.005$\\
Sampling period $T$ & $100$\\
Lambda Return $\lambda$ & $0.95$\\
\bottomrule
\end{tabular}
\end{table}

\begin{table}[H]
\label{tab:hyperparameters-dcg-me}
\caption{\dcgme hyperparameters}
\centering
\begin{tabular}{l | c}
\toprule
Parameter & Value\\
\midrule
Number of centroids & $1024$\\
Evaluation batch size $b$ & $256$\\
Policy networks & [128, 128, $|\mathcal{A}|$]\\
Number of GA variations $g$ & 128\\
\hline
GA variation param. 1 $\sigma_1$ & $0.005$\\
GA variation param. 2 $\sigma_2$ & $0.05$\\
\hline
Actor network & [256, 256, $|\mathcal{A}|$]\\
Critic network & [256, 256, 1]\\
TD3 batch size $N$ & $100$\\
Critic training steps $n$ & $3000$\\
PG training steps $m$ & $150$\\
Optimizer & Adam\\
Policy learning rate & $5 \times 10^{-3}$\\
Actor learning rate & $3 \times 10^{-4}$\\
Critic learning rate & $3 \times 10^{-4}$\\
Replay buffer size & $10^6$\\
Discount factor $\discount$ & $0.99$\\
Actor delay $\Delta$ & $2$\\
Target update rate & $0.005$\\
Smoothing noise var. $\sigma$ & $0.2$\\
Smoothing noise clip & $0.5$\\
\hline
lengthscale $l$ & 0.008\\
Descriptor sigma $\sigma_d$ & 0.0004\\
\bottomrule
\end{tabular}
\end{table}

\begin{table}[H]
\label{tab:hyperparameters-qd-pg}
\caption{\qdpg hyperparameters}
\centering
\begin{tabular}{l | c}
\toprule
Parameter & Value\\
\midrule
Number of centroids & $1024$\\
Evaluation batch size $b$ & $256$\\
Policy networks & [128, 128, $|\mathcal{A}|$]\\
Number of GA variations $g$ & $86$ \\
Number of Quality-PG variations & $85$ \\
Number of Diversity-PG variations & $85$ \\
\hline
GA variation param. 1 $\sigma_1$ & $0.005$\\
GA variation param. 2 $\sigma_2$ & $0.05$\\
\hline
Critic network --- task reward & [256, 256, 1]\\
Critic network --- diversity reward & [256, 256, 1]\\
TD3 batch size $N$ & $100$\\
Critic training steps --- task reward & $300$\\
Critic training steps --- diversity reward & $300$\\
PG training steps $m$ & $150$\\
Optimizer & Adam\\
Policy learning rate & $5 \times 10^{-3}$\\
Actor learning rate & $3 \times 10^{-4}$\\
Critic learning rate & $3 \times 10^{-4}$\\
Replay buffer size & $10^6$\\
Discount factor $\discount$ & $0.99$\\
Actor delay $\Delta$ & $2$\\
Target update rate & $0.005$\\
Smoothing noise var. $\sigma$ & $0.2$\\
Smoothing noise clip & $0.5$\\
\hline
Archive acceptance threshold & $0.1$ \\
Archive maximal size & $10^4$ \\
K-nearest neighbors & $3$ \\
\bottomrule
\end{tabular}
\end{table}

\begin{table}[H]
\label{tab:hyperparameters-domino}
\caption{\domino hyperparameters}
\centering
\begin{tabular}{l | c}
\toprule
Parameter & Value\\
\midrule
Actor network & [256, 256, $|\mathcal{A}|$]\\
Critic network --- task reward & [256, 256, 1]\\
Critic network --- diversity reward & [256, 256, 1]\\
Online batch size & $60$  \\
Batch size & $600$\\
Optimizer & Adam\\
Learning rate & $1 \times 10^{-4}$\\
Replay buffer size & $1.5\times 10^6$\\
Discount factor $\discount$ & $0.99$\\
Target smoothing coefficient $\tau$ & $0.005$\\
\hline
Number of policies & $10$ \\
Optimality ratio $\alpha$ & $0.9$\\
$\widetilde{v}^{\text{avg}}_{\pi^i}$ decay factor $\alpha_d^{\widetilde{v}^{\text{avg}}}$ & $0.9$\\
$\widetilde{\SF}^{\text{avg}}_{\pi^i}$ decay factor $\alpha_d^{\widetilde{\SF}^{\text{avg}}}$ & $0.99$\\
Lagrange learning rate & $1\times 10^{-3}$ \\
Lagrange optimizer & Adam \\
\bottomrule
\end{tabular}
\end{table}

\begin{table}[H]
\label{tab:hyperparameters-smerl}
\caption{\smerl hyperparameters}
\centering
\begin{tabular}{l | c}
\toprule
Parameter & Value\\
\midrule
Actor network & [256, 256, $|\mathcal{A}|$]\\
Critic network & [256, 256, 1]\\
Batch size & $256$\\
Optimizer & Adam\\
Learning rate & $3 \times 10^{-4}$\\
Replay buffer size & $10^6$\\
Discount factor $\discount$ & $0.99$\\
Target smoothing coefficient $\tau$ & $0.005$\\
\hline
Skill distribution & Normal distribution\\
Diversity reward scale & 10.0\\
SMERL target & $R_{\text{SAC}}$\\
SMERL margin & $0.1 R_{\text{SAC}}$\\
\bottomrule
\end{tabular}
\end{table}

\begin{table}[H]
\label{tab:hyperparameters-smerl-reverse}
\caption{\smerlReverse hyperparameters}
\centering
\begin{tabular}{l | c}
\toprule
Parameter & Value\\
\midrule
Actor network & [256, 256, $|\mathcal{A}|$]\\
Critic network & [256, 256, 1]\\
Batch size & $256$\\
Optimizer & Adam\\
Learning rate & $3 \times 10^{-4}$\\
Replay buffer size & $10^6$\\
Discount factor $\discount$ & $0.99$\\
Target smoothing coefficient $\tau$ & $0.005$\\
\hline
Skill distribution & Normal distribution\\
Diversity reward scale & 10.0\\
SMERL target & $R_{\text{SAC}}$\\
SMERL margin & $0.1 R_{\text{SAC}}$\\
\bottomrule
\end{tabular}
\end{table}

\begin{table}[H]
\label{tab:hyperparameters-sep-z}
\caption{\oursSepSkill hyperparameters}
\centering
\begin{tabular}{l | c}
\toprule
Parameter & Value\\
\midrule
Actor network & [512, 512, $|\mathcal{A}|$]\\
Critic network & [512, 512, 1]\\
Lagrangian network & [8, 1]\\
Imagination batch size $N$ & $1024$\\
Real environment exploration batch size & $16$\\
Total timesteps & $1 \times 10^{7}$ \\
Optimizer & Adam\\
Learning rate & $3 \times 10^{-4}$\\
Replay buffer size & $10^6$\\
Discount factor $\discount$ & $0.997$\\
Imagination horizon $H$ & $15$\\
Target smoothing coefficient $\tau$ & $0.005$\\
Sampling period $T$ & $100$\\
Lambda Return $\lambda$ & $0.95$\\
\bottomrule
\end{tabular}
\end{table}

\begin{table}[H]
\label{tab:hyperparameters-fixed-lambda}
\caption{\oursFixedLambda hyperparameters}
\centering
\begin{tabular}{l | c}
\toprule
Parameter & Value\\
\midrule
Actor network & [512, 512, $|\mathcal{A}|$]\\
Critic network & [512, 512, 1]\\
Successor Feature network & [512, 512, $|\skillSpace|$]\\
Lagrangian network & [8, 1]\\
Imagination batch size $N$ & $1024$\\
Real environment exploration batch size & $16$\\
Total timesteps & $1 \times 10^{7}$ \\
Optimizer & Adam\\
Learning rate & $3 \times 10^{-4}$\\
Replay buffer size & $10^6$\\
Discount factor $\discount$ & $0.997$\\
Imagination horizon $H$ & $15$\\
Target smoothing coefficient $\tau$ & $0.005$\\
Sampling period $T$ & $100$\\
Lambda Return $\lambda$ & $0.95$ \\
Lambda $\lagrange$ & 0.5\\
\bottomrule
\end{tabular}
\end{table}

\begin{table}[H]
\label{tab:hyperparameters-uvfa}
\caption{\uvfa hyperparameters}
\centering
\begin{tabular}{l | c}
\toprule
Parameter & Value\\
\midrule
Actor network & [512, 512, $|\mathcal{A}|$]\\
Critic network & [512, 512, 1]\\
Lagrangian network & [8, 1]\\
Imagination batch size $N$ & $1024$\\
Real environment exploration batch size & $16$\\
Total timesteps & $1 \times 10^{7}$ \\
Optimizer & Adam\\
Learning rate & $3 \times 10^{-4}$\\
Replay buffer size & $10^6$\\
Discount factor $\discount$ & $0.997$\\
Imagination horizon $H$ & $15$\\
Target smoothing coefficient $\tau$ & $0.005$\\
Sampling period $T$ & $100$\\
Lambda Return $\lambda$ & $0.95$ \\
Lagrange multiplier $\lagrange$ & 0.66\\
\bottomrule
\end{tabular}
\end{table}

\end{document}